\ifdefined\pdfmapfile
  \pdfmapfile{=ps2pk35.map}
  \pdfmapfile{+cm.map}
  \pdfmapfile{+cmextra.map}
  \pdfmapfile{+latxfont.map}
  \pdfmapfile{+symbols.map}
  \pdfmapfile{+rsfs.map}
\fi
\documentclass[conference]{IEEEtran}
\usepackage{amsmath,amssymb,bm}
\usepackage{times}
\usepackage[table,dvipsnames]{xcolor}
\usepackage[numbers]{natbib}
\usepackage{multicol,algorithm,array,algpseudocode}
\usepackage{graphicx,capt-of,caption,float,booktabs,placeins,afterpage}
\usepackage{multirow,pifont,wrapfig,verbatim,tabularx,makecell,arydshln}
\usepackage[bookmarks=true,linkcolor=blue,colorlinks=true,citecolor=orange]{hyperref}
\usepackage{cleveref}
\hypersetup{pdftitle={MSK-Bench},pdfauthor={Mengtao Ou et al.},pdfsubject={},pdfkeywords={}}
\newcommand{\cmark}{\textcolor{green!70!black}{\ding{51}}}
\newcommand{\xmark}{\textcolor{red}{\ding{55}}}
\newcommand{\pmark}{\textcolor{orange}{\Large $\bullet$}}
\definecolor{myblue}{rgb}{0.9,0.95,1}
\newcommand{\mskResidualFigure}{\includegraphics[width=0.90\columnwidth]{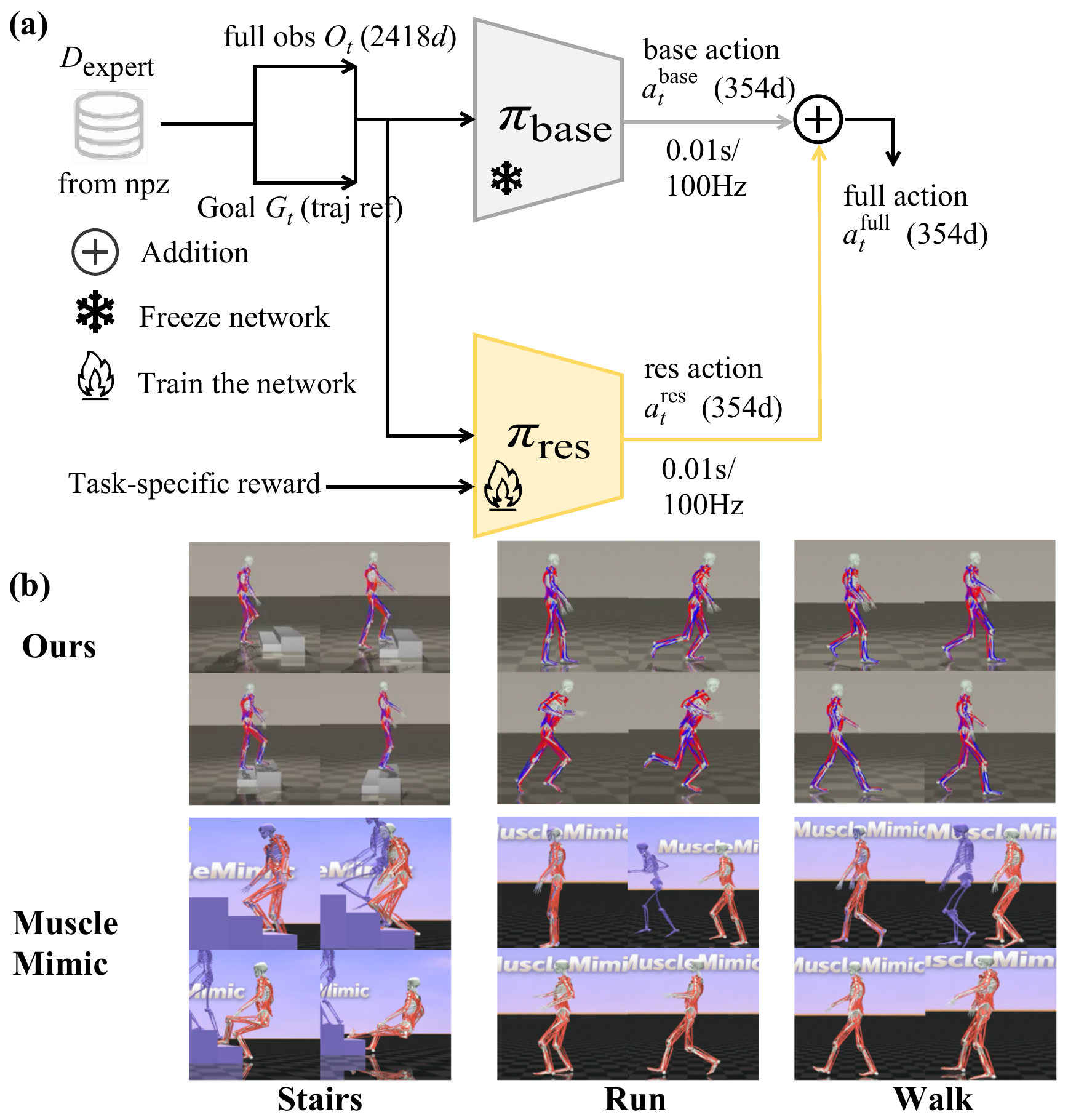}}
\title{MSK-Bench: Benchmarking Full-Body Musculoskeletal Motor Control Across Tasks, Control Paradigms, and Physiological Metrics}
\author{Mengtao Ou$^{*1}$, Zongzheng Zhang$^{*1}$, Zhenghao Xiao$^{1}$, Yixuan Pan$^{2}$, Ziwen Zhuang$^{3}$,\\ Hang Zhao$^{3}$, Hongyang Li$^{2}$, Yanan Sui$^{4}$, Libin Liu$^{5}$, Hao Zhao\textdagger$^{1}$ \vspace{0.2cm}\\
 $^1$ Institute for AI Industry Research (AIR), Tsinghua University  $^2$ The University of Hong Kong (HKU) \\
 $^3$ Institute for Interdisciplinary Information Sciences (IIIS), Tsinghua University \\
 $^4$ School of Aerospace Engineering, Tsinghua University
 $^5$ Peking University (PKU)\\
 $^*$ Equal contribution  \textdagger Corresponding author\\
 \href{https://zzongzheng0918.github.io/MSK-Bench/}{\textcolor{orange}{https://zzongzheng0918.github.io/MSK-Bench/}}
}

\begin{document}
\maketitle
\begin{figure}[t]
    \centering
    \includegraphics[width=\columnwidth]{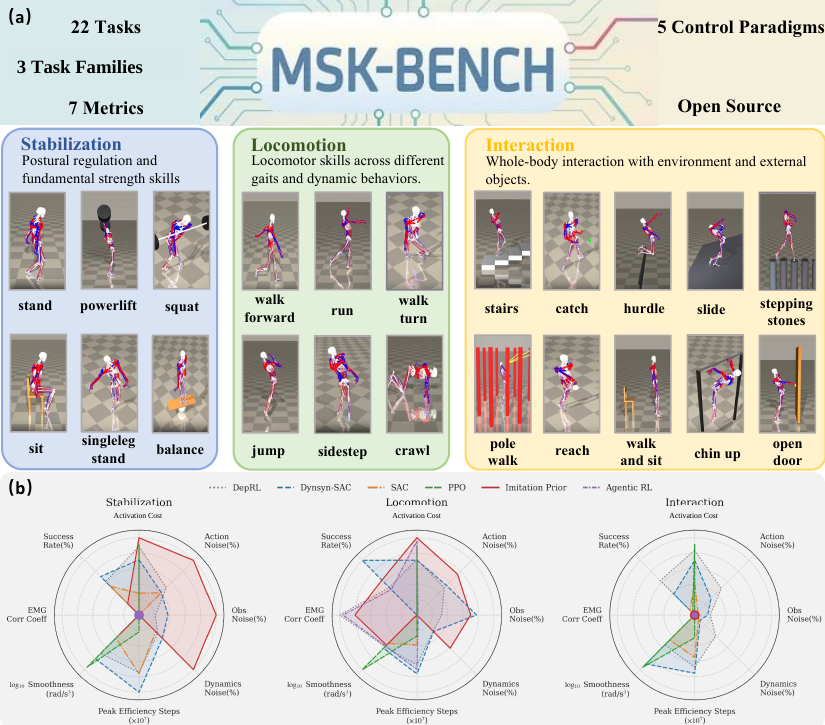}
    \caption{\textbf{Overview of the Musculoskeletal Benchmark (MSK-Bench).}
    (a) \textbf{Benchmark Components:} An open-source suite featuring
    \textbf{22} tasks across stabilization, locomotion, and interaction,
    evaluated under \textbf{5} control paradigms and \textbf{7} metric families.
    (b) \textbf{Method Evaluation:} Radar charts compare control methods
    across the \textbf{3} task families using \textbf{8} reported axes derived
    from the benchmark metrics. Each value represents the average performance
    across all applicable tasks within the respective category.
    EMG-envelope similarity is evaluated only where matched human reference
    data are available, while cumulative reward is reported separately in
    Fig.~\ref{fig:main20}.}
    \label{fig:teaser}
    \vspace{-0.5cm}
\end{figure}

\begin{abstract}
Musculoskeletal (MSK) humanoids provide a physiologically grounded embodiment for studying full-body motor control, but their high-dimensional muscle actuation, delayed activation dynamics, and redundant muscle--tendon structures make learning substantially harder than torque-driven humanoid control. Existing MSK benchmarks remain fragmented across gait, prosthetics, dexterous hands, or challenge-specific tracks, leaving full-body muscle-actuated control insufficiently evaluated under standardized tasks, methods, and metrics. We introduce MSK-Bench, a benchmark of 22 full-body motor-control tasks organized into three progressively challenging categories: postural stabilization, common locomotor behaviors, and contact-rich environmental interaction. Under unified task protocols and robustness perturbations, MSK-Bench evaluates 5 representative control paradigms, including reward-based RL, agentic reward tuning, latent-action RL, imitation-prior control, and residual adaptation over imitation priors. Beyond task success and reward, MSK-Bench further reports robustness analysis and physiology-oriented diagnostics, including activation cost, joint smoothness, and EMG-envelope similarity. Our empirical study shows that embodiment-aware exploration and structured action representations improve task coverage in high-dimensional muscle spaces, imitation priors enhance reference-compatible stabilization and locomotion but degrade under contact-rich terrain mismatch, and residual adaptation can recover successful behaviors when fixed references fail. We further find that improved task success does not necessarily imply improved physiological agreement, highlighting the importance of evaluating task performance, robustness, and physiological behavior jointly. MSK-Bench provides a task--method--metric testbed for full-body muscle-actuated humanoid control. 
\end{abstract}

\IEEEpeerreviewmaketitle
\section{Introduction}
\label{sec: intro}

Human motion emerges from the coordinated control of hundreds of muscles acting through a highly redundant and biomechanically constrained musculoskeletal system~\cite{zajac2002biomechanics, massion1992movement}. Unlike torque-driven humanoids, where actions directly specify joint torques, biological movement is generated through muscle excitation, activation-contraction dynamics and tendon transmission \citep{delp2007opensim, zajac2003biomechanics}. This physiological hierarchy introduces substantial control challenges including delayed actuation, muscle redundancy, and complex nonlinear dynamics \citep{valero2015exploring}. Consequently, musculoskeletal (MSK) models have become an increasingly important platform for studying physically grounded control \citep{seth2018opensim, caggiano2022myosuite, damsgaard2006anybody}.

Despite this progress, our understanding of how to learn effective control policies for large-scale musculoskeletal systems remains limited. 
Recent torque-driven humanoid benchmarks have substantially expanded task diversity for whole-body locomotion, loco-manipulation, and scene interaction, but they typically rely on rigid-body torque actuation (Tab.~\ref{tab:benchmark_comparison}). Conversely, existing MSK benchmarks provide physiologically grounded actuation, but they remain largely fragmented across specific domains, including gait optimization~\citep{song2021deep, kidzinski2018learning1}, prosthetics~\citep{kidzinski2019artificial}, athletic locomotion~\citep{caggiano2024myochallenge, wang2026myochallenge}, and dexterous hand control~\citep{caggiano2023myodex, xu2026music}. While these benchmarks have significantly advanced individual subfields, they are still limited in at least one of the dimensions needed for general-purpose full-body muscle-actuated control: broad whole-body task diversity, contact-rich terrain and object interaction, or standardized evaluation across multiple control paradigms. As a result, it remains unclear which control paradigms are most effective for full-body musculoskeletal agents and under what physical conditions they succeed or fail.

A second limitation of existing research lies in how musculoskeletal control is evaluated. Most existing benchmarks primarily focus on task-level metrics such as reward, success rate, or locomotion speed \citep{duan2016benchmarking}. However, unlike conventional humanoid control, musculoskeletal systems provide access to physiologically meaningful internal variables, including muscle activations, muscle synergies, and energy expenditure \citep{d2003combinations, selinger2015humans}. Multiple muscle patterns can often generate similar body motions, implying that successful task completion alone is insufficient for evaluating musculoskeletal control \citep{valero2015exploring, latash2018muscle}. A policy may achieve the correct movement while relying on unrealistic muscle coordination strategies. A recent controlled comparison makes this gap explicit: motion-only imitation and kinetics-aware imitation obtained comparable kinematic tracking, yet motion-only imitation produced substantially larger errors in ground-reaction forces, center of pressure, and joint moments~\citep{liu2026imitation}. Therefore, evaluating whether a policy behaves in a biomechanically plausible manner is equally important as evaluating whether it completes a task successfully \citep{michaud2021fair, simos2025reinforcement}.

To address these challenges, we propose \textbf{MSK-Bench}, a comprehensive full-body musculoskeletal benchmark designed to resolve current evaluation deficiencies: the lack of humanoid-level task diversity, the scarcity of rich physical interactions, and the absence of standardized physiological metrics (Tab.~\ref{tab:benchmark_comparison} provides a detailed comparison) \cite{sferrazza2024humanoidbench,kuang2025skillblender,liu2024mimicking,zhang2026lhmhumanoidlearningunifiedpolicy,he2024omnih2o,simos2025reinforcement,berg2024sar,caggiano2023myodex,li2026towards,zuo2024self,wei2026scaling,boo2025comprehensive,caggiano2025myochallenge,wang2026myochallenge2025newbenchmark}. 
Specifically, MSK-Bench encompasses \textbf{22} tasks organized into three progressively challenging categories: \textbf{stabilization}, \textbf{locomotion}, and \textbf{complex interactions} (e.g., load transport and whole-body coordination), as illustrated in Fig.~\ref{fig:teaser}a. 
Moving beyond mere task success, we establish a physiology-aware evaluation protocol that assesses biomechanical fidelity through standardized metrics, including robustness, muscle activation cost, joint smoothness and EMG-envelope similarity. 
We evaluate \textbf{five} representative control paradigms under this unified protocol to provide a comprehensive multi-dimensional analysis of their control capabilities (Fig.~\ref{fig:teaser}b). 
This evaluation establishes a strong baseline, demonstrating that residual reinforcement learning effectively combines motor priors with task adaptation to offer a robust foundation for future research.

Through \textbf{MSK-Bench}, we identify three empirical findings about full-body muscle-actuated control. \textbf{First}, controller performance is strongly task- and metric-dependent: reward-based methods exhibit different advantages across task families
and individual tasks, while focused studies reveal trade-offs in reward adaptation and action representation rather than a uniformly superior design. \textbf{Second}, motion-prior control is also task-dependent: the evaluated
imitation controller achieves strong measured success on several stabilization and locomotion skills, but fails on the tested stair task
even at zero dynamics perturbation; tracking errors characterize the resulting reference divergence without identifying its cause. \textbf{Third}, task-level success does not fully characterize muscle use. Consistent with evidence that similar kinematic tracking can conceal substantial kinetic errors~\citep{liu2026imitation}, physiology-aware evaluation reveals that success and EMG-envelope
agreement can change differently across tasks, that agreement varies across individual muscles and control settings, and that recruitment
profiles depend on task conditions and anatomical resolution. These findings motivate a diagnosis-to-validation workflow in which broad task evaluation identifies controller limitations, task-specific analysis characterizes them, and muscle-level diagnostics
assess behavior beyond task completion.

\section{Related Work}
\label{sec:related work}

\begin{table*}[t]
\centering
\scriptsize
\renewcommand{\arraystretch}{1.12}
\setlength{\tabcolsep}{3.2pt}

\caption{
\textbf{Benchmark comparison.}
\cmark/\pmark/\xmark denote full/partial/no support in the cited evaluations.
We compare actuation, full-body embodiment, scene/terrain support,
contact-rich interaction, task hierarchy, benchmark scale,
physiological evaluation, and evaluated control families.
}
\label{tab:benchmark_comparison}

\resizebox{\textwidth}{!}{%
\begin{tabular}{@{}l cc cc cc cc l@{}}
\toprule

\multirow{2}{*}{\textbf{Benchmark / Framework}}
& \multicolumn{2}{c}{\textbf{Embodiment}}
& \multicolumn{2}{c}{\textbf{Environment}}
& \multicolumn{2}{c}{\textbf{Task Design}}
& \multicolumn{2}{c}{\textbf{Evaluation}}
& \multirow{2}{*}{\textbf{Main Focus}}
\\

\cmidrule(lr){2-3}
\cmidrule(lr){4-5}
\cmidrule(lr){6-7}
\cmidrule(lr){8-9}

& \makecell{\textbf{Actuation}\\\textbf{(Dim.)}}
& \makecell{\textbf{Full}\\\textbf{Body}}
& \makecell{\textbf{3D Scene /}\\\textbf{Terrain}}
& \makecell{\textbf{Contact}\\\textbf{Rich}}
& \makecell{\textbf{Task}\\\textbf{Hierarchy}}
& \makecell{\textbf{Tasks /}\\\textbf{Data Scale}}
& \makecell{\textbf{Physio. Eval.}\\\textbf{Metrics}}
& \makecell{\textbf{Control}\\\textbf{Families}}
&
\\

\midrule
\multicolumn{10}{c}{\textbf{Torque-Driven Humanoid Benchmarks}}
\\
\midrule

HumanoidBench~\cite{sferrazza2024humanoidbench}
& Torque (19/61)
& \cmark
& \cmark
& \cmark
& \cmark
& 27 tasks
& \xmark
& 1
& Loco-manipulation
\\

SkillBench / SkillBlender~\cite{kuang2025skillblender}
& Torque (19)
& \cmark
& \pmark
& \cmark
& \cmark
& \makecell{4 skills /\\8 tasks}
& \xmark
& 1
& Skill Blending
\\

Mimicking-Bench~\cite{liu2024mimicking}
& Torque (19)
& \cmark
& \cmark
& \cmark
& \xmark
& 6 tasks
& \xmark
& 1
& Scene Interaction
\\

\midrule
\multicolumn{10}{c}{\textbf{Musculoskeletal Benchmarks}}
\\
\midrule

LocoMuJoCo~\cite{al2023locomujoco}
& Mixed
& \cmark
& \xmark
& \xmark
& \pmark
& \makecell{12 envs /\\27 tasks}
& \xmark
& 2
& Locomotion Imitation
\\

MyoSuite~\cite{caggiano2022myosuite}
& Muscle ($\leq 80$)
& \xmark
& \xmark
& \cmark
& \pmark
& 204 tasks
& \xmark
& 1
& Dexterity and Agility
\\

MyoDex~\cite{caggiano2023myodex}
& Muscle (39)
& \xmark
& \xmark
& \cmark
& \xmark
& \makecell{14 train /\\34 eval.}
& \xmark
& 1
& Hand Manipulation
\\

MS-Human-700 / MsGym~\cite{zuo2024self}
& Muscle (700)
& \cmark
& \xmark
& \pmark
& \xmark
& 3 tasks
& \cmark
& 1
& MSK Locomotion
\\

MyoChallenge 2024~\cite{wang2026myochallenge}
& Muscle ($\leq 80$)
& \pmark
& \cmark
& \cmark
& \pmark
& 2 tracks
& \xmark
& 1
& Prosthetics
\\

MyoChallenge 2025~\cite{caggiano2025myochallenge}
& Muscle ($\leq 80$)
& \pmark
& \pmark
& \cmark
& \pmark
& \makecell{2 tasks /\\4 tracks}
& \xmark
& 1
& Athletic Control
\\

\rowcolor{myblue}
\textbf{MSK-Bench (ours)}
& \textbf{Muscle (416)}
& \cmark
& \cmark
& \cmark
& \cmark
& \textbf{22 tasks}
& \cmark
& \textbf{5}
& \textbf{Full-body Terrain MSK}
\\

\bottomrule
\end{tabular}%
}

\vspace{-0.3cm}
\end{table*}

\textbf{Musculoskeletal Models and Control.}
Musculoskeletal (MSK) agents differ from torque-driven humanoids: muscle excitations are transformed by activation-contraction dynamics into muscle forces and then mapped to joint torques through tendon paths and moment arms, introducing redundancy, delay, nonlinear muscle-tendon dynamics, and biomechanical constraints~\citep{delp2007opensim,seth2011opensim,seth2018opensim}. 
Classical platforms established workflows for anatomically grounded model construction, scaling, and motion-data analysis~\citep{delp2007opensim,seth2011opensim,seth2018opensim,damsgaard2006anybody}, while MyoSim and MyoSuite brought physiologically realistic MSK models into MuJoCo-style contact-rich RL settings~\citep{todorov2012mujoco,brockman2016openai,wang2022myosim,caggiano2022myosuite}. 
Recent work further advances MSK control through model-based muscle controllers, hierarchical planning, markerless-capture imitation, and physical musculoskeletal humanoids~\citep{feng2023musclevae,wei2025motion,cotton2025kintwin,kawaharazuka2026characteristics, xu2026music}, while full-body systems have scaled toward large-scale motion imitation, richer anatomy, and whole-body behavior emulation~\citep{li2026towards,zuo2024self,wei2026scaling}. Arnold extends this direction to a transformer policy spanning 14 tasks and multiple embodiments through a compositional sensorimotor vocabulary~\citep{an2025arnold}. Complementarily, morphology--control co-design can optimize muscle strength, contraction velocity, and stiffness jointly with control, improving learning efficiency and locomotion stability across varied terrain~\citep{sun2026evolving}.
These advances motivate a systematic evaluation of full-body MSK control.

\textbf{Benchmarks for Embodied Motor Control.}
Robot-learning benchmarks have standardized progress in continuous control, offline RL~\citep{tassa2018deepmind, fu2020d4rl}, and manipulation~\citep{yu2020meta,james2020rlbench, mees2022calvin,JiangYunfan2023VGRM, liu2023libero, zhu2020robosuite,kumar2023robohive, gu2023maniskill2,tao2024maniskill3, nasiriany2024robocasa, cui2026libero, nasiriany2026robocasa365,mu2025robotwin, walke2023bridgedata, lin2021softgym}.
Recent humanoid benchmarks further move from isolated locomotion or manipulation toward whole-body interaction~\citep{sferrazza2024humanoidbench,kuang2025skillblender,liu2024mimicking,zhang2026lhmhumanoidlearningunifiedpolicy,he2024omnih2o}. However, these benchmarks primarily target torque-driven rigid-body robots rather than physiological muscle actuation.
For the MSK benchmark, early OpenSim-based challenges focused on gait, running, prosthetics, and neuromechanical locomotion control~\citep{kidzinski2018learning1, kidzinski2018learning2, kidzinski2019artificial, song2021deep, al2023locomujoco}. 
MyoSuite~\citep{caggiano2022myosuite} and MyoChallenge~\citep{caggiano2025myochallenge, wang2026myochallenge2025newbenchmark,caggiano2024myochallenge, wang2026myochallenge} introduce contact-rich MSK tasks for physiologically realistic models, spanning hand manipulation, locomotion, and athletic intelligence tracks. 
Our benchmark organizes full-body MSK tasks into a graded spectrum from postural regulation to common locomotor behaviors and contact-rich environmental interaction. It evaluates multiple control paradigms under a task-method-metric matrix, where metrics go beyond task success to include EMG-envelope activation~\cite{li2026towards, boo2025comprehensive,simos2025reinforcement} for assessing biomechanically plausible muscle coordination. This emphasis is supported by predictive locomotion results showing that a muscle-synergy prior can reduce non-physiological knee kinematics and keep joint moments closer to experimental envelopes than unconstrained muscle control~\citep{park2026synergy}.

\textbf{Learning Paradigms for Musculoskeletal Control.}
Reward-based RL provides a general route to motor control~\citep{schulman2017proximal,haarnoja2018soft}, but the high dimensionality, redundancy, and delay of muscle actuation make direct exploration inefficient in MSK agents, motivating MSK-specific exploration, action-structure, and weak-prior methods~\citep{schumacher2022dep,he2024dynsyn,wei2026scalable,berg2024sar,caggiano2023myodex}. Physiology-structured interfaces offer a complementary route: muscle-synergy actions improve biomechanical fidelity and generalization~\citep{park2026synergy}, while Lambda-Hold control predicts per-muscle equilibrium-point threshold lengths and lets a stretch-reflex law generate excitations, reducing both action dimensionality and policy-query frequency~\citep{lee2026lambda}.
Agentic reward generation further reduces manual reward engineering in robotics~\citep{yu2023language,xie2024text2reward,ma2024eureka,wang2024gensim}, yet its reliability under physiologically constrained MSK dynamics remains unclear. 
A complementary family relies on motion references and motor priors: references can be reconstructed from human motion or generated by large motion models~\citep{uhlrich2023opencap,gilon2026opencap,rempe2026kimodo,wen2025hy,wang2026motionbricks}, while multimodal resources such as ULTRA-MoCap pair wearable IMU and sEMG measurements with high-fidelity upper-limb kinematics for physiological training and validation~\citep{fritsche2026ultra}. Imitation and prior-based controllers have enabled both physics-based character control and muscle-actuated motion imitation~\citep{li2026towards,simos2025reinforcement,peng2018deepmimic,peng2021amp,peng2022ase,tessler2023calm,mu2025smp}.
However, reference tracking alone may fail on contact-rich or out-of-distribution tasks, motivating residual adaptation that learns corrective actions on top of existing controllers or skill policies~\citep{johannink2019residual,silver2018residual,alakuijala2021residual,rana2023residual,lin2026peel,xu2026compliant,ankile2025residual,zhao2025resmimic,sun2025robotdancing,yan2024ctrl,han2026muscle}. Reflex-informed neuromuscular RL follows this prior-plus-residual pattern: a fixed phase-dependent reflex controller supplies the base behavior, while RL adjusts four biomechanically meaningful reflex gains and thresholds, retaining robustness under muscle weakness and external perturbations~\citep{zhou2026reflex}.
Our benchmark jointly evaluates reward-based RL, agentic reward tuning, latent-action RL, imitation-prior control, and residual adaptation under the same full-body MSK task suite.

\section{MSK-Bench}
\label{sec: benchmark}

As shown in Fig.~\ref{fig:teaser}a, MSK-Bench evaluates three foundational functional regimes of gross full-body motor control: maintaining postural stability, generating self-propelled movement, and coordinating physical interaction with external terrain or objects~\cite{newell2020fundamental}. These regimes progressively broaden the control demands from body stabilization to dynamic locomotion and environmental interaction. Accordingly, the 22 tasks are organized into three families:
1) \textbf{Stabilization} comprises six tasks---standing, single-leg standing, squatting, sitting, balancing, and powerlifting---that evaluate static or quasi-static postural control under body weight and external loading, emphasizing balance and trunk stabilization~\cite{jonsson2004one,bressel2009effect,coratella2022front,stokes2011abdominal};
2) \textbf{Locomotion} comprises six tasks---walking, running, jumping, crawling, sidestepping, and turning---that evaluate self-propelled movement, requiring coordinated lower-limb recruitment, dynamic balance, and impact handling~\cite{kuo2010dynamic,neptune2019dynamic,zajac2002biomechanics,liu2008muscle,hamner2010muscle};
3) \textbf{Interaction} comprises ten tasks---stairs, hurdles, stepping stones, sliding terrain, reaching, catching, chin-up, door opening, pole walking, and walk-and-sit---that evaluate whole-body coordination under terrain variation and object contact, combining lower-body support, trunk stabilization, and task-dependent upper-limb or grip control~\cite{caggiano2022myosuite,sferrazza2024humanoidbench,moffet1993load,doma2013kinematic,snarr2017electromyographical}.
Together, these families assess whether muscle-actuated agents can balance, move, and interact with environment, while supporting physiological analysis across diverse movement regimes.

\begin{figure*}[t!]
    \centering
    \includegraphics[width=0.90\linewidth]{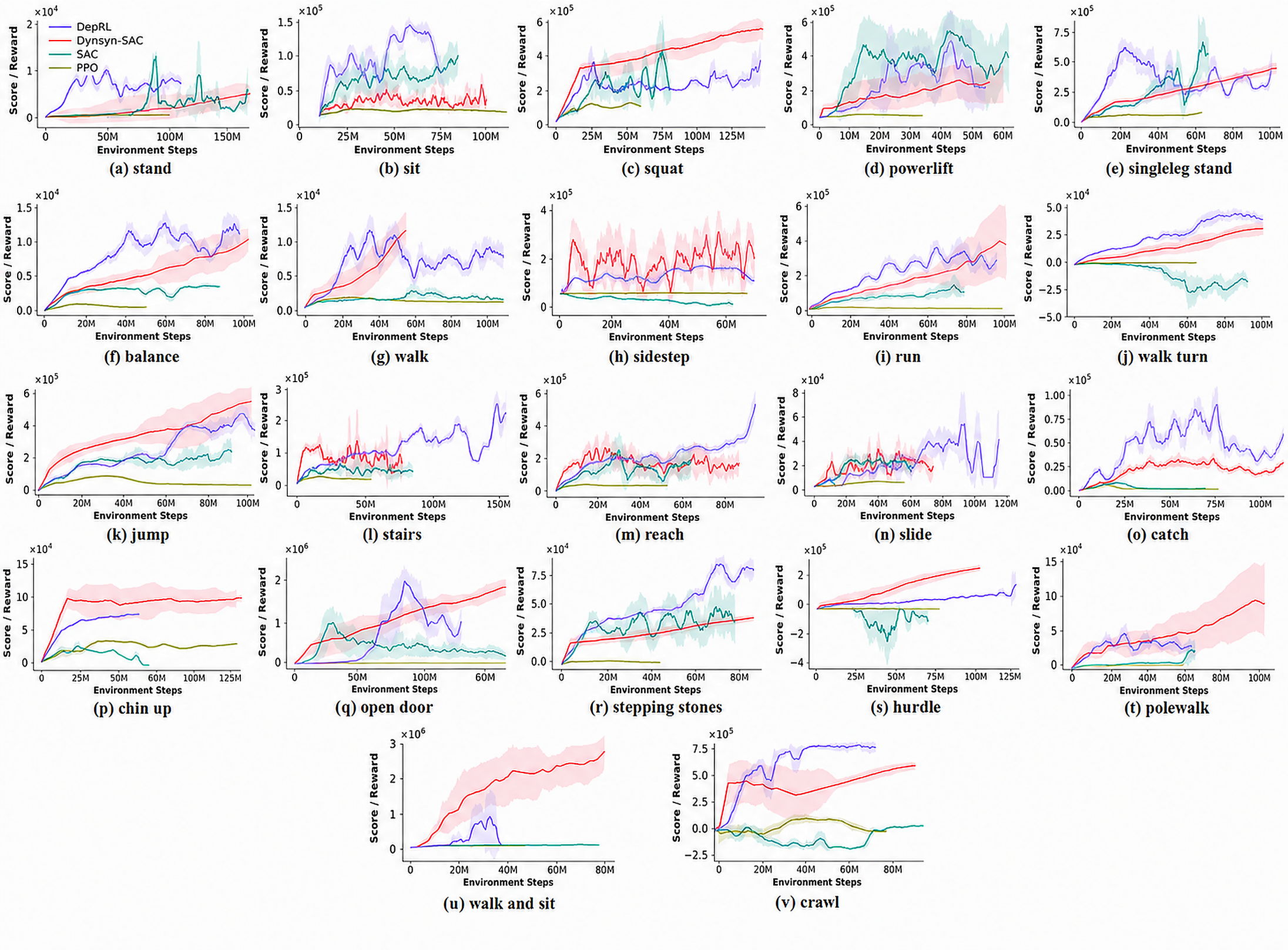}
    \caption{\textbf{Task-specific reward-learning curves for 22 tasks.} Four reward-based algorithms show task-dependent advantages. Shading shows across-run variation; Table~\ref{tab:aggregate_results} separately reports success, robustness, and nonzero-success coverage over all 22 tasks.}
    \label{fig:main20}
    \vspace{-0.4cm}
\end{figure*}

\textbf{Experimental Setup.}
All 22 tasks use MuJoCo and the same 416-muscle elastic-tendon full-body model; the 700-muscle model is used only in the anatomy study. Tasks share a muscle-action interface and base proprioceptive observations, supplemented by task-dependent phase, terrain-height, or obstacle signals. Rewards combine task progress with regime-specific physical constraints. 

\textbf{Evaluation Protocol.} Episode success requires $\mathcal{C}_{\mathrm{task}}^{(i)}\land\mathcal{C}_{\mathrm{alive}}^{(i)}\land\mathcal{C}_{\mathrm{safe}}^{(i)}$; criteria and termination rules are fixed per task, with exact thresholds provided in the supplementary material. Within each task, the four full-suite baselines share observations, rewards, termination conditions, and evaluation. Focused cross-paradigm studies use separately specified matched protocols applied identically to all methods within each comparison; their success rates are not pooled with the full-suite benchmark results. We evaluate seven complementary metric families across three dimensions: \textbf{task effectiveness and learning efficiency} (Success Rate, Cumulative Reward, and Peak-Efficiency Steps), \textbf{robustness} (Perturbation Robustness), and \textbf{physiological behavior} (Activation Cost, Joint Smoothness, and EMG-Envelope Similarity). 1) \textbf{Success Rate} ($\mathrm{SR}_{\mathrm{train-env}}$) is the fraction of evaluation episodes satisfying the above success conditions in the restored training environment, including native noise. 2) \textbf{Cumulative Reward} is the task-specific evaluation return and is never pooled across tasks. 3) \textbf{Peak-Efficiency Steps} is $T_{\max}=\arg\max_{t\in\mathcal{T}}\mathbb{E}[R_t]$, the evaluated training step attaining the highest mean return. All-task and family success rates are task-wise arithmetic means. 4) \textbf{Perturbation Robustness} evaluates separate action, observation, and muscle-dynamics sweeps using Gaussian action/observation noise and uniformly sampled multiplicative muscle-force-gain perturbations. Action/dynamics scales are $(0,0.05,0.10,0.15,0.20)$ and observation scales are $(0,0.02,0.05,0.08,0.10)$; the score is the normalized trapezoidal area under success versus perturbation scale, averaged equally over tasks and perturbation types. 5) \textbf{Activation Cost}, $E=(TM)^{-1}\sum_{t,m}a_{m,t}^{2}$, is the mean squared muscle activation and serves as an effort proxy. 6) \textbf{Joint Smoothness} uses $j_{q,t}=(v_{q,t}-2v_{q,t-1}+v_{q,t-2})/\Delta t^2$; its squared magnitude is averaged over joints and time before logarithmic display. Both are interpreted only for successful, behaviorally comparable policies. 7) \textbf{EMG-Envelope Similarity}, following the activation-envelope interpretation used in MuscleMimic~\cite{li2026towards}, computes per-muscle Pearson correlation $r_m=\mathrm{corr}(a_{\mathrm{sim},m},e_{\mathrm{EMG},m})$ and reports $r_{\mathrm{EMG}}=M^{-1}\sum_m r_m$. Signals are resampled to 101 gait-cycle points, min--max normalized per muscle, cycle-averaged, and cyclically shifted to maximize correlation, so the metric reflects phase-optimized waveform shape rather than absolute amplitude or timing.

\textbf{Baselines.}
We consider \textbf{five} control paradigms within a common policy--action--objective formulation. Let $p$ index the paradigm, $s_t$ denote the simulator state, $o_t$ its observation, and $c_t$ optional reference or prior information:
\begin{equation}
\label{eq:unified_control}
\begin{aligned}
\mathbf{u}_t &\sim \pi_{\theta}^{p}(\cdot\mid o_t,c_t),\\
\mathbf{a}_t &= F_p(o_t,\mathbf{u}_t,c_t)\in[-1,1]^M,\\
J_p(\theta;\mathbf{w}_{p,k})
&=\mathbb{E}_{\tau\sim P,\pi_\theta^p,F_p}\!\left[\sum_{t=0}^{T-1}\gamma^t
\widetilde r_{p,t}\right],\\
\widetilde r_{p,t}
&=\mathbf{w}_{p,k}^{\top}\mathbf{f}_{p,t}
-\lambda_{p,k}\Omega_{p,t}+\beta_p\mathcal{H}_{p,t}.
\end{aligned}
\end{equation}
Here $P$ is the shared transition model, $M$ the number of muscles, $\mathbf{u}_t$ the learned command, and $\mathbf{a}_t$ the executed excitation. The action map $F_p$, reward features $\mathbf{f}_{p,t}$, penalties $\Omega_{p,t}$, and entropy $\mathcal{H}_{p,t}$ specify each paradigm; adaptive weights remain fixed within stage $k$.

1) \textbf{Reward-based RL} uses fixed task rewards with direct muscle actions or an algorithm-specific action map $F_p$. We evaluate PPO~\cite{schulman2017proximal}, SAC~\cite{haarnoja2018soft}, DepRL~\cite{schumacher2022dep}, and DynSyn-SAC~\cite{he2024dynsyn}, covering standard, exploration-enhanced, and action-structured methods.
2) \textbf{Agentic reward tuning} uses DepRL as the base learner and updates reward coefficients through bounded reasoner feedback, $\mathbf{w}_{k+1}=U(\mathbf{w}_k,\mathbf{h}_k)$. We evaluate DeepSeek-V4-Flash~\cite{deepseekai2026deepseekv4} and GPT-6 Astra~\cite{openai2026gpt6astra} as reasoners.
3) \textbf{Latent-action RL} filters commands through an expert-trained, state-conditioned encoder--decoder. The latent dimension constrains the representation bottleneck while the executed muscle-action space remains unchanged.
4) \textbf{Imitation-prior control} uses a pretrained MuscleMimic~\cite{li2026towards} policy, introducing reference information through $c_t$ and motion-tracking objectives.
5) \textbf{Residual adaptation over imitation priors} combines a frozen prior with a bounded, task-specific muscle-action correction:
\begin{equation}
\mathbf{a}_t^{\mathrm{full}}
=
\operatorname{clip}\!\left(
\mathbf{a}_t^{\mathrm{base}}
+\alpha\operatorname{clip}(\mathbf{a}_t^{\mathrm{res}},-1,1),
-1,1
\right).
\label{eq:residual_rl}
\end{equation}
The four \textbf{reward-based} algorithms are trained and evaluated on all 22 tasks to form the full-suite leaderboard. \textbf{Agentic} reward tuning and \textbf{latent-action RL} are focused studies; \textbf{MuscleMimic} is evaluated on five tasks, and \textbf{residual adaptation} on walk, run, and stairs. These studies differ in task coverage, and some modify reward coefficients or introduce reference/prior objectives, so they are reported separately from the full-suite ranking.

\begin{table}[t!]
\centering
\setlength{\tabcolsep}{3pt}
\renewcommand{\arraystretch}{1.05}
\caption{Full-suite point estimates. SR retains training noise. Means weight tasks equally. Robustness areas are normalized to 0--100; Avg. weights the three perturbation types equally. Coverage: SR $>0$. Bold: row best.}
\label{tab:aggregate_results}

\begin{tabular}{@{}llrrrr@{}}
\toprule
Group & Metric & DynSyn-SAC & DepRL & SAC & PPO \\
\midrule

\multirow{4}{*}{SR (\%) $\uparrow$}
& Stabilization (6) & \textbf{57.33} & 50.33 & 42.33 & 0.00 \\
& Locomotion (6)    & \textbf{81.33} & 41.33 & 0.00  & 0.00 \\
& Interaction (10)  & 31.20 & \textbf{51.40} & 2.80 & 7.40 \\
& Average (22)          & \textbf{52.00} & 48.36 & 12.82 & 3.36 \\

\midrule

\multirow{4}{*}{Robust. $\uparrow$}
& Action      & 46.66 & \textbf{53.06} & 15.36 & 3.10 \\
& Observation & \textbf{43.37} & 38.10 & 4.26 & 3.40 \\
& Dynamics    & 34.75 & \textbf{39.84} & 14.50 & 2.92 \\
& Average     & 41.59 & \textbf{43.67} & 11.37 & 3.14 \\

\midrule

Coverage $\uparrow$
& SR $>0$
& 16/22 & \textbf{22/22} & 4/22 & 1/22 \\

\bottomrule
\end{tabular}
\vspace{-0.4cm}
\end{table}

\textbf{Results \& Analysis.}
We use MSK-Bench to address three questions.
First, \textit{do aggregate scores adequately characterize controller performance?}
Task-family comparisons reveal ranking reversals and differences in coverage and robustness, while focused studies examine the effects of reward adaptation and action compression \textbf{(T1)}.
Second, \textit{where do motion priors succeed, and how can their failures be characterized?}
Task-specific robustness tests reveal contrasting outcomes across movements, and tracking-error analysis localizes divergence on stairs without establishing its cause \textbf{(T2)}.
Third, \textit{what aspects of muscle use are omitted by task-level scores, and how can they be assessed?}
Paired success and EMG results expose differences beyond task completion, while muscle-resolved, task-dependent, and anatomical analyses characterize where those differences occur \textbf{(T3)}.

\begin{figure*}[t!]
    \centering
    \includegraphics[width=0.90\textwidth]{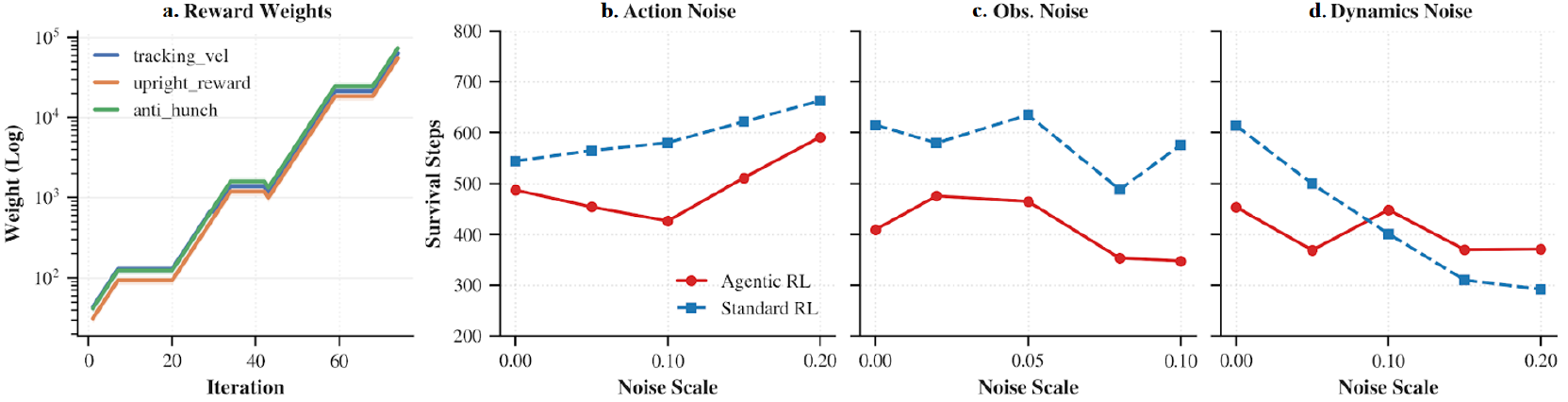}
    \caption{\textbf{Reward adaptation and robustness.} (a) Coefficient updates; (b--d) survival under perturbations. Agentic-DepRL and fixed-reward DepRL share learner, architecture, and training budget. Each point averages 50 held-out seeds; survival is capped at 1000 steps.}
    \label{fig:agentic_main}
    \vspace{-0.3cm}
\end{figure*}

\textbf{T1--Breadth: controller rankings depend on task family and metric.}
Fig.~\ref{fig:main20} shows task-dependent reward-learning trajectories for the
four reward-based algorithms on all 22 tasks. Over the displayed
overlapping training ranges, DynSyn-SAC generally attains higher returns
than DepRL on jump and hurdle, whereas DepRL attains higher returns on
walk-turn and catch (Fig.~\ref{fig:main20}j,k,o,s). SAC also remains competitive on
powerlift (Fig.~\ref{fig:main20}d). Thus, the learning curves reveal differences not
only across task families, but also among tasks within the same family. Tab.~\ref{tab:aggregate_results} separately quantifies success, coverage, and robustness over
the full suite. DynSyn-SAC attains higher mean success, driven primarily by locomotion, whereas DepRL performs better on interaction tasks. DepRL achieves nonzero observed success on 22/22 tasks, compared with 16/22 for DynSyn-SAC, and has higher action- and dynamics-robustness scores; DynSyn-SAC has a higher observation-robustness score. These results concern separately trained task-specific policies and show why mean success alone is insufficient to characterize controller performance.

\begin{figure*}[t!]
    \centering
    \includegraphics[width=1.0\textwidth]{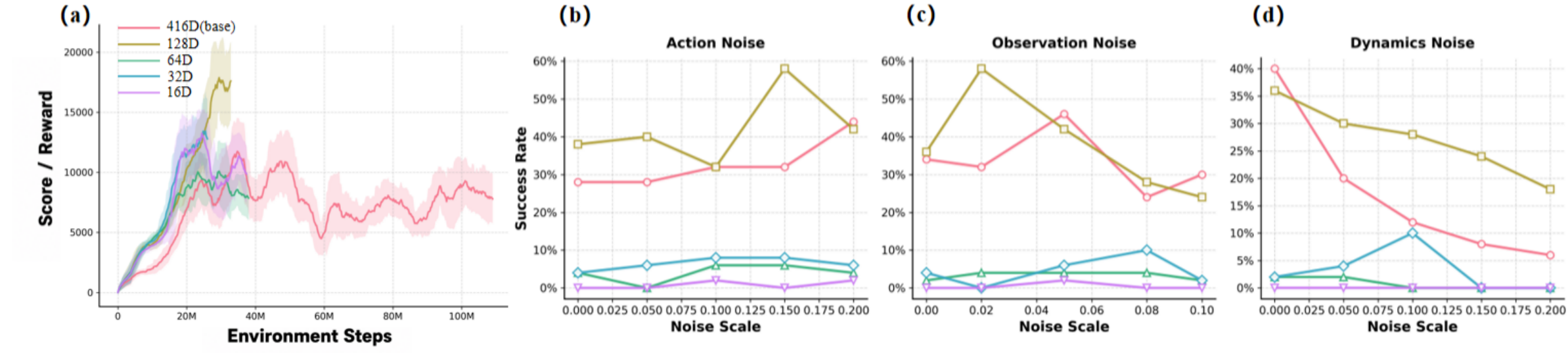}
    \caption{\textbf{Action compression.} Learning and perturbation tests for 416 actions and 128-, 64-, 32-, and 16-dimensional representations. The focused study improves at 128 dimensions but degrades under stronger compression; it is not a full-suite comparison.}
    \label{fig:latent_main}
    \vspace{-0.5cm}
\end{figure*}

\textit{Complementary diagnostics: objectives and action representations.} The following focused studies examine additional design choices. In the original
Agentic-DepRL setting, the reasoner receives mean forward speed and
torso pitch and updates reward coefficients every 20,000 steps, with
each nonzero-weight change limited to 20\%. Despite these bounded
updates, the unnormalized coefficient magnitudes grow over time (Fig.~\ref{fig:agentic_main}a). Fixed-reward DepRL maintains higher survival under the
tested action and observation perturbations, whereas agentic tuning
yields higher survival at the larger dynamics-perturbation scales (Fig.~\ref{fig:agentic_main}b-d). Reward adaptation therefore provides no uniform survival
advantage.

\begin{table}[!t]
\centering
\setlength{\tabcolsep}{4pt}
\caption{\textbf{Matched reasoner comparison}.}
\label{tab:reasoner_comparison}
\resizebox{\linewidth}{!}{%
\begin{tabular}{@{}lrrr@{}}
\toprule
Setting & Survival (steps) $\uparrow$ & Tracking RMSE $\downarrow$ & Activation $\downarrow$ \\
\midrule
Fixed reward & 417 & 0.4446 & 0.4769 \\
DeepSeek & 246.6 & 0.4355 & 0.3999 \\
GPT-6 Astra & 360 & 0.4324 & 0.3879 \\
\bottomrule
\end{tabular}%
}
\vspace{-0.5cm}
\end{table}

In a matched comparison, DeepSeek and GPT-6 Astra use the same DepRL
learner, architecture, training budget, and evaluation, with an
identical reward-coefficient constraint $\|w\|_1=40$. GPT-6 Astra has better reported point estimates than DeepSeek for
survival, tracking RMSE, and activation (Tab.~\ref{tab:reasoner_comparison}). Relative to
fixed-reward DepRL, however, it yields shorter survival
(360 vs.\ 417 steps) despite lower RMSE and activation, illustrating
a trade-off among the reported metrics.

A separate diagnostic tests GPT-6 Astra without training or a learned
low-level controller. Full experimental details are provided in
App.~\ref{sec:app-gpt6_direct_details}. It directly outputs 354-dimensional muscle
excitations at 5~Hz in a 100-Hz simulation, with each action held for
0.2~s. One seed-0 rollout is evaluated for each of walk, run, and
stairs; all three terminate in falls within the 3-s horizon
(Tab.~\ref{tab:gpt_direct_control}). These outcomes characterize the tested
direct-action configuration rather than establishing a general limitation of
GPT-6 control.

\begin{table}[t!]
\centering
\setlength{\tabcolsep}{3.5pt}
\caption{\textbf{GPT-6 direct muscle-control diagnostic.}
Div.: first sustained tracking divergence; N/A indicates that the
fixed divergence criterion was not met.}
\label{tab:gpt_direct_control}
{\small
\begin{tabular}{@{}lcccc@{}}
\toprule
Task & Duration (s) $\uparrow$ & Div. (s) $\uparrow$
& Tracking RMSE $\downarrow$ & Outcome \\
\midrule
Walk   & 0.75 & 0.53 & 0.158 & Fall \\
Run    & 2.02 & 0.94 & 0.653 & Fall \\
Stairs & 0.80 & N/A  & 0.173 & Fall \\
\bottomrule
\end{tabular}}
\vspace{-0.5cm}
\end{table}

Finally, latent-action RL uses an expert-trained, state-conditioned
encoder--decoder while retaining the executed muscle-action space.
The 128-dimensional setting achieves higher returns over the
displayed overlapping training interval and higher success at
several perturbation scales than the 416-dimensional baseline,
whereas stronger compression degrades performance (Fig.~\ref{fig:latent_main}). The reported reconstruction and explained-variance statistics also
deteriorate under stronger compression, consistent with a
representation trade-off rather than an unrestricted benefit from
dimensionality reduction. Together, these focused studies reinforce
the need to evaluate reward adaptation and action representations
against task-specific outcomes and multiple metrics.

\begin{figure*}[t!]
    \centering
    \includegraphics[width=1.0\linewidth]{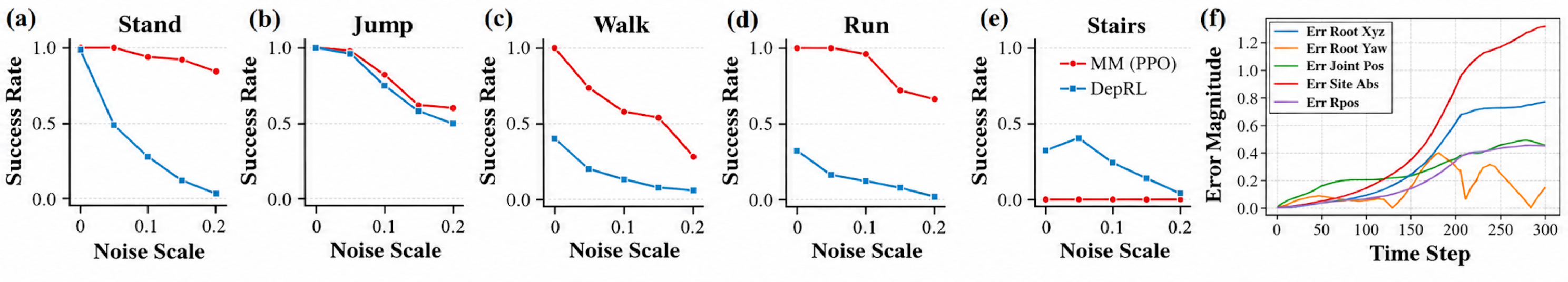}
    \caption{\textbf{Task-dependent performance under dynamics perturbations
and stair-tracking errors.} (a--e) Matched-protocol success rates of MuscleMimic/MM (PPO) and DepRL. Within each task, both methods use identical success conditions and perturbation settings. This focused cross-paradigm protocol is distinct from the full-suite benchmark protocol reported in the robustness tables; its values are not directly comparable with those tables and are excluded from aggregate robustness scores.
MuscleMimic has no observed stair successes at any tested
scale, including zero dynamics perturbation. (f) Reported tracking-error components on stairs characterize
deviation from the reference, without identifying a timing
mismatch or the cause of task failure.}
    \label{fig:musclemimic_vs_deprl}
    \vspace{-0.4cm}
\end{figure*}

\begin{figure*}[t!]
    \centering
    \includegraphics[width=1.0\linewidth]{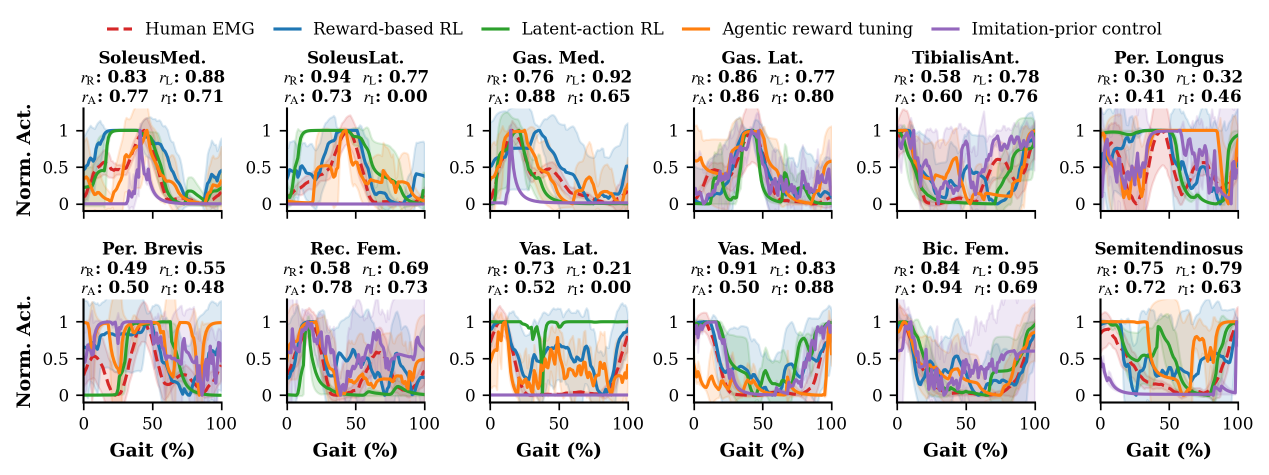}
    \caption{\textbf{Walking EMG-envelope comparison.} Human recordings~\cite{boo2025comprehensive} versus simulated activations. Normalized, phase-optimized correlations ($r_{\mathrm{R}},r_{\mathrm{L}},r_{\mathrm{A}},r_{\mathrm{I}}$) correspond to reward-based RL, latent-action RL, agentic tuning, and imitation-prior control.}
    \label{fig:emg_similarity}
    \vspace{-0.5cm}
\end{figure*}

\textbf{T2--Diagnosis: task-specific tests characterize
a stair-control failure.}
Under the matched cross-paradigm protocol in Fig.~\ref{fig:musclemimic_vs_deprl}, MuscleMimic generally
achieves higher diagnostic success than DepRL on stand, jump,
walk, and run (Fig.~\ref{fig:musclemimic_vs_deprl}a--d), with a smaller gap on jump.
On stairs, however, MuscleMimic has no observed successes
at any tested scale, including zero dynamics perturbation,
whereas DepRL retains nonzero success (Fig.~\ref{fig:musclemimic_vs_deprl}e). The stair limitation is therefore present at the
zero-perturbation setting, rather than arising only when
perturbations increase. Success on the other evaluated skills
does not establish suitability for this task.

The reported stair-tracking errors characterize increasing
deviation from the reference (Fig.~\ref{fig:musclemimic_vs_deprl}f). Between steps 150 and 300, absolute-site error rises from
0.35 to 1.31, and root-XYZ error from 0.25 to 0.77.
A mismatch between reference progression and executed motion
is one possible explanation, but these errors neither establish
a timing mismatch nor identify the cause of task failure.
Motivated by this task-specific failure, Sec.~\ref{sec:residual_rl_policy} evaluates
a combined adaptation pipeline involving residual actions,
task-objective changes, and reference pacing; \textbf{T3} assesses
task success and EMG-envelope similarity separately.

\textbf{T3--Physiology-aware evaluation: task success does not characterize muscle use.} MSK-Bench complements task outcomes with muscle-resolved
diagnostics and, where human recordings are available,
EMG-envelope comparisons. These analyses distinguish task
completion from the muscle activity underlying the reported
behavior.

\emph{Task outcomes and human-reference agreement.}
Tab.~\ref{tab:residual_results} illustrates this distinction using the combined
adaptation pipeline described in Sec.~\ref{sec:residual_rl_policy}.
Walking retains 100\% SR while correlations for the three
selected muscles increase from 0.71/0.69/0.63 to
0.75/0.80/0.67, revealing changes not captured by success
alone. Running improves from 90\% to 100\% SR and from
0.43 to 0.77 in mean EMG correlation. In contrast, stairs
recovers from 0\% to 100\% SR while mean correlation changes
only from 0.51 to 0.52, remaining below DepRL's 0.56.
Thus, an unchanged success rate can accompany changes in
muscle-level agreement, while a large success gain can
accompany only a small numerical change in the EMG metric.

\emph{Muscle-specific differences across control settings.}
Fig.~\ref{fig:emg_similarity} further resolves walking agreement by muscle.
Latent-action RL has higher correlation than reward-based
RL for biceps femoris (0.95 vs.\ 0.84), but lower correlation
for vastus lateralis (0.21 vs.\ 0.73). Such reversals reveal
muscle-specific differences that an aggregate score does
not describe. Human references are Gait120~\cite{boo2025comprehensive} for walking
and stairs and the 3-m/s recordings of Van Hooren and
Meijer~\cite{vanhooren2024running} for running. Because the signals are normalized
and phase-optimized, these comparisons assess
activation-envelope shape rather than absolute amplitude
or timing.

\begin{figure}[tb]
    \centering
    \includegraphics[width=1.0\linewidth]{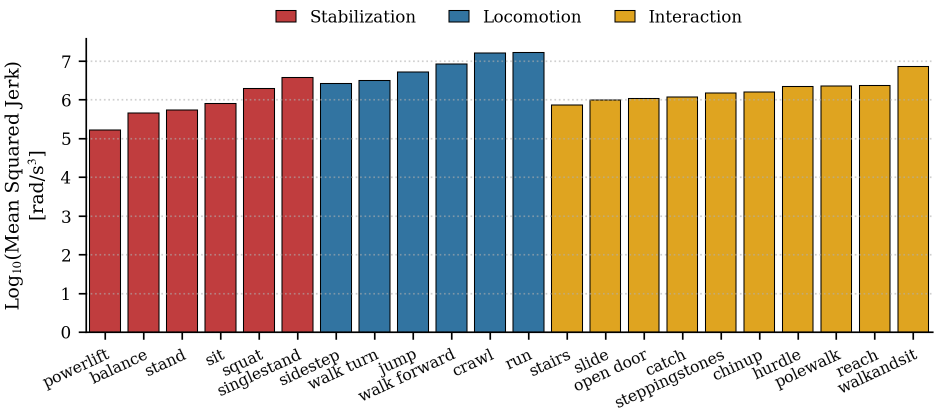}
    \caption{\textbf{Task-wise joint smoothness.} Bars report $\log_{10}$ mean squared angular jerk. Stabilization shows the lowest measured jerk, locomotion the highest, and interaction intermediate values. These characterize motion fluctuations, not success or human similarity.}
    \label{fig:joint_smoothness}
    \vspace{-0.5cm}
\end{figure}

\begin{figure}[tb]
    \centering
    \includegraphics[width=1.0\linewidth]{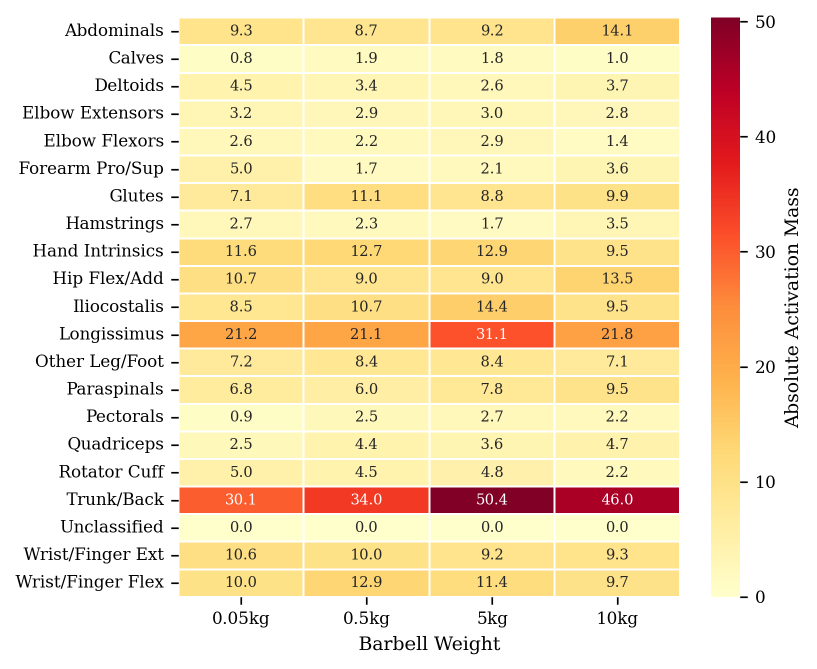}
    \caption{\textbf{Load-dependent recruitment in powerlift.} Muscle-family activation is computed by summing the time-averaged activations of its constituent actuators. Bars report this activation mass under different barbell weights; the metric does not represent instantaneous activation or agreement with human recordings.}
    \label{fig:powerweight}
\end{figure}

\begin{figure}[tb]
    \centering
    \includegraphics[width=1.0\linewidth]{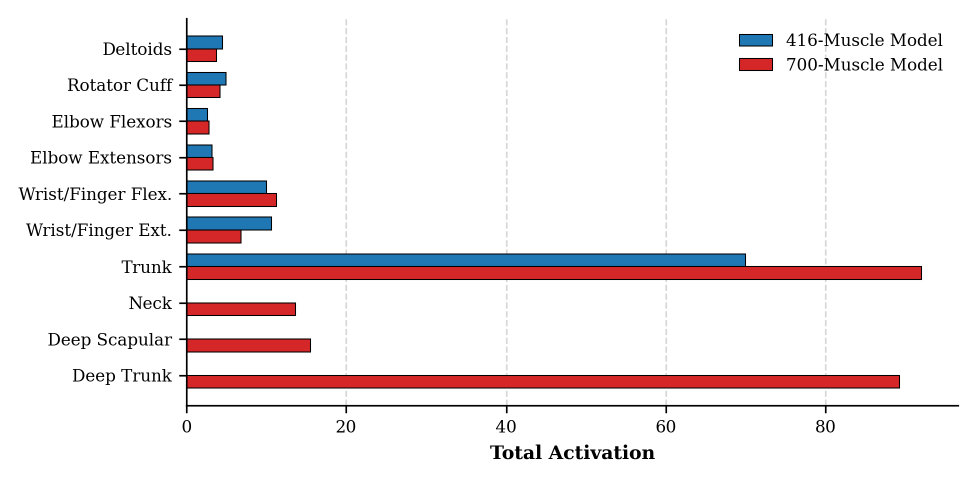}
    \caption{\textbf{Anatomical resolution and powerlift recruitment.} Activation mass for the 416- and 700-muscle models. The latter adds neck, deep-trunk, and deep-scapular groups, expanding available recruitment pathways; this does not establish greater human similarity.}
    \label{fig:416vs700}
    \vspace{-0.5cm}
\end{figure}

\emph{Task and anatomical context.}
Complementary diagnostics characterize movement fluctuations
and recruitment across task settings. Fig.~\ref{fig:joint_smoothness} reports
task-dependent joint-jerk levels with overlapping ranges
across task families, providing movement-level context
rather than a cross-task ranking of control quality.
In powerlift, muscle-family recruitment varies across loads:
trunk/back activation mass is 30.1, 50.4, and 46.0 at
0.05, 5, and 10~kg, respectively (Fig.~\ref{fig:powerweight}).
These profiles describe load-specific muscle use over the
recorded motions, rather than a uniform increase in
activation or agreement with human recordings. Fig.~\ref{fig:416vs700} extends this analysis to anatomical resolution.
The 700-muscle model reports additional neck, deep-trunk,
and deep-scapular recruitment groups, showing that the
available anatomical representation also determines what
muscle-use diagnostics can reveal. These model-specific
profiles do not establish greater human similarity.
Together, the analyses extend evaluation beyond task
completion to muscle-wise human-reference agreement and
task- and anatomy-specific descriptions of muscle use.

\section{Residual Adaptation Case Study}
\label{sec:residual_rl_policy}
\begin{figure}[t]
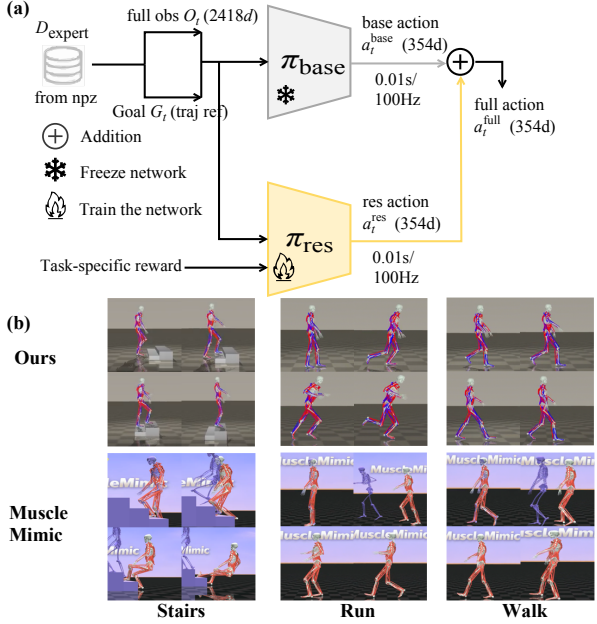

    \centering
    \mskResidualFigure
    \caption{\textbf{Residual architecture and motions.} Frozen prior plus trainable correction; blue skeletons are references. The original lower-branch label ``ref action'' denotes the residual correction in our equation. The 354d labels exclude hand actuation. Rewards and reference timing also change.}
    \label{fig:residual_main}
\end{figure}

T2 identifies stairs as an adaptation target; T3 asks separately whether adaptation recovers the task and improves EMG agreement. We use residual adaptation to demonstrate this diagnosis-to-validation transition, not as a standalone state-of-the-art method. Fig.~\ref{fig:residual_main} shows the controller and example motions. Specializing Eq.~\eqref{eq:unified_control}, a frozen MuscleMimic policy~\cite{li2026towards} supplies base actions, and a trainable 3-layer MLP supplies the bounded corrections defined in Eq.~\eqref{eq:residual_rl}.

Besides the residual correction, adaptation changes task objectives and timing: stairs relaxes vertical tracking and adapts reference pacing; walking uses phase-dependent residual authority and activation regularization; running adjusts flight-phase rewards and penalizes ground-reaction forces. A small L2 residual penalty ($c_{\mathrm{res}}=-0.02$) replaces the generic action-rate and activation-cost terms. Results therefore characterize the combined pipeline.

\textbf{Task-specific adaptation.}
For stairs, planar root error controls reference progression: below 0.4~m the reference advances normally; otherwise it slows by 50\%. Vertical root displacement is excluded from tracking, and progress targets the reference state 1~s ahead. Walking uses residual authority 0.6 in swing and 0.2 in stance and penalizes activation magnitude and temporal change. Running halves flight-phase tracking penalties, rewards toe-off, and softly penalizes vertical GRF above $2.5\times$ body weight. These choices permit larger swing corrections, preserve support, and avoid treating the residual study as an unchanged-objective ablation.

\begin{table}[!t]
\centering
\setlength{\tabcolsep}{3pt}
\caption{Residual adaptation results. MM: MuscleMimic. SR retains training
noise. EMG $r$ reports SoleusMed./Bic.Fem./Semitend. for walking, an
11-channel phase-aligned mean for running, and a 12-muscle phase-aligned mean
for stairs. Bold: best in each comparison.}
\label{tab:residual_results}
\begin{tabular}{@{}lccccc@{}}
\toprule
& \multicolumn{2}{c}{SR (\%) $\uparrow$}
& \multicolumn{3}{c}{EMG $r$ $\uparrow$} \\
\cmidrule(lr){2-3}
\cmidrule(lr){4-6}

Task
& MM
& \cellcolor{myblue}Residual
& MM
& DepRL
& \cellcolor{myblue}Residual \\

\midrule

Walking
& \textbf{100}
& \cellcolor{myblue}\textbf{100}
& 0.71/0.69/0.63
& --
& \cellcolor{myblue}\textbf{0.75/0.80/0.67} \\

Running
& 90
& \cellcolor{myblue}\textbf{100}
& 0.43
& 0.70
& \cellcolor{myblue}\textbf{0.77} \\

Stairs
& 0
& \cellcolor{myblue}\textbf{100}
& 0.51
& \textbf{0.56}
& \cellcolor{myblue}0.52 \\

\bottomrule
\end{tabular}
\end{table}

Tab.~\ref{tab:residual_results} summarizes task recovery and EMG agreement.
Walking retains 100\% SR while improving correlations for the three selected
muscles, and running improves both SR and mean EMG correlation. Stairs shows
the largest success gain but only a 0.01 correlation increase over MuscleMimic,
remaining below DepRL (0.56). Thus, task recovery does not imply uniformly
better human EMG agreement. Stair height rises from 0.94 to 1.21~m and
tracking error falls from 0.1968 to 0.1079~rad; running tracking error falls
from 0.2267 to 0.1338~rad and peak GRF from 6.40 to 6.07 times body weight.
Together, these results close the T1--T3 chain: broad evaluation motivates
task-specific diagnosis, which in turn motivates multi-metric validation
after adaptation.

\section{Limitations and Future Work}
\label{sec:limitations}

MSK-Bench is designed to compare control strategies under a unified task suite, rather than to claim that one method is universally superior. The reported results show the performance differences under the evaluated tasks and perturbations, but they do not imply statistical dominance or direct transfer to unseen embodiments.

The benchmark uses a fixed 416-muscle humanoid to reduce variations caused by different body designs. However, conclusions may not directly generalize to other anatomical models. Physiological evaluation is also limited by available human references: current EMG analysis covers three tasks using task-dependent sets of lower-limb signals---three selected muscles for walking, 11 independent EMG channels for running, and 12 matched muscles for stairs---and the correlation mainly reflects activation patterns rather than complete biological similarity.

Residual adaptation combines a reference prior with learned corrections, reward design, and task-specific adjustments. Therefore, its improvement should be interpreted as the effect of the complete adaptation pipeline rather than an isolated contribution from each component. Future extensions will include more dexterous manipulation tasks, longer-horizon behaviors, real-world validation, and broader physiological measurements (see App.~\ref{sec:app-limitations} for details).

\section{Conclusion}
\label{sec:conclusion}

In this work, we introduced \textbf{MSK-Bench}, a full-body musculoskeletal benchmark built on a 416-muscle humanoid, with 22 tasks spanning stabilization, locomotion, and contact-rich interaction. MSK-Bench provides a unified evaluation protocol covering task
performance, robustness, activation cost, joint smoothness, and EMG-envelope similarity. Our results show that embodiment-aware exploration and action-structured representations improve task coverage in the evaluated settings; motion-prior control performs strongly on several reference-compatible skills but fails on the tested stair task; and physiology-aware diagnostics reveal muscle-level differences not
captured by task success alone. We further use residual adaptation as a case study that combines a motion prior with task-specific corrections to recover task performance while separately assessing EMG agreement. MSK-Bench provides a testbed for physiologically grounded whole-body control.

\bibliographystyle{plainnat}
\bibliography{main}

@article{schumacher2022dep,
  title={Dep-rl: Embodied exploration for reinforcement learning in overactuated and musculoskeletal systems},
  author={Schumacher, Pierre and H{\"a}ufle, Daniel and B{\"u}chler, Dieter and Schmitt, Syn and Martius, Georg},
  journal={arXiv preprint arXiv:2206.00484},
  year={2022}
}

@article{schulman2017proximal,
  title={Proximal policy optimization algorithms},
  author={Schulman, John and Wolski, Filip and Dhariwal, Prafulla and Radford, Alec and Klimov, Oleg},
  journal={arXiv preprint arXiv:1707.06347},
  year={2017}
}

@inproceedings{haarnoja2018soft,
  title={Soft actor-critic: Off-policy maximum entropy deep reinforcement learning with a stochastic actor},
  author={Haarnoja, Tuomas and Zhou, Aurick and Abbeel, Pieter and Levine, Sergey},
  booktitle={International conference on machine learning},
  pages={1861--1870},
  year={2018},
  organization={Pmlr}
}

@article{wen2025hy,
  title={HY-Motion 1.0: Scaling Flow Matching Models for Text-To-Motion Generation},
  author={Wen, Yuxin and Shuai, Qing and Kang, Di and Li, Jing and Wen, Cheng and Qian, Yue and Jiao, Ningxin and Chen, Changhai and Chen, Weijie and Wang, Yiran and others},
  journal={arXiv preprint arXiv:2512.23464},
  year={2025}
}

@article{rempe2026kimodo,
  title={Kimodo: Scaling Controllable Human Motion Generation},
  author={Rempe, Davis and Petrovich, Mathis and Yuan, Ye and Zhang, Haotian and Peng, Xue Bin and Jiang, Yifeng and Wang, Tingwu and Iqbal, Umar and Minor, David and de Ruyter, Michael and others},
  journal={arXiv preprint arXiv:2603.15546},
  year={2026}
}

@article{li2026towards,
  title={Towards Embodied AI with MuscleMimic: Unlocking full-body musculoskeletal motor learning at scale},
  author={Li, Chengkun and Wang, Cheryl and Ziliotto, Bianca and Simos, Merkourios and Kovecses, Jozsef and Durandau, Guillaume and Mathis, Alexander},
  journal={arXiv preprint arXiv:2603.25544},
  year={2026}
}

@article{uhlrich2023opencap,
  title={OpenCap: Human movement dynamics from smartphone videos},
  author={Uhlrich, Scott D and Falisse, Antoine and Kidzi{\'n}ski, {\L}ukasz and Muccini, Julie and Ko, Michael and Chaudhari, Akshay S and Hicks, Jennifer L and Delp, Scott L},
  journal={PLoS computational biology},
  volume={19},
  number={10},
  pages={e1011462},
  year={2023},
  publisher={Public Library of Science San Francisco, CA USA}
}

@article{gilon2026opencap,
  title={OpenCap Monocular: 3D Human Kinematics and Musculoskeletal Dynamics from a Single Smartphone Video},
  author={Gilon, Selim and Miller, Emily Y and Uhlrich, Scott D},
  journal={arXiv preprint arXiv:2603.24733},
  year={2026}
}

@inproceedings{caggiano2024myochallenge,
  title={Myochallenge 2024: Physiological dexterity and agility in bionic humans},
  author={Caggiano, Vittorio and Durandau, Guillaume and Song, Seungmoon and Tan, Chun Kwang and Wang, Huiyi and Hodossy, Balint and Schumacher, Pierre and Gionfrida, Letizia and Sartori, Massimo and Kumar, Vikash},
  booktitle={NeurIPS 2024 Competition Track},
  year={2024}
}

@article{wang2026myochallenge,
  title={MyoChallenge 2024: A New Benchmark for Physiological Dexterity and Agility in Bionic Humans},
  author={Wang, Huiyi and Tan, Chun Kwang and Hodossy, Balint and Lyu, Shirui and Schumacher, Pierre and Heald, James and Biegun, Kai and Hromadka, Samo and Sahani, Maneesh and Park, Gunwoo and others},
  journal={Advances in Neural Information Processing Systems},
  volume={38},
  year={2026}
}

@incollection{caggiano2025myochallenge,
  title={MyoChallenge 2025: Towards Human Athletic Intelligence},
  author={Caggiano, Vittorio and Wang, Cheryl and Tan, Chun Kwang and Hodossy, Balint K and Lyu, Shirui and Sartori, Massimo and Song, Seungmoon and Gionfrida, Letizia and Durandau, Guillaume and Kumar, Vikash},
  booktitle={In NeurIPS 2024 Competition Track},
  year={2025}
}

@misc{wang2026myochallenge2025newbenchmark,
  title={MyoChallenge 2025: A New Benchmark for Human Athletic Intelligence}, 
  author={Cheryl Wang and Chun Kwang Tan and Balint K. Hodossy and Eric Lyu and Jun Guo and Wentao Zhao and Huaping Liu and Chengkun Li and Merkourios Simos and Bianca Ziliotto and Alexander Mathis and Siyuan Liu and Jiahao Chen and Shanlin Zhong and Bo Jiang and Ci Song and Yaoye Zhu and Chenhui Zuo and Yanan Sui and Mohamed Irfan Refai and Massimo Sartori and Guillaume Durandau and Vikash Kumar and Vittorio Caggiano},
  year={2026},
  eprint={2605.15650},
  archivePrefix={arXiv},
  primaryClass={cs.RO},
  url={https://arxiv.org/abs/2605.15650}, 
}

@article{sferrazza2024humanoidbench,
  title={Humanoidbench: Simulated humanoid benchmark for whole-body locomotion and manipulation},
  author={Sferrazza, Carmelo and Huang, Dun-Ming and Lin, Xingyu and Lee, Youngwoon and Abbeel, Pieter},
  journal={arXiv preprint arXiv:2403.10506},
  year={2024}
}

@inproceedings{yu2020meta,
  title={Meta-world: A benchmark and evaluation for multi-task and meta reinforcement learning},
  author={Yu, Tianhe and Quillen, Deirdre and He, Zhanpeng and Julian, Ryan and Hausman, Karol and Finn, Chelsea and Levine, Sergey},
  booktitle={Conference on robot learning},
  pages={1094--1100},
  year={2020},
  organization={PMLR}
}

@misc{james2020rlbench,
  title={RLBench: The Robot Learning Benchmark and Learning Environment. IEEE Robotics and Automation Letters 5, 2 (2020), 3019--3026},
  author={James, Stephen and Ma, Zicong and Arrojo, David Rovick and Davison, Andrew J},
  year={2020}
}

@article{kuang2025skillblender,
  title={Skillblender: Towards versatile humanoid whole-body loco-manipulation via skill blending},
  author={Kuang, Yuxuan and Geng, Haoran and Elhafsi, Amine and Do, Tan-Dzung and Abbeel, Pieter and Malik, Jitendra and Pavone, Marco and Wang, Yue},
  journal={arXiv preprint arXiv:2506.09366},
  year={2025}
}

@article{liu2024mimicking,
  title={Mimicking-bench: A benchmark for generalizable humanoid-scene interaction learning via human mimicking},
  author={Liu, Yun and Yang, Bowen and Zhong, Licheng and Wang, He and Yi, Li},
  journal={arXiv preprint arXiv:2412.17730},
  year={2024}
}

@article{he2024omnih2o,
  title={Omnih2o: Universal and dexterous human-to-humanoid whole-body teleoperation and learning},
  author={He, Tairan and Luo, Zhengyi and He, Xialin and Xiao, Wenli and Zhang, Chong and Zhang, Weinan and Kitani, Kris and Liu, Changliu and Shi, Guanya},
  journal={arXiv preprint arXiv:2406.08858},
  year={2024}
}

@article{al2023locomujoco,
  title={Locomujoco: A comprehensive imitation learning benchmark for locomotion},
  author={Al-Hafez, Firas and Zhao, Guoping and Peters, Jan and Tateo, Davide},
  journal={arXiv preprint arXiv:2311.02496},
  year={2023}
}

@article{delp2007opensim,
  title={OpenSim: open-source software to create and analyze dynamic simulations of movement},
  author={Delp, Scott L and Anderson, Frank C and Arnold, Allison S and Loan, Peter and Habib, Ayman and John, Chand T and Guendelman, Eran and Thelen, Darryl G},
  journal={IEEE transactions on biomedical engineering},
  volume={54},
  number={11},
  pages={1940--1950},
  year={2007},
  publisher={IEEE}
}

@article{seth2018opensim,
  title={OpenSim: Simulating musculoskeletal dynamics and neuromuscular control to study human and animal movement},
  author={Seth, Ajay and Hicks, Jennifer L and Uchida, Thomas K and Habib, Ayman and Dembia, Christopher L and Dunne, James J and Ong, Carmichael F and DeMers, Matthew S and Rajagopal, Apoorva and Millard, Matthew and others},
  journal={PLoS computational biology},
  volume={14},
  number={7},
  pages={e1006223},
  year={2018},
  publisher={Public Library of Science San Francisco, CA USA}
}

@article{seth2011opensim,
  title={OpenSim: a musculoskeletal modeling and simulation framework for in silico investigations and exchange},
  author={Seth, Ajay and Sherman, Michael and Reinbolt, Jeffrey A and Delp, Scott L},
  journal={Procedia Iutam},
  volume={2},
  pages={212--232},
  year={2011},
  publisher={Elsevier}
}

@article{caggiano2022myosuite,
  title={MyoSuite--A contact-rich simulation suite for musculoskeletal motor control},
  author={Caggiano, Vittorio and Wang, Huawei and Durandau, Guillaume and Sartori, Massimo and Kumar, Vikash},
  journal={arXiv preprint arXiv:2205.13600},
  year={2022}
}

@inproceedings{zuo2024self,
  title={Self model for embodied intelligence: Modeling full-body human musculoskeletal system and locomotion control with hierarchical low-dimensional representation},
  author={Zuo, Chenhui and He, Kaibo and Shao, Jing and Sui, Yanan},
  booktitle={2024 IEEE International Conference on Robotics and Automation (ICRA)},
  pages={13062--13069},
  year={2024},
  organization={IEEE}
}

@article{wei2026scaling,
  title={Scaling Whole-Body Human Musculoskeletal Behavior Emulation for Specificity and Diversity},
  author={Wei, Yunyue and Zuo, Chenhui and Zhuang, Shanning and Gong, Haixin and Liu, Yaming and Sui, Yanan},
  journal={arXiv preprint arXiv:2603.29332},
  year={2026}
}

@article{fu2020d4rl,
  title={D4rl: Datasets for deep data-driven reinforcement learning},
  author={Fu, Justin and Kumar, Aviral and Nachum, Ofir and Tucker, George and Levine, Sergey},
  journal={arXiv preprint arXiv:2004.07219},
  year={2020}
}

@inproceedings{todorov2012mujoco,
  title={Mujoco: A physics engine for model-based control},
  author={Todorov, Emanuel and Erez, Tom and Tassa, Yuval},
  booktitle={2012 IEEE/RSJ international conference on intelligent robots and systems},
  pages={5026--5033},
  year={2012},
  organization={IEEE}
}

@article{tassa2018deepmind,
  title={Deepmind control suite},
  author={Tassa, Yuval and Doron, Yotam and Muldal, Alistair and Erez, Tom and Li, Yazhe and Casas, Diego de Las and Budden, David and Abdolmaleki, Abbas and Merel, Josh and Lefrancq, Andrew and others},
  journal={arXiv preprint arXiv:1801.00690},
  year={2018}
}

@article{zhu2020robosuite,
  title={robosuite: A modular simulation framework and benchmark for robot learning},
  author={Zhu, Yuke and Wong, Josiah and Mandlekar, Ajay and Mart{\'\i}n-Mart{\'\i}n, Roberto and Joshi, Abhishek and Lin, Kevin and Maddukuri, Abhiram and Nasiriany, Soroush and Zhu, Yifeng},
  journal={arXiv preprint arXiv:2009.12293},
  year={2020}
}

@article{gu2023maniskill2,
  title={Maniskill2: A unified benchmark for generalizable manipulation skills},
  author={Gu, Jiayuan and Xiang, Fanbo and Li, Xuanlin and Ling, Zhan and Liu, Xiqiang and Mu, Tongzhou and Tang, Yihe and Tao, Stone and Wei, Xinyue and Yao, Yunchao and others},
  journal={arXiv preprint arXiv:2302.04659},
  year={2023}
}

@article{tao2024maniskill3,
  title={Maniskill3: Gpu parallelized robotics simulation and rendering for generalizable embodied ai},
  author={Tao, Stone and Xiang, Fanbo and Shukla, Arth and Qin, Yuzhe and Hinrichsen, Xander and Yuan, Xiaodi and Bao, Chen and Lin, Xinsong and Liu, Yulin and Chan, Tse-kai and others},
  journal={arXiv preprint arXiv:2410.00425},
  year={2024}
}

@article{mees2022calvin,
  title={Calvin: A benchmark for language-conditioned policy learning for long-horizon robot manipulation tasks},
  author={Mees, Oier and Hermann, Lukas and Rosete-Beas, Erick and Burgard, Wolfram},
  journal={IEEE Robotics and Automation Letters},
  volume={7},
  number={3},
  pages={7327--7334},
  year={2022},
  publisher={IEEE}
}

@article{JiangYunfan2023VGRM,
  author = {Jiang, Yunfan and Gupta, Agrim and Zhang, Zichen and Wang, Guanzhi and Dou, Yongqiang and Chen, Yanjun and Li, Fei-Fei and Anandkumar, Anima and Zhu, Yuke and Fan, Linxi},
  address = {Ithaca},
  copyright = {2023. This work is published under http://creativecommons.org/licenses/by/4.0/ (the “License”). Notwithstanding the ProQuest Terms and Conditions, you may use this content in accordance with the terms of the License.},
  issn = {2331-8422},
  journal = {arXiv.org},
  language = {eng},
  publisher = {Cornell University Library, arXiv.org},
  title = {VIMA: General Robot Manipulation with Multimodal Prompts},
  year = {2023},
}

@article{liu2023libero,
  title={Libero: Benchmarking knowledge transfer for lifelong robot learning},
  author={Liu, Bo and Zhu, Yifeng and Gao, Chongkai and Feng, Yihao and Liu, Qiang and Zhu, Yuke and Stone, Peter},
  journal={Advances in Neural Information Processing Systems},
  volume={36},
  pages={44776--44791},
  year={2023}
}

@article{nasiriany2024robocasa,
  title={Robocasa: Large-scale simulation of everyday tasks for generalist robots},
  author={Nasiriany, Soroush and Maddukuri, Abhiram and Zhang, Lance and Parikh, Adeet and Lo, Aaron and Joshi, Abhishek and Mandlekar, Ajay and Zhu, Yuke},
  journal={arXiv preprint arXiv:2406.02523},
  year={2024}
}

@inproceedings{mu2025robotwin,
  title={Robotwin: Dual-arm robot benchmark with generative digital twins},
  author={Mu, Yao and Chen, Tianxing and Chen, Zanxin and Peng, Shijia and Lan, Zhiqian and Gao, Zeyu and Liang, Zhixuan and Yu, Qiaojun and Zou, Yude and Xu, Mingkun and others},
  booktitle={Proceedings of the computer vision and pattern recognition conference},
  pages={27649--27660},
  year={2025}
}

@article{kumar2023robohive,
  title={Robohive: A unified framework for robot learning},
  author={Kumar, Vikash and Shah, Rutav and Zhou, Gaoyue and Moens, Vincent and Caggiano, Vittorio and Gupta, Abhishek and Rajeswaran, Aravind},
  journal={Advances in Neural Information Processing Systems},
  volume={36},
  pages={44323--44340},
  year={2023}
}

@article{peng2018deepmimic,
  title={Deepmimic: Example-guided deep reinforcement learning of physics-based character skills},
  author={Peng, Xue Bin and Abbeel, Pieter and Levine, Sergey and Van de Panne, Michiel},
  journal={ACM Transactions On Graphics (TOG)},
  volume={37},
  number={4},
  pages={1--14},
  year={2018},
  publisher={ACM New York, NY, USA}
}

@incollection{kidzinski2018learning1,
  title={Learning to run challenge: Synthesizing physiologically accurate motion using deep reinforcement learning},
  author={Kidzi{\'n}ski, {\L}ukasz and Mohanty, Sharada P and Ong, Carmichael F and Hicks, Jennifer L and Carroll, Sean F and Levine, Sergey and Salath{\'e}, Marcel and Delp, Scott L},
  booktitle={The NIPS'17 Competition: Building Intelligent Systems},
  pages={101--120},
  year={2018},
  publisher={Springer}
}

@incollection{kidzinski2018learning2,
  title={Learning to run challenge solutions: Adapting reinforcement learning methods for neuromusculoskeletal environments},
  author={Kidzi{\'n}ski, {\L}ukasz and Mohanty, Sharada Prasanna and Ong, Carmichael F and Huang, Zhewei and Zhou, Shuchang and Pechenko, Anton and Stelmaszczyk, Adam and Jarosik, Piotr and Pavlov, Mikhail and Kolesnikov, Sergey and others},
  booktitle={The NIPS'17 Competition: Building Intelligent Systems},
  pages={121--153},
  year={2018},
  publisher={Springer}
}

@article{xu2026music,
  title={MUSIC: Learning Muscle-Driven Dexterous Hand Control},
  author={Xu, Pei and Ye, Yufei and Sun, Shuchun and Ding, Yu and Schumann, Elizabeth and Liu, C Karen},
  journal={arXiv preprint arXiv:2604.23886},
  year={2026}
}

@article{song2021deep,
  title={Deep reinforcement learning for modeling human locomotion control in neuromechanical simulation},
  author={Song, Seungmoon and Kidzi{\'n}ski, {\L}ukasz and Peng, Xue Bin and Ong, Carmichael and Hicks, Jennifer and Levine, Sergey and Atkeson, Christopher G and Delp, Scott L},
  journal={Journal of neuroengineering and rehabilitation},
  volume={18},
  number={1},
  pages={126},
  year={2021},
  publisher={Springer}
}

@inproceedings{lin2021softgym,
  title={Softgym: Benchmarking deep reinforcement learning for deformable object manipulation},
  author={Lin, Xingyu and Wang, Yufei and Olkin, Jake and Held, David},
  booktitle={Conference on Robot Learning},
  pages={432--448},
  year={2021},
  organization={PMLR}
}

@inproceedings{walke2023bridgedata,
  title={Bridgedata v2: A dataset for robot learning at scale},
  author={Walke, Homer Rich and Black, Kevin and Zhao, Tony Z and Vuong, Quan and Zheng, Chongyi and Hansen-Estruch, Philippe and He, Andre Wang and Myers, Vivek and Kim, Moo Jin and Du, Max and others},
  booktitle={Conference on Robot Learning},
  pages={1723--1736},
  year={2023},
  organization={PMLR}
}

@article{nasiriany2026robocasa365,
  title={Robocasa365: A large-scale simulation framework for training and benchmarking generalist robots},
  author={Nasiriany, Soroush and Nasiriany, Sepehr and Maddukuri, Abhiram and Zhu, Yuke},
  journal={arXiv preprint arXiv:2603.04356},
  year={2026}
}

@misc{zhang2026lhmhumanoidlearningunifiedpolicy,
  title={LHM-Humanoid: Learning a Unified Policy for Long-Horizon Humanoid Whole-Body Loco-Manipulation in Diverse Messy Environments}, 
  author={Haozhuo Zhang and Jingkai Sun and Michele Caprio and Jian Tang and Shanghang Zhang and Qiang Zhang and Wei Pan},
  year={2026},
  eprint={2508.16943},
  archivePrefix={arXiv},
  primaryClass={cs.RO},
  url={https://arxiv.org/abs/2508.16943}, 
}

@article{brockman2016openai,
  title={Openai gym},
  author={Brockman, Greg and Cheung, Vicki and Pettersson, Ludwig and Schneider, Jonas and Schulman, John and Tang, Jie and Zaremba, Wojciech},
  journal={arXiv preprint arXiv:1606.01540},
  year={2016}
}

@inproceedings{caggiano2023myodex,
  title={Myodex: a generalizable prior for dexterous manipulation},
  author={Caggiano, Vittorio and Dasari, Sudeep and Kumar, Vikash},
  booktitle={International Conference on Machine Learning},
  pages={3327--3346},
  year={2023},
  organization={PMLR}
}

@article{simos2025reinforcement,
  title={Reinforcement learning-based motion imitation for physiologically plausible musculoskeletal motor control},
  author={Simos, Merkourios and Silvio Chiappa, Alberto and Mathis, Alexander},
  journal={arXiv e-prints},
  pages={arXiv--2503},
  year={2025}
}

@inproceedings{johannink2019residual,
  title={Residual reinforcement learning for robot control},
  author={Johannink, Tobias and Bahl, Shikhar and Nair, Ashvin and Luo, Jianlan and Kumar, Avinash and Loskyll, Matthias and Ojea, Juan Aparicio and Solowjow, Eugen and Levine, Sergey},
  booktitle={2019 international conference on robotics and automation (ICRA)},
  pages={6023--6029},
  year={2019},
  organization={IEEE}
}

@article{yan2024ctrl,
  title={I-ctrl: Imitation to control humanoid robots through constrained reinforcement learning},
  author={Yan, Yashuai and Mascaro, Esteve Valls and Egle, Tobias and Lee, Dongheui},
  journal={arXiv preprint arXiv:2405.08726},
  year={2024}
}

@article{han2026muscle,
  title={Muscle Synergy-Guided Reinforcement Learning for Embodied Musculoskeletal Motion Skill Learning},
  author={Han, Lijun and Cheng, Long and Ma, Muyuan and Yang, Feiyang and Li, Houcheng},
  journal={IEEE Transactions on Biomedical Engineering},
  year={2026},
  publisher={IEEE}
}

@article{berg2024sar,
  title={Sar: Generalization of physiological agility and dexterity via synergistic action representation},
  author={Berg, Cameron and Caggiano, Vittorio and Kumar, Vikash},
  journal={Autonomous Robots},
  volume={48},
  number={8},
  pages={28},
  year={2024},
  publisher={Springer}
}

@article{silver2018residual,
  title={Residual policy learning},
  author={Silver, Tom and Allen, Kelsey and Tenenbaum, Josh and Kaelbling, Leslie},
  journal={arXiv preprint arXiv:1812.06298},
  year={2018}
}

@article{lin2026peel,
  title={How to Peel with a Knife: Aligning Fine-Grained Manipulation with Human Preference},
  author={Lin, Toru and Deng, Shuying and Yin, Zhao-Heng and Abbeel, Pieter and Malik, Jitendra},
  journal={arXiv preprint arXiv:2603.03280},
  year={2026}
}

@article{xu2026compliant,
  title={Compliant residual dagger: Improving real-world contact-rich manipulation with human corrections},
  author={Xu, Xiaomeng and Hou, Yifan and Liu, Zeyi and Song, Shuran},
  journal={Advances in Neural Information Processing Systems},
  volume={38},
  pages={139559--139581},
  year={2026}
}

@article{ankile2025residual,
  title={Residual off-policy rl for finetuning behavior cloning policies},
  author={Ankile, Lars and Jiang, Zhenyu and Duan, Rocky and Shi, Guanya and Abbeel, Pieter and Nagabandi, Anusha},
  journal={arXiv preprint arXiv:2509.19301},
  year={2025}
}

@article{zhao2025resmimic,
  title={Resmimic: From general motion tracking to humanoid whole-body loco-manipulation via residual learning},
  author={Zhao, Siheng and Ze, Yanjie and Wang, Yue and Liu, C Karen and Abbeel, Pieter and Shi, Guanya and Duan, Rocky},
  journal={arXiv preprint arXiv:2510.05070},
  year={2025}
}

@article{sun2025robotdancing,
  title={RobotDancing: Residual-Action Reinforcement Learning Enables Robust Long-Horizon Humanoid Motion Tracking},
  author={Sun, Zhenguo and Peng, Yibo and Meng, Yuan and Li, Xukun and Huang, Bo-Sheng and Bing, Zhenshan and Wang, Xinlong and Knoll, Alois},
  journal={arXiv preprint arXiv:2509.20717},
  year={2025}
}

@article{yu2023language,
  title={Language to rewards for robotic skill synthesis},
  author={Yu, Wenhao and Gileadi, Nimrod and Fu, Chuyuan and Kirmani, Sean and Lee, Kuang-Huei and Arenas, Montse Gonzalez and Chiang, Hao-Tien Lewis and Erez, Tom and Hasenclever, Leonard and Humplik, Jan and others},
  journal={arXiv preprint arXiv:2306.08647},
  year={2023}
}

@inproceedings{xie2024text2reward,
  title={Text2reward: Reward shaping with language models for reinforcement learning},
  author={Xie, Tianbao and Zhao, Siheng and Wu, Chen and Liu, Yitao and Luo, Qian and Zhong, Victor and Yang, Yanchao and Yu, Tao},
  booktitle={International Conference on Learning Representations},
  volume={2024},
  pages={35663--35699},
  year={2024}
}

@inproceedings{ma2024eureka,
  title={Eureka: Human-level reward design via coding large language models},
  author={Ma, Yecheng Jason and Liang, William and Wang, Guanzhi and Huang, De-An and Bastani, Osbert and Jayaraman, Dinesh and Zhu, Yuke and Fan, Jim and others},
  booktitle={International conference on learning Representations},
  volume={2024},
  pages={26516--26560},
  year={2024}
}

@inproceedings{wang2024gensim,
  title={Gensim: Generating robotic simulation tasks via large language models},
  author={Wang, Lirui and Ling, Yiyang and Yuan, Zhecheng and Shridhar, Mohit and Bao, Chen and Qin, Yuzhe and Wang, Bailin and Xu, Huazhe and Wang, Xiaolong},
  booktitle={International Conference on Learning Representations},
  volume={2024},
  pages={4890--4924},
  year={2024}
}

@article{alakuijala2021residual,
  title={Residual reinforcement learning from demonstrations},
  author={Alakuijala, Minttu and Dulac-Arnold, Gabriel and Mairal, Julien and Ponce, Jean and Schmid, Cordelia},
  journal={arXiv preprint arXiv:2106.08050},
  year={2021}
}

@article{peng2021amp,
  title={Amp: Adversarial motion priors for stylized physics-based character control},
  author={Peng, Xue Bin and Ma, Ze and Abbeel, Pieter and Levine, Sergey and Kanazawa, Angjoo},
  journal={ACM Transactions on Graphics (ToG)},
  volume={40},
  number={4},
  pages={1--20},
  year={2021},
  publisher={ACM New York, NY, USA}
}

@article{wang2026motionbricks,
  title={MotionBricks: Scalable Real-Time Motions with Modular Latent Generative Model and Smart Primitives},
  author={Wang, Tingwu and Dionne, Olivier and De Ruyter, Michael and Minor, David and Rempe, Davis and Zhao, Kaifeng and Petrovich, Mathis and Yuan, Ye and Li, Chenran and Luo, Zhengyi and others},
  journal={arXiv preprint arXiv:2604.24833},
  year={2026}
}

@article{peng2022ase,
  title={Ase: Large-scale reusable adversarial skill embeddings for physically simulated characters},
  author={Peng, Xue Bin and Guo, Yunrong and Halper, Lina and Levine, Sergey and Fidler, Sanja},
  journal={ACM Transactions On Graphics (TOG)},
  volume={41},
  number={4},
  pages={1--17},
  year={2022},
  publisher={ACM New York, NY, USA}
}

@inproceedings{tessler2023calm,
  title={Calm: Conditional adversarial latent models for directable virtual characters},
  author={Tessler, Chen and Kasten, Yoni and Guo, Yunrong and Mannor, Shie and Chechik, Gal and Peng, Xue Bin},
  booktitle={ACM SIGGRAPH 2023 conference proceedings},
  pages={1--9},
  year={2023}
}

@inproceedings{rana2023residual,
  title={Residual skill policies: Learning an adaptable skill-based action space for reinforcement learning for robotics},
  author={Rana, Krishan and Xu, Ming and Tidd, Brendan and Milford, Michael and S{\"u}nderhauf, Niko},
  booktitle={Conference on Robot Learning},
  pages={2095--2104},
  year={2023},
  organization={PMLR}
}

@article{damsgaard2006anybody,
  title={Analysis of musculoskeletal systems in the AnyBody Modeling System},
  author={Damsgaard, Michael and Rasmussen, John and Christensen, S{\o}ren T{\o}rholm and Surma, Egidijus and De Zee, Mark},
  journal={Simulation Modelling Practice and Theory},
  volume={14},
  number={8},
  pages={1100--1111},
  year={2006},
  publisher={Elsevier}
}

@inproceedings{wang2022myosim,
  title={MyoSim: Fast and physiologically realistic MuJoCo models for musculoskeletal and exoskeletal studies},
  author={Wang, Huawei and Caggiano, Vittorio and Durandau, Guillaume and Sartori, Massimo and Kumar, Vikash},
  booktitle={2022 International Conference on Robotics and Automation (ICRA)},
  pages={8104--8111},
  year={2022},
  organization={IEEE}
}

@article{cotton2025kintwin,
  title={KinTwin: Imitation Learning with Torque and Muscle Driven Biomechanical Models Enables Precise Replication of Able-Bodied and Impaired Movement from Markerless Motion Capture},
  author={Cotton, R James},
  journal={arXiv preprint arXiv:2505.13436},
  year={2025}
}

@article{wei2025motion,
  title={Motion control of high-dimensional musculoskeletal systems with hierarchical model-based planning},
  author={Wei, Yunyue and Zhuang, Shanning and Zhuang, Vincent and Sui, Yanan},
  journal={arXiv preprint arXiv:2505.08238},
  year={2025}
}

@article{he2024dynsyn,
  title={Dynsyn: Dynamical synergistic representation for efficient learning and control in overactuated embodied systems},
  author={He, Kaibo and Zuo, Chenhui and Ma, Chengtian and Sui, Yanan},
  journal={arXiv preprint arXiv:2407.11472},
  year={2024}
}

@inproceedings{feng2023musclevae,
  title={Musclevae: Model-based controllers of muscle-actuated characters},
  author={Feng, Yusen and Xu, Xiyan and Liu, Libin},
  booktitle={SIGGRAPH Asia 2023 Conference Papers},
  pages={1--11},
  year={2023}
}

@article{mu2025smp,
  title={SMP: Reusable Score-Matching Motion Priors for Physics-Based Character Control},
  author={Mu, Yuxuan and Zhang, Ziyu and Shi, Yi and Matsumoto, Minami and Imamura, Kotaro and Tevet, Guy and Guo, Chuan and Taylor, Michael and Shu, Chang and Xi, Pengcheng and others},
  journal={arXiv preprint arXiv:2512.03028},
  year={2025}
}

@article{kawaharazuka2026characteristics,
  title={Characteristics, Management, and Utilization of Muscles in Musculoskeletal Humanoids: Empirical Study on Kengoro and Musashi},
  author={Kawaharazuka, Kento and Okada, Kei and Inaba, Masayuki},
  journal={Advanced Intelligent Systems},
  pages={e202500741},
  year={2026},
  publisher={Wiley Online Library}
}

@incollection{kidzinski2019artificial,
  title={Artificial intelligence for prosthetics: Challenge solutions},
  author={Kidzi{\'n}ski, {\L}ukasz and Ong, Carmichael and Mohanty, Sharada Prasanna and Hicks, Jennifer and Carroll, Sean and Zhou, Bo and Zeng, Hongsheng and Wang, Fan and Lian, Rongzhong and Tian, Hao and others},
  booktitle={The NeurIPS'18 Competition: From Machine Learning to Intelligent Conversations},
  pages={69--128},
  year={2019},
  publisher={Springer}
}

@article{radosavovic2024real,
  title={Real-world humanoid locomotion with reinforcement learning},
  author={Radosavovic, Ilija and Xiao, Tete and Zhang, Bike and Darrell, Trevor and Malik, Jitendra and Sreenath, Koushil},
  journal={Science Robotics},
  volume={9},
  number={89},
  pages={eadi9579},
  year={2024},
  publisher={American Association for the Advancement of Science}
}

@article{selinger2015humans,
  title={Humans can continuously optimize energetic cost during walking},
  author={Selinger, Jessica C and O’Connor, Shawn M and Wong, Jeremy D and Donelan, J Maxwell},
  journal={Current Biology},
  volume={25},
  number={18},
  pages={2452--2456},
  year={2015},
  publisher={Elsevier}
}

@article{michaud2021fair,
  title={A fair and EMG-validated comparison of recruitment criteria, musculotendon models and muscle coordination strategies, for the inverse-dynamics based optimization of muscle forces during gait},
  author={Michaud, Florian and Lamas, Mario and Lugr{\'\i}s, Urbano and Cuadrado, Javier},
  journal={Journal of neuroengineering and rehabilitation},
  volume={18},
  number={1},
  pages={17},
  year={2021},
  publisher={Springer}
}

@article{flash1985coordination,
  title={The coordination of arm movements: an experimentally confirmed mathematical model},
  author={Flash, Tamar and Hogan, Neville},
  journal={Journal of neuroscience},
  volume={5},
  number={7},
  pages={1688--1703},
  year={1985},
  publisher={Society for Neuroscience}
}

@article{balasubramanian2015analysis,
  title={On the analysis of movement smoothness},
  author={Balasubramanian, Sivakumar and Melendez-Calderon, Alejandro and Roby-Brami, Agnes and Burdet, Etienne},
  journal={Journal of neuroengineering and rehabilitation},
  volume={12},
  number={1},
  pages={112},
  year={2015},
  publisher={Springer}
}

@inproceedings{duan2016benchmarking,
  title={Benchmarking deep reinforcement learning for continuous control},
  author={Duan, Yan and Chen, Xi and Houthooft, Rein and Schulman, John and Abbeel, Pieter},
  booktitle={International conference on machine learning},
  pages={1329--1338},
  year={2016},
  organization={PMLR}
}

@article{hollenstein2022action,
  title={Action noise in off-policy deep reinforcement learning: Impact on exploration and performance},
  author={Hollenstein, Jakob and Auddy, Sayantan and Saveriano, Matteo and Renaudo, Erwan and Piater, Justus},
  journal={arXiv preprint arXiv:2206.03787},
  year={2022}
}

@inproceedings{desai2020stochastic,
  title={Stochastic grounded action transformation for robot learning in simulation},
  author={Desai, Siddharth and Karnan, Haresh and Hanna, Josiah P and Warnell, Garrett and Stone, Peter},
  booktitle={2020 IEEE/RSJ International Conference on Intelligent Robots and Systems (IROS)},
  pages={6106--6111},
  year={2020},
  organization={IEEE}
}

@inproceedings{tobin2017domain,
  title={Domain randomization for transferring deep neural networks from simulation to the real world},
  author={Tobin, Josh and Fong, Rachel and Ray, Alex and Schneider, Jonas and Zaremba, Wojciech and Abbeel, Pieter},
  booktitle={2017 IEEE/RSJ international conference on intelligent robots and systems (IROS)},
  pages={23--30},
  year={2017},
  organization={IEEE}
}

@inproceedings{peng2018sim,
  title={Sim-to-real transfer of robotic control with dynamics randomization},
  author={Peng, Xue Bin and Andrychowicz, Marcin and Zaremba, Wojciech and Abbeel, Pieter},
  booktitle={2018 IEEE international conference on robotics and automation (ICRA)},
  pages={3803--3810},
  year={2018},
  organization={IEEE}
}

@article{tan2018sim,
  title={Sim-to-real: Learning agile locomotion for quadruped robots},
  author={Tan, Jie and Zhang, Tingnan and Coumans, Erwin and Iscen, Atil and Bai, Yunfei and Hafner, Danijar and Bohez, Steven and Vanhoucke, Vincent},
  journal={arXiv preprint arXiv:1804.10332},
  year={2018}
}

@article{akkaya2019solving,
  title={Solving rubik's cube with a robot hand},
  author={Akkaya, Ilge and Andrychowicz, Marcin and Chociej, Maciek and Litwin, Mateusz and McGrew, Bob and Petron, Arthur and Paino, Alex and Plappert, Matthias and Powell, Glenn and Ribas, Raphael and others},
  journal={arXiv preprint arXiv:1910.07113},
  year={2019}
}

@article{nimbarte2014risk,
  title={Risk of neck musculoskeletal disorders among males and females in lifting exertions},
  author={Nimbarte, Ashish D},
  journal={International Journal of Industrial Ergonomics},
  volume={44},
  number={2},
  pages={253--259},
  year={2014},
  publisher={Elsevier}
}

@article{latash2018muscle,
  title={Muscle coactivation: definitions, mechanisms, and functions},
  author={Latash, Mark L},
  journal={Journal of neurophysiology},
  volume={120},
  number={1},
  pages={88--104},
  year={2018},
  publisher={American Physiological Society Bethesda, MD}
}

@article{boo2025comprehensive,
  title={Comprehensive human locomotion and electromyography dataset: Gait120},
  author={Boo, Junyo and Seo, Dongwook and Kim, Minseung and Koo, Seungbum},
  journal={Scientific Data},
  volume={12},
  number={1},
  pages={1023},
  year={2025},
  publisher={Nature Publishing Group UK London}
}

@misc{deepseekai2026deepseekv4,
      title={DeepSeek-V4: Towards Highly Efficient Million-Token Context Intelligence},
      author={DeepSeek-AI},
      year={2026},
}

@article{d2003combinations,
  title={Combinations of muscle synergies in the construction of a natural motor behavior},
  author={d'Avella, Andrea and Saltiel, Philippe and Bizzi, Emilio},
  journal={Nature neuroscience},
  volume={6},
  number={3},
  pages={300--308},
  year={2003},
  publisher={Nature Publishing Group US New York}
}

@article{calatayud2015core,
  title={Core muscle activity during the clean and jerk lift with barbell versus sandbags and water bags},
  author={Calatayud, Joaquin and Colado, Juan C and Martin, Fernando and Casa{\~n}a, Jos{\'e} and Jakobsen, Markus D and Andersen, Lars L},
  journal={International journal of sports physical therapy},
  volume={10},
  number={6},
  pages={803},
  year={2015}
}

@article{coratella2022front,
  title={Front vs back and barbell vs machine overhead press: an electromyographic analysis and implications for resistance training},
  author={Coratella, Giuseppe and Tornatore, Gianpaolo and Longo, Stefano and Esposito, Fabio and C{\`e}, Emiliano},
  journal={Frontiers in physiology},
  volume={13},
  pages={825880},
  year={2022},
  publisher={Frontiers Media SA}
}

@article{geisler2023effects,
  title={Effects of expertise on muscle activity during the hang power clean and hang power snatch compared to snatch and clean pulls--an explorative analysis},
  author={Geisler, Stephan and Havers, Tim and Isenmann, Eduard and Schulze, Jonas and Lourens, Leonie K and Nowak, Jannik and Held, Steffen and Haff, G Gregory},
  journal={Journal of Sports Science \& Medicine},
  volume={22},
  number={4},
  pages={778},
  year={2023}
}

@article{wei2026scalable,
  title={Scalable Exploration for High-Dimensional Continuous Control via Value-Guided Flow},
  author={Wei, Yunyue and Zuo, Chenhui and Sui, Yanan},
  journal={arXiv preprint arXiv:2601.19707},
  year={2026}
}

@article{valero2015exploring,
  title={Exploring the high-dimensional structure of muscle redundancy via subject-specific and generic musculoskeletal models},
  author={Valero-Cuevas, Francisco J and Cohn, BA and Yngvason, HF and Lawrence, Emily L},
  journal={Journal of biomechanics},
  volume={48},
  number={11},
  pages={2887--2896},
  year={2015},
  publisher={Elsevier}
}

@article{jonsson2004one,
  title={One-leg stance in healthy young and elderly adults: a measure of postural steadiness?},
  author={Jonsson, Erika and Seiger, {\AA}ke and Hirschfeld, Helga},
  journal={Clinical biomechanics},
  volume={19},
  number={7},
  pages={688--694},
  year={2004},
  publisher={Elsevier}
}

@article{bressel2009effect,
  title={Effect of instruction, surface stability, and load intensity on trunk muscle activity},
  author={Bressel, Eadric and Willardson, Jeffrey M and Thompson, Brennan and Fontana, Fabio E},
  journal={Journal of Electromyography and Kinesiology},
  volume={19},
  number={6},
  pages={e500--e504},
  year={2009},
  publisher={Elsevier}
}

@article{kuo2010dynamic,
  title={Dynamic principles of gait and their clinical implications},
  author={Kuo, Arthur D and Donelan, J Maxwell},
  journal={Physical therapy},
  volume={90},
  number={2},
  pages={157--174},
  year={2010},
  publisher={Oxford University Press}
}

@article{neptune2019dynamic,
  title={Dynamic balance during human movement: measurement and control mechanisms},
  author={Neptune, Richard R and Vistamehr, Arian},
  journal={Journal of Biomechanical Engineering},
  volume={141},
  number={7},
  pages={070801},
  year={2019},
  publisher={American Society of Mechanical Engineers}
}

@article{zajac2003biomechanics,
  title={Biomechanics and muscle coordination of human walking: part II: lessons from dynamical simulations and clinical implications},
  author={Zajac, Felix E and Neptune, Richard R and Kautz, Steven A},
  journal={Gait \& posture},
  volume={17},
  number={1},
  pages={1--17},
  year={2003},
  publisher={Elsevier}
}

@article{newell2020fundamental,
  title={What are fundamental motor skills and what is fundamental about them?},
  author={Newell, Karl M},
  journal={Journal of Motor Learning and Development},
  volume={8},
  number={2},
  pages={280--314},
  year={2020},
  publisher={Human Kinetics}
}

@article{stokes2011abdominal,
  title={Abdominal muscle activation increases lumbar spinal stability: analysis of contributions of different muscle groups},
  author={Stokes, Ian AF and Gardner-Morse, Mack G and Henry, Sharon M},
  journal={Clinical biomechanics},
  volume={26},
  number={8},
  pages={797--803},
  year={2011},
  publisher={Elsevier}
}

@article{liu2008muscle,
  title={Muscle contributions to support and progression over a range of walking speeds},
  author={Liu, May Q and Anderson, Frank C and Schwartz, Michael H and Delp, Scott L},
  journal={Journal of biomechanics},
  volume={41},
  number={15},
  pages={3243--3252},
  year={2008},
  publisher={Elsevier}
}

@article{hamner2010muscle,
  title={Muscle contributions to propulsion and support during running},
  author={Hamner, Samuel R and Seth, Ajay and Delp, Scott L},
  journal={Journal of biomechanics},
  volume={43},
  number={14},
  pages={2709--2716},
  year={2010},
  publisher={Elsevier}
}

@article{doma2013kinematic,
  title={Kinematic and electromyographic comparisons between chin-ups and lat-pull down exercises},
  author={Doma, Kenji and Deakin, Glen B and Ness, Kevin F},
  journal={Sports biomechanics},
  volume={12},
  number={3},
  pages={302--313},
  year={2013},
  publisher={Taylor \& Francis}
}

@article{moffet1993load,
  title={Load-carrying during stair ascent: a demanding functional test},
  author={Moffet, H and Richards, CL and Malouin, F and Bravo, G},
  journal={Gait \& Posture},
  volume={1},
  number={1},
  pages={35--44},
  year={1993},
  publisher={Elsevier}
}

@article{snarr2017electromyographical,
  title={Electromyographical comparison of a traditional, suspension device, and towel pull-up},
  author={Snarr, Ronald L and Hallmark, Ashleigh V and Casey, Jason C and Esco, Michael R},
  journal={Journal of Human Kinetics},
  volume={58},
  pages={5},
  year={2017}
}

@article{dickinson2000animals,
  title={How animals move: an integrative view},
  author={Dickinson, Michael H and Farley, Claire T and Full, Robert J and Koehl, MAR and Kram, Rodger and Lehman, Steven},
  journal={science},
  volume={288},
  number={5463},
  pages={100--106},
  year={2000},
  publisher={American Association for the Advancement of Science}
}

@article{zajac2002biomechanics,
  title={Biomechanics and muscle coordination of human walking: Part I: Introduction to concepts, power transfer, dynamics and simulations},
  author={Zajac, Felix E and Neptune, Richard R and Kautz, Steven A},
  journal={Gait \& posture},
  volume={16},
  number={3},
  pages={215--232},
  year={2002},
  publisher={Elsevier}
}

@article{massion1992movement,
  title={Movement, posture and equilibrium: interaction and coordination},
  author={Massion, Jean},
  journal={Progress in neurobiology},
  volume={38},
  number={1},
  pages={35--56},
  year={1992},
  publisher={Elsevier}
}

@article{bouisset2008posture,
  title={Posture, dynamic stability, and voluntary movement},
  author={Bouisset, Simon and Do, Manh-Cuong},
  journal={Neurophysiologie Clinique/Clinical Neurophysiology},
  volume={38},
  number={6},
  pages={345--362},
  year={2008},
  publisher={Elsevier}
}

@article{crowninshield1981physiologically,
  title={A physiologically based criterion of muscle force prediction in locomotion},
  author={Crowninshield, Roy D and Brand, Richard A},
  journal={Journal of biomechanics},
  volume={14},
  number={11},
  pages={793--801},
  year={1981},
  publisher={Elsevier}
}

@article{vanhooren2024running,
  title   = {Dataset of running kinematics, kinetics and muscle activation at different speeds, surface gradients, cadences and with forward trunk lean},
  author  = {Van Hooren, Bas and Meijer, Kenneth},
  journal = {Data in Brief},
  volume  = {54},
  pages   = {110312},
  year    = {2024},
  doi     = {10.1016/j.dib.2024.110312}
}

@misc{openai2026gpt6astra,
  author       = {{OpenAI}},
  title        = {{GPT-6 Astra}: A New Generation of Intelligence},
  year         = {2026},
  month        = sep,
  howpublished = {[Online]. Available: \url{https://openai.com/index/gpt-6-astra/}},
}

@article{liu2026imitation,
  title={Imitation Learning from Human Motion Alone Does Not Guarantee Biomechanically Plausible Gait Kinetics},
  author={Liu, Xinyi and Ahn, Jangwhan and Lobaton, Edgar and Si, Jennie and Huang, He},
  journal={arXiv preprint arXiv:2603.12408},
  year={2026}
}

@article{park2026synergy,
  title={Muscle Synergy Priors Enhance Biomechanical Fidelity in Predictive Musculoskeletal Locomotion Simulation},
  author={Park, Ilseung and Choi, Eunsik and Ahn, Jangwhan and Ahn, Jooeun},
  journal={arXiv preprint arXiv:2603.10474},
  year={2026}
}

@article{lee2026lambda,
  title={{Lambda}-Hold Control: Human-Like Movement Emerges from a Minimal Task Reward in Predictive Musculoskeletal Simulation},
  author={Lee, Jun Hyuk and Lee, Chihyeong and Ahn, Jooeun},
  journal={arXiv preprint arXiv:2608.17030},
  year={2026}
}

@article{zhou2026reflex,
  title={Reflex-Informed Neuromuscular Reinforcement Learning for Muscle-Driven Locomotion},
  author={Zhou, Jian and Zhang, Xingyu and Ma, Rui and Cao, Yu and Xie, Shane and Zhang, Zhi-qiang},
  journal={arXiv preprint arXiv:2609.11733},
  year={2026}
}

@article{an2025arnold,
  title={Arnold: A Multi-Task, Multi-Embodiment Muscle Transformer Policy},
  author={An, Boshi and Chiappa, Alberto Silvio and Simos, Merkourios and Li, Chengkun and Mathis, Alexander},
  journal={arXiv preprint arXiv:2508.18066},
  year={2025}
}

@article{fritsche2026ultra,
  title={{ULTRA-MoCap}: A Multimodal {IMU} and {sEMG} Dataset for Upper Body Joint Kinematics Analysis},
  author={Fritsche, Oliver and Camacho, Steven and Hossain, Md Sanzid Bin and Halfpenny, Tyler and Arciniegas, Carlos and Dranetz, Joseph and Hadley, Dexter and Guo, Zhishan and Choi, Hwan},
  journal={Scientific Data},
  volume={13},
  number={1},
  pages={622},
  year={2026},
  doi={10.1038/s41597-026-06687-5}
}

@article{sun2026evolving,
  title={Evolving the Complete Muscle: Efficient Morphology-Control Co-design for Musculoskeletal Locomotion},
  author={Sun, Lidong and Zhao, Wentao and Wang, Ye and Liu, Huaping and Sun, Fuchun},
  journal={arXiv preprint arXiv:2604.12855},
  year={2026}
}

@inproceedings{cui2026libero,
  title={LIBERO-Safety: A Comprehensive Benchmark for Physical and Semantic Safety in Vision-Language-Action Models},
  author={Cui, Rongxu and Zhang, Zongzheng and Pang, Jingrui and Chi, Haohan and Guo, Jinbang and Zhang, Saining and Xie, Shaoxuan and Jin, Xin and Mu, Yao and Yang, Jiaolong and others},
  booktitle={European Conference on Computer Vision},
  pages={512--530},
  year={2026},
  organization={Springer}
}
\clearpage
\appendix

\section{Appendix}
\label{sec:app}

This appendix provides supplementary technical details, mathematical derivations, implementation descriptions, and extended experimental results to support the main findings of the paper. The content is organized as follows:

\begin{itemize}
    \item \textbf{App.~\ref{sec:app-task_spec}: Task Specification.} 
    Defines the common kinematic variables, orientation quantities, control symbols, reward terms, task settings, observations, initialization protocols, and termination criteria for the full musculoskeletal humanoid benchmark.

    \item \textbf{App.~\ref{sec:app-evaluation_details}: Evaluation Details.} 
    Describes the evaluation metrics used in the benchmark, including the task-specific success criteria, robustness under action noise, observation noise and dynamics randomization, activation cost, joint smoothness, peak efficiency steps, and EMG-envelope similarity.

    \item \textbf{App.~\ref{appendix: pure_rl_baselines}: Detailed Descriptions of Reward-Based RL Baselines.} 
    Provides detailed explanations and released training configurations for PPO, SAC, Dep-RL, and DynSyn-SAC, including network sizes, optimization schedules, update frequencies, training budgets, and checkpoint selection.

    \item \textbf{App.~\ref{appendix: muscle_heatmaps}: Detailed Muscle Activation Analysis.} 
    Provides time-step-resolved muscle activation heatmaps and biomechanical analysis for the 416-muscle and 700-muscle models, highlighting how anatomical fidelity affects load distribution and recruitment strategies.

    \item \textbf{App.~\ref{sec:app-agentic_reward_tuning}: Agentic Reward Tuning.} 
    Describes the proposed closed-loop reward refinement mechanism, where an external reasoning module periodically adjusts reward weights according to recent gait quality, posture stability, and task-progress statistics. It also details the matched DeepSeek-V4-Flash and GPT-6 Astra comparison under the same learner, architecture, training budget, evaluation protocol, and reward-weight constraint $\|w\|_1=40$.

    \item \textbf{App.~\ref{sec:app-gpt6_direct_details}: Direct GPT-6 Control.}
    Describes the direct-action GPT-6 Astra diagnostic for walk, run, and stairs, including the 354-dimensional muscle-action interface, controller-visible observations, paused rollout protocol, fall and divergence criteria, and tracking-error evaluation.

    \item \textbf{App.~\ref{sec:app-latent_action_rl}: Latent-Action RL.} 
    Explains the expert-informed hard-mode action filtering framework, where a pretrained anatomical encoder-decoder constrains RL actions within a bounded neighborhood of biologically plausible muscle activations.

    \item \textbf{App.~\ref{sec:app-emg_data_analysis}: EMG Data Processing and Comparative Analysis.} 
    Details the gait-cycle normalization, multi-cycle averaging, phase alignment, and Pearson-correlation-based comparison between simulated muscle activations and human EMG envelopes.

    \item \textbf{App.~\ref{sec:app-muscle_categorization}: Muscle Recruitment Categorization.} 
    Summarizes the anatomical keyword-based grouping of 416 muscle actuators into interpretable muscle families, allowing whole-body activation patterns to be analyzed at the functional group level.
    \item \textbf{App.~\ref{appendix: residual_rl}: Detailed Descriptions of Residual RL Policy.}
    Describes the residual reinforcement learning framework used to adapt the
    base tracking policy to challenging locomotion settings. This section details
    the residual action formulation, residual scaling strategy, Z-axis-agnostic
    imitation objective for stair navigation, elastic trajectory pacing,
    phase-dependent residual authority, activation-based muscle regularization,
    and high-impact locomotion adaptations for running.

    \item \textbf{App.~\ref{sec:app-limitations}: Limitations and Future Directions.}
    Discusses the current limitations of the benchmark and outlines future
    research directions, including dexterous manipulation, compositional and
    long-horizon task learning, broader physiological validation with EMG data,
    and sim-to-real transfer for physical musculoskeletal humanoid systems.
\end{itemize}
\subsection{Task Specification}
\label{sec:app-task_spec}

Before enumerating the environment details below, let us define common auxiliary variables, symbols, and regularizations that are broadly employed across the reward functions of multiple environments.

\textbf{Biomechanical background.}
The task families draw on prior work on movement mechanics~\cite{dickinson2000animals,zajac2002biomechanics}, postural control and anticipatory adjustments~\cite{massion1992movement,bouisset2008posture}, and muscle-force estimation under musculoskeletal redundancy~\cite{crowninshield1981physiologically}. Studies of core recruitment and lifting provide context for examining load-dependent muscle use~\cite{calatayud2015core,geisler2023effects,coratella2022front,nimbarte2014risk}. These studies motivate the biomechanical variables examined here; they do not establish that the benchmark policies reproduce the mechanisms or human recruitment patterns described in those studies.
 
\begin{figure*}[htbp]
    \centering
    \includegraphics[width=1.0\linewidth]{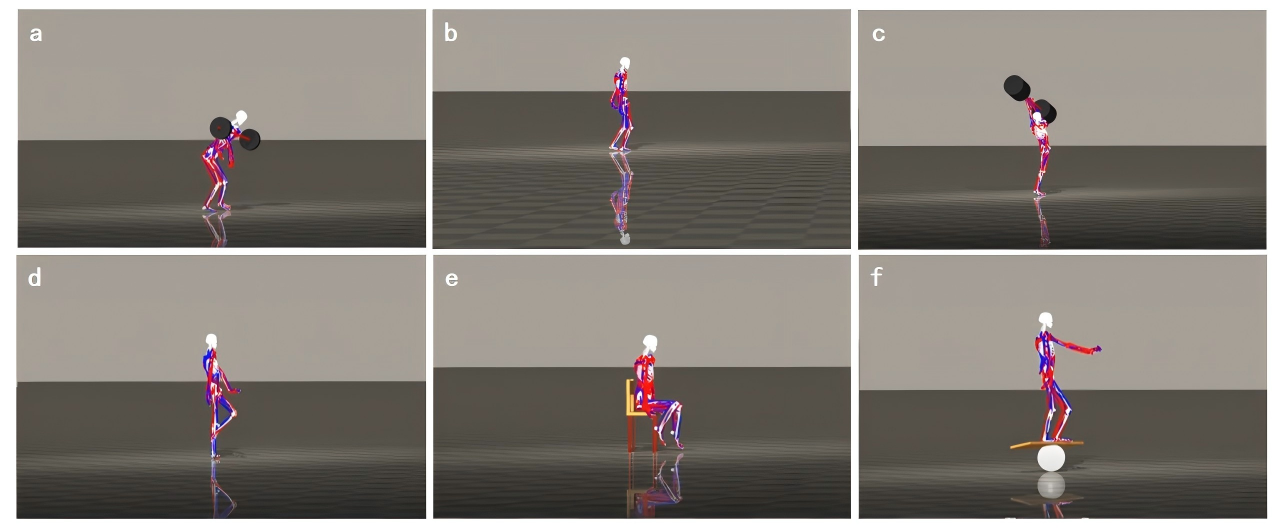} 
    \caption{
    \textbf{Illustration of six representative whole-body stabilization tasks.}
    (a) \textbf{squat}: The control policy must accurately track the dynamic vertical trajectories of the hip and knee based on a phase clock $\phi$, while maintaining a strict center-of-mass balance lock between the two feet. A specially designed ``landmine penalty'' is introduced to immediately eliminate abnormal strategies that attempt to exploit the height-tracking reward by kneeling on both knees.
    (b) \textbf{stand}: The robot is required to maintain an absolutely static upright posture from an initial neutral pose. The key challenge is that the system must use high muscle stiffness to lock the joints, while even slight center-of-mass drift or foot sliding is severely penalized. The reward function is strongly biased toward absolute stillness and posture maintenance.
    (c) \textbf{powerlift}: To sustain the extreme load, the reward function explicitly encourages a biomechanically plausible forward-leaning hip-hinge posture through the $R_{\mathrm{hinge}}$ reward, thereby effectively recruiting the powerful posterior-chain muscle groups. Once dangerous lumbar collapse is detected, the episode is immediately terminated to ensure structural safety.
    (d) \textbf{Singlelegstand}: The robot must lift one leg into a ``flamingo'' posture and strictly align the center-of-mass projection with the support foot. The task severely penalizes lateral trunk lean, and after an initial physical tolerance period of $10$ steps, the episode is immediately terminated once the lifted foot touches the ground.
    (e) \textbf{sit}: The robot is required to approach a chair and execute a precise sitting transition. To prevent cheating behaviors such as stiff-legged falling, the task enforces active knee flexion, requires the feet to be retracted beneath the support area, and introduces a ``soft-landing'' reward to limit pelvic descent velocity, ensuring a smooth motion.
    (f) \textbf{balance}: The robot is required to maintain dynamic balance on an unconstrained wobble board placed on a sphere. To overcome the common failure mode of falling backward, the task explicitly enforces a forward-leaning ``fighting defensive stance'', with a target pitch angle of $-0.15\,\mathrm{rad}$, and requires precise center-of-mass control to keep the balance board level.
    }
    \label{fig:stabilization}
\end{figure*}

\begin{figure*}[htbp]
    \centering
    \includegraphics[width=0.96\linewidth]{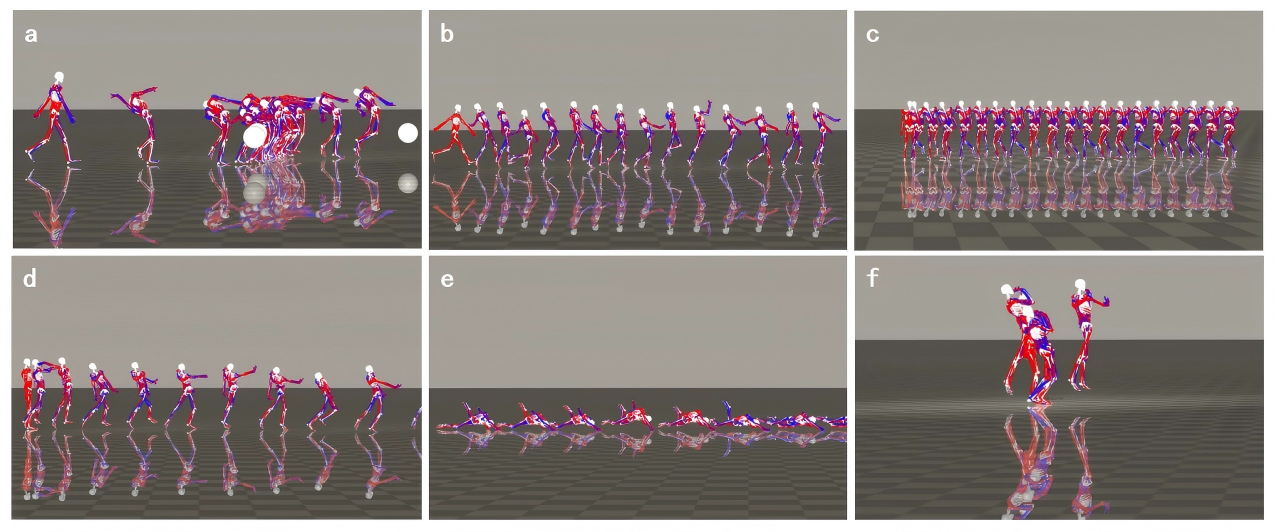} 
    \caption{
    \textbf{Visualization of representative locomotion behaviors generated by the humanoid robot.} 
    Each subfigure shows a sequence of motion frames, with red and blue skeletons indicating the two legs and body segments over time. 
    (a) \textbf{walk turn}: The robot moves while interacting with spherical obstacles. 
    (b) \textbf{walk forward}: The robot demonstrates a stable periodic gait with alternating leg support. 
    (c) \textbf{sidestep}: The robot demonstrates sustained rhythmic whole-body coordination. 
    (d) \textbf{run}: The robot runs forward with alternating leg motion while maintaining dynamic balance. 
    (e) \textbf{crawl}: The robot performs low-posture locomotion with the body close to the floor. 
    (f) \textbf{Jump}: The robot performs a dynamic aerial transition while maintaining body balance.
    }
    \label{fig:locomotion}
\end{figure*}

\begin{figure*}[htbp]
    \centering
    \includegraphics[width=0.96\linewidth]{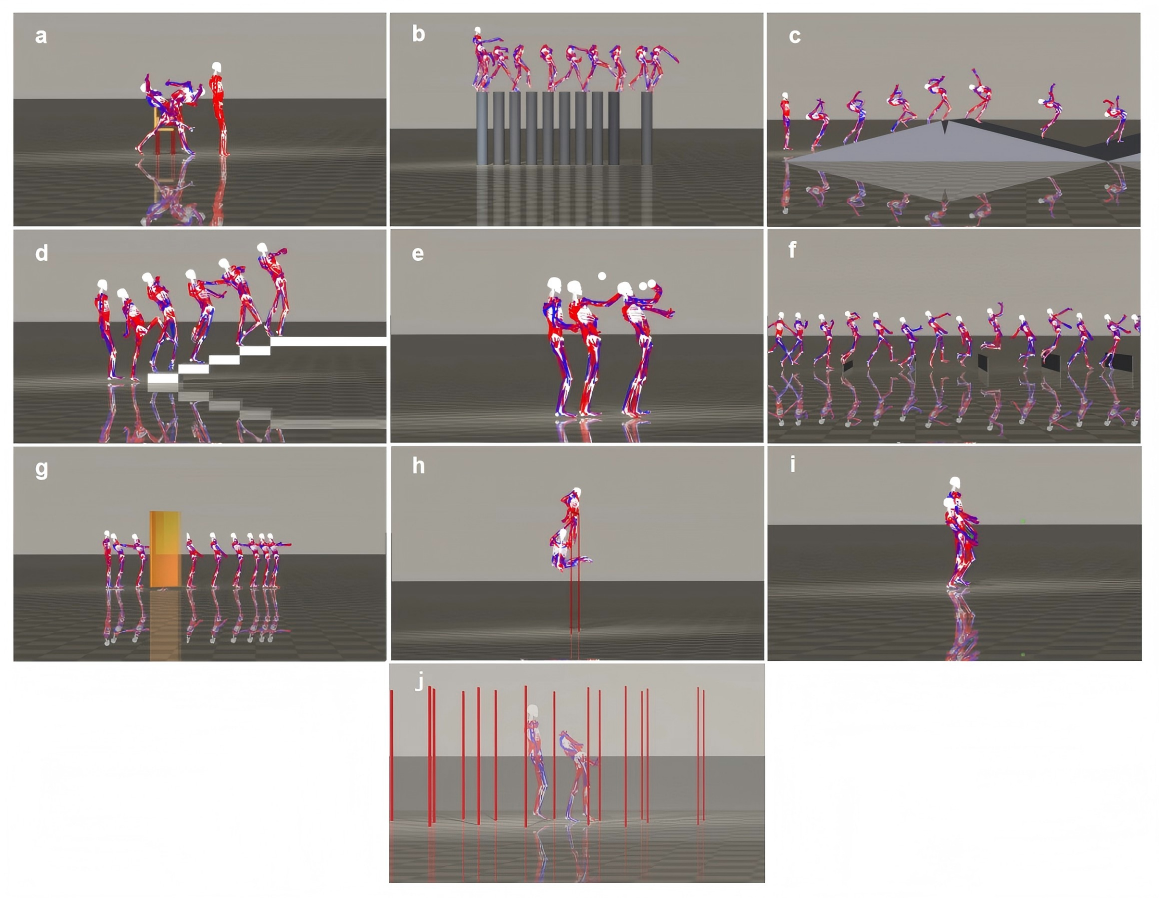} 
    \caption{
    \textbf{Illustration of ten representative terrain and object interaction tasks.}
    (a) \textbf{walkandsit}: The robot is required to approach a chair backward and sit down. Instead of using a conventional collision-avoidance penalty, the task rewards the robot for confidently utilizing downward inertia to complete the sitting motion and for comfortably leaning back against the chair after sitting, thereby reproducing human-like sitting biomechanics.
    (b) \textbf{stepstones}: A magnetic attraction reward is used to guide the foot toward the target landing positions, while strict penalties on trunk lateral bending, yaw deviation, and excessive knee flexion encourage an upright and stable traversal posture.
    (c) \textbf{slide}: The robot is required to adaptively lean forward to counteract inertia during slope transitions. A split-prevention penalty is introduced to terminate the episode once the lateral separation of the legs exceeds the biomechanical limit.
    (d) \textbf{stairs}: The robot is required to climb stairs using look-ahead terrain perception from a terrain radar. When upcoming steps are detected, an active leg-lifting reward is activated to encourage anticipatory step planning.
    (e) \textbf{reach}: The robot is required to rapidly and explosively reach randomly reset spatial targets with its dominant hand while maintaining a slightly crouched defensive posture. Since the target height varies, the task strictly constrains knee height to prevent cheating strategies such as kneeling.
    (f) \textbf{hurdle}: The task enforces a complete aerial flight phase and introduces a strict anti-flamingo-jump penalty to prevent abnormal strategies in which the robot crosses the hurdle by curling and dragging a single leg.
    (g) \textbf{open door}: The robot is required to approach, push open, and pass through a hinged door. A global attraction field guides the hand toward the door handle, while fall-prevention penalties discourage shortcut behaviors such as headbutting the door or diving through it.
    (h) \textbf{chinup}: The robot is required to pull its head above the horizontal bar from a hanging posture. The task rewards the activation of major pulling muscle groups such as the latissimus dorsi and biceps, while imposing strict penalties on body swinging and leg-kipping, forcing the motion to be completed through pure muscular strength.
    (i) \textbf{catch}: The robot is required to coordinate both arms to catch a flying cube thrown toward the chest while maintaining an upright posture. The key objective is to track the distance between both hands and the target, and a large sparse reward is given when the object is successfully caught and stabilized in mid-air.
    (j) \textbf{polewalk}: The robot uses terrain radar to navigate through a field of poles while avoiding obstacles. In accordance with the task specification and collision penalty, head or torso contact with a pole terminates the episode.
    }
    \label{fig:interaction}
\end{figure*}

The complete task suite is visually summarized in Figs.~\ref{fig:stabilization}--\ref{fig:interaction}, where Fig.~\ref{fig:stabilization} presents whole-body stabilization tasks, Fig.~\ref{fig:locomotion} illustrates representative locomotion behaviors and Fig.~\ref{fig:interaction} shows terrain and object interaction tasks.

\textbf{Kinematics and Dynamics:}
\begin{itemize}
    \item $q, \dot{q}$ denote the proprioceptive joint positions and velocities of the robot. Specifically, $q_{\text{joints}}$ denotes the subset of non-root joint angles, and $\bar{q}_{\text{hip}}, \bar{q}_{\text{knee}}$ represent the average angles of the left and right hip/knee joints.
    \item $\mathbf{p}_{(\cdot)}$ represents the 3D position of a specific body part or object in the global frame (e.g., $\mathbf{p}_{\text{pelvis}}, \mathbf{p}_{\text{foot,L}}$). The superscript $xy$ (e.g., $\mathbf{p}_{\text{pelvis}}^{xy}$) denotes its 2D horizontal projection.
    \item $\mathbf{v}_{(\cdot)}$ represents the 3D linear velocity vector. Specifically, $\mathbf{v}_{\text{CoM}}$ is the center-of-mass velocity, with $v_{x, \text{CoM}}, v_{y, \text{CoM}}, v_{z, \text{CoM}}$ being its scalar components.
    \item $z_{(\cdot)}$ denotes the vertical height of a specific body part or object from the ground (e.g., $z_{\text{CoM}}, z_{\text{pelvis}}, z_{\text{torso}}, z_{\text{head}}, z_{\text{foot}}$).
    \item $\Delta \mathbf{p}_{\text{feet}} = \mathbf{p}_{\text{foot,L}} - \mathbf{p}_{\text{foot,R}}$ denotes the relative 3D vector between the left and right feet. We use $d_{xy, \text{feet}}$ to denote the horizontal Euclidean separation between the feet, and $d_{x, \text{feet}}, d_{y, \text{feet}}, d_{z, \text{feet}}$ for its absolute scalar components.
    \item $\Delta z_{\text{leg}} = z_{\text{pelvis}} - \min(z_{\text{foot,L}}, z_{\text{foot,R}})$ defines the robot's vertical leg extension, primarily used to enforce an upright posture and prevent deep-squatting or kneeling exploits.
\end{itemize}

\textbf{Orientations and Vectors:}
\begin{itemize}
    \item $\mathbf{u}_{\text{torso}}$ is the normalized 3D unit vector pointing from the robot's pelvis to its head, used to evaluate general core verticality.
    \item $\mathbf{u}_{\text{chest}}$ is the normalized unit vector pointing forward from the robot's chest, used to penalize hunching or twisting.
    \item $\mathbf{u}_{\text{fwd}}$ is the normalized forward-facing vector derived from the pelvis orientation.
    \item $\mathbf{\hat{x}} = [1, 0, 0]^\top, \mathbf{\hat{y}} = [0, 1, 0]^\top, \mathbf{\hat{z}} = [0, 0, 1]^\top$ are the global forward, lateral, and vertical unit vectors.
    \item $y_{\text{yaw,dev}}$ denotes the forward yaw angular deviation derived from the robot's pelvis orientation, utilized to restrict unintended turning.
\end{itemize}

\textbf{Control, Environment, and Regularization:}
\begin{itemize}
    \item $a \in \mathbb{R}^M$ denotes the muscle activations at the current control step, with $a_t$ and $a_{t-1}$ specifying temporal steps.
    \item $\phi$ is the time-varying gait phase clock, represented in observations as $(\sin\phi, \cos\phi)$, with $\omega$ being its angular frequency.
    \item $\mathbf{z}_{\text{terrain}}$ is the local terrain observation grid representing relative ground heights.
    \item $\Delta \mathbf{h} = [\Delta x, h_{\text{target}}, 0]^\top$ denotes the relative 3D vector to an upcoming obstacle/hurdle.
    \item $\mathbb{I}_{\text{alive}}$ represents a constant survival bonus awarded at each step.
    \item $R_{\text{efficiency}} = \text{mean}(a^2)$ is the standard action regularization term applied across tasks to penalize excessive muscle force and promote energy-efficient behaviors. $\sum \text{penalties}$ accumulates minor supplementary regularizations (e.g., joint limits).
\end{itemize}

The detailed reward functions, task settings, specific observations, and termination criteria across all 22 tasks in the Musculoskeletal Humanoid Benchmark are systematically summarized in Tables~\ref{tab:rewards_stab} to~\ref{tab:rewards_interact} (Reward Functions) and Tables~\ref{tab:task_setup_stab} to~\ref{tab:task_setup_interact} (Setup and Termination).

\begin{table*}[!htbp]
    \centering
    \caption{Reward Functions for All 22 Musculoskeletal Benchmark Tasks (Part 1: Stabilization Series). Note: Action regularization ($R_{\text{efficiency}}$) and minor limit penalties ($\sum \text{penalties}$) are applied broadly across tasks and abbreviated where standard.}
    \label{tab:rewards_stab}
    \renewcommand{\arraystretch}{1.15}
    \begin{tabular}{
        @{} 
        >{\raggedright\arraybackslash}p{2.8cm} 
        >{\raggedright\arraybackslash}p{3.8cm} 
        >{\raggedright\arraybackslash}p{8.7cm} 
        >{\centering\arraybackslash}p{1.5cm} 
        @{}
    }
    \toprule
    \textbf{Task} & \textbf{Term} & \textbf{Formula} & \textbf{Weight} \\ \midrule
    \textbf{1. Stand} 
    & Pose preservation & $\exp(-\|q_{\text{joints}} - q_{0,\text{joints}}\|)$ & $10$ \\
    & Absolute stillness & $\exp(-10.0 \|\mathbf{v}_{\text{CoM}}\|) \cdot \exp(-0.1 \|\dot{q}\|)$ & $15$ \\
    & Feet planted & $\exp(-10.0 (\|\mathbf{v}_{\text{foot,L}}\| + \|\mathbf{v}_{\text{foot,R}}\|))$ & $10$ \\
    & Torso balance & $\exp(-5.0 \max(0, 1 - \mathbf{u}_{\text{torso}} \cdot \mathbf{\hat{z}}))$ & $10$ \\
    & Survival & $\mathbb{I}_{\text{alive}}$ & $5$ \\ \addlinespace

    \textbf{2. Powerlift} 
    & Hold stability & $\mathbb{I}(z_{\text{bar}} > 1.6\text{m} \land |\dot{z}_{\text{bar}}| < 0.1\text{m/s})$ & $100$ \\
    & Height threshold & $\mathbb{I}(z_{\text{bar}} > 1.6\text{m})$ & $50$ \\
    & Upright alignment & $\exp(-20.0(1 - \mathbf{u}_{\text{torso}} \cdot \mathbf{\hat{z}}))$ & $100$ \\
    & Lumbar spine safety& $\exp(-15.0|q_{\text{lumbar}} - (-0.4)|)$ & $80$ \\
    & Muscular effort & $\text{mean}(a)$ & $30$ \\
    & Hip hinge posture & $\mathbb{I}(q_{\text{lumbar}} > 0.1)$ & $200$ \\ \addlinespace

    \textbf{3. Singleleg} 
    & Balance (CoM) & $\exp(-15.0 \|\mathbf{p}_{\text{pelvis}}^{xy} - \mathbf{p}_{\text{foot,R}}^{xy}\|^2)$ & $500$ \\
    & Lifted foot clear & $\text{clip}(z_{\text{foot,L}} / 0.3, 0, 1)$ & $300$ \\
    & Absolute stillness & $\exp(-5.0 \|\mathbf{v}_{\text{CoM}}\|^2)$ & $400$ \\
    & Upright alignment & $\exp(-10.0(1 - \mathbf{u}_{\text{torso}} \cdot \mathbf{\hat{z}})^2)$ & $200$ \\
    & Foot touch penalty & $\mathbb{I}(z_{\text{foot,L}} < 0.05\text{m})$ & $-800$ \\
    & Lateral lean penalty & $|\mathbf{u}_{\text{torso}} \cdot \mathbf{\hat{y}}|$ & $-200$ \\ \addlinespace

    \textbf{4. Sit} 
    & Chair approach & $2.0 / (\|\Delta \mathbf{p}_{\text{chair}}^{xy}\| + 0.2)$ & $10$ \\
    & 3D Contact docking & $\exp(-\frac{\Delta x^2}{2 \cdot 0.15^2}) \cdot \exp(-\frac{\Delta z^2}{2 \cdot 0.15^2})$ & $30$ \\
    & Lateral centering & $\exp(-\frac{\Delta y^2}{2 \cdot 0.05^2})$ & $30$ \\
    & Seated leg geometry & $\exp(-2.0 |\bar{q}_{\text{hip}} - 1.5| - k_{\text{knee}} |\bar{q}_{\text{knee}} - 1.6|)$ & $40$ \\
    & Feet tucked under & $\exp(-5.0 \sum_{i} |(x_{\text{foot}, i} - x_{\text{pelvis}}) - 0.2|)$ & $30$ \\
    & Soft landing & $\exp(-2.0 |\dot{z}_{\text{pelvis}}|)$ & $5$ \\ \addlinespace

    \textbf{5. Balance} 
    & Forward lean stance & $\exp(-10.0(\theta_{\text{pitch}} - (-0.15))^2)$ (gated) & $15$ \\
    & CoM over board & $\exp(-20.0 \|\mathbf{p}_{\text{CoM}}^{xy} - \mathbf{p}_{\text{board}}^{xy}\|^2)$ & $15$ \\
    & Board over ball & $\exp(-5.0 \|\mathbf{p}_{\text{board}}^{xy} - \mathbf{p}_{\text{ball}}^{xy}\|)$ & $5$ \\
    & Board level & $\exp(-5.0 (1 - \mathbf{u}_{\text{board}} \cdot \mathbf{\hat{z}}))$ & $5$ \\
    & Stand tall & $\exp(-10.0(1.30 - z_{\text{pelvis}})^2)$ if $z<1.3$ & $4$ \\ \addlinespace

    \textbf{6. Squat} 
    & Height tracking & $\exp(-30.0(z_{\text{pelvis}} - z_{\text{target}})^2)$ & $200$ \\
    & Joint tracking & $\exp(-5.0(\bar{q} - q_{\text{target}})^2)$ & $300 / 200$ \\
    & Vertical vel tracking& $\exp(-5.0(v_{z, \text{CoM}} - v_{z, \text{target}})^2)$ & $200$ \\
    & CoM centered & $\exp(-15.0 \|\mathbf{p}_{\text{pelvis}}^{xy} - \mathbf{p}_{\text{mid\_feet}}^{xy}\|^2)$ & $300$ \\
    & Kneeling penalty & $\mathbb{I}(\min(z_{\text{knee,L}}, z_{\text{knee,R}}) < 0.25\text{m})$ & $-1000$ \\ 
    \bottomrule
    \end{tabular}
\end{table*}

\begin{table*}[!htbp]
    \centering
    \caption{Reward Functions for Musculoskeletal Benchmark Tasks (Part 2: Locomotion Series).}
    \label{tab:rewards_loco}
    \renewcommand{\arraystretch}{1.15}
    \begin{tabular}{
        @{} 
        >{\raggedright\arraybackslash}p{2.8cm} 
        >{\raggedright\arraybackslash}p{3.8cm} 
        >{\raggedright\arraybackslash}p{8.7cm} 
        >{\centering\arraybackslash}p{1.5cm} 
        @{}
    }
    \toprule
    \textbf{Task} & \textbf{Term} & \textbf{Formula} & \textbf{Weight} \\ \midrule
    \textbf{1. Walk} 
    & Velocity tracking & $\exp(-2.0 \|\mathbf{v}_{\text{CoM}} - \mathbf{v}_{\text{target}}\|^2)$ & $10$ \\
    & Kinematic imitation & $\exp(-2.0 \|q_{\text{joints}} - q_{\text{ref,joints}}\|^2)$ & $3$ \\
    & Upright / No hunch & $\exp(-10.0(1 - \mathbf{u}_{\text{torso}} \cdot \mathbf{\hat{z}}))$, $\exp(-20.0(\mathbf{u}_{\text{chest}} \cdot \mathbf{\hat{z}} + 0.1)^2)$ & $6 / 8$ \\
    & Forward face & $\exp(-5.0(1 - \mathbf{u}_{\text{chest}} \cdot \mathbf{\hat{x}}))$ & $8$ \\ \addlinespace

    \textbf{2. Crawl} 
    & Backwards velocity & $\text{clip}(-v_{x, \text{CoM}} / v_{\text{target}}, -0.2, 1.0)$ & $500$ \\
    & Prone posture & $\exp(-5(\mathbf{u}_{\text{chest}} \cdot \mathbf{\hat{z}} + 1)^2) \cdot \dots$ & $100$ \\
    & Alternating stride & $\text{clip}(|\Delta x_{\text{hands}}| / 0.3, 0, 1) \cdot \dots$ (diag gait) & $150$ \\
    & Drift penalty & $2.0|v_{y, \text{CoM}}| + |v_{x, \text{CoM}}|\mathbb{I}(v_{x} > 0)$ & $-150$ \\ \addlinespace

    \textbf{3. Run} 
    & High velocity & $\text{clip}(v_{x, \text{CoM}}, -0.5, 6.0)$, bonus if $v>3.0$ & $200 / 100$ \\
    & Asymmetric stride & $\text{clip}(d_{x, \text{feet}} + 2.0 \cdot d_{z, \text{feet}}, 0.0, 1.0)$ & $200$ \\
    & Knee flexion (swing) & $\text{clip}((\max(q_{\text{knee}}) - 0.1) / 1.0, 0, 1)$ & $200$ \\
    & Scissor penalty & $\text{clip}((0.15 - d_{y, \text{feet}}) / 0.15, 0, 1)$ & $-300$ \\
    & Kangaroo penalty & Penalize $d_{x, \text{feet}} < 0.2 \land d_{z, \text{feet}} < 0.1$ & $-300$ \\ \addlinespace

    \textbf{4. Jump} 
    & Flight gradient & $\text{clip}(\frac{1}{2}(z_{\text{foot,L}} + z_{\text{foot,R}}) / 0.1, 0, 1)$ & $600$ \\
    & Upward explosion & $\text{clip}(\dot{z}_{\text{pelvis}}, 0, 2.0)$ & $300$ \\
    & Feet separation pen. & $d_{xy, \text{feet}}$ & $-200$ \\
    & Desync jump pen. & $|q_{\text{knee,L}} - q_{\text{knee,R}}| + |q_{\text{hip,L}} - q_{\text{hip,R}}|$ & $-150$ \\ \addlinespace

    \textbf{5. Walk turn} 
    & Targeted velocity & $\text{clip}(\mathbf{v}_{\text{CoM}} \cdot \mathbf{u}_{\text{target}} / 3.0, -1.0, 1.0)$ & $200$ \\
    & Waypoint approach & $\exp(-\|\Delta \mathbf{p}_{\text{target}}\|)$ & $50$ \\
    & Target facing & $\text{clip}(\mathbf{u}_{\text{fwd}} \cdot \mathbf{u}_{\text{target}}, 0, 1)$ & $50$ \\
    & Waypoint reached & $\mathbb{I}(\|\Delta \mathbf{p}_{\text{target}}\| < 0.5\text{m})$ & $500$ \\
    & Wall collision & $\mathbb{I}_{\text{contact}}(\text{wall})$ & $-200$ \\ \addlinespace

    \textbf{6. Sidestep} 
    & Lateral velocity & $\text{clip}(v_{y, \text{CoM}} / v_{\text{target}}, -0.2, 1.0)$ & $400$ \\
    & Phase sync & $\exp(-15.0(e_L + e_R))$ & $200$ \\
    & CoM mid-feet lock & $|y_{\text{pelvis}} - \frac{1}{2}(y_{\text{foot,L}} + y_{\text{foot,R}})|$ & $-500$ \\
    & Rubber band splits & $\max(0, d_{xy, \text{feet}} - 0.3)$ & $-400$ \\ 
    \bottomrule
    \end{tabular}
\end{table*}

\begin{table*}[!htbp]
    \centering
    \caption{Reward Functions for Musculoskeletal Benchmark Tasks (Part 3: Interaction with Environment Series).}
    \label{tab:rewards_interact}
    \renewcommand{\arraystretch}{1.15}
    \begin{tabular}{
        @{} 
        >{\raggedright\arraybackslash}p{2.8cm} 
        >{\raggedright\arraybackslash}p{3.8cm} 
        >{\raggedright\arraybackslash}p{8.7cm} 
        >{\centering\arraybackslash}p{1.5cm} 
        @{}
    }
    \toprule
    \textbf{Task} & \textbf{Term} & \textbf{Formula} & \textbf{Weight} \\ \midrule
    \textbf{1. Stairs} 
    & Fwd trajectory & $\text{clip}(v_{x, \text{CoM}}, -0.2, 2.0)$ & $50$ \\
    & Upward trajectory & $\text{clip}(v_{z, \text{CoM}}, -0.5, 3.0)$ & $30$ \\
    & Vertical stable & $\exp(-5.0(0.98 - \mathbf{u}_{\text{torso}} \cdot \mathbf{\hat{z}})^2)$ & $20$ \\
    & Lateral stable & $\exp(-5.0 y_{\text{pelvis}}^2)$ & $20$ \\
    & Anticipatory lift & $\mathbb{I}(\text{radar}) \cdot \max(\dot{z}_{\text{swing}}, 0)$ & $40$ \\ \addlinespace

    \textbf{2. Hurdle} 
    & Forward velocity & $\text{clip}(v_{x, \text{CoM}}, -0.1, 4.0)$ & $300$ \\
    & Clearance (Flight) & $\mathbb{I}(\max(z_{\text{feet}}) > h_{\text{target}})$ & $100$ \\
    & Sparse clear bonus & $\mathbb{I}(\Delta x < 0 \land |y_{\text{pelvis}}| < 0.5\text{m})$ & $3000$ \\
    & Crane hopping pen. & $\text{clip}((|z_{\text{foot,L}} - z_{\text{foot,R}}| - 0.2)/0.5, 0, 1)$ & $-150$ \\ \addlinespace

    \textbf{3. Step stones} 
    & Target stone progress& Bonus when crossing stone X-coord & $800$ \\
    & Magnetic reaching & $\exp(-2.0 \cdot d_{\text{lead}}) \cdot \mathbb{I}(v_{x, \text{CoM}} > 0.1)$ & $150$ \\
    & Stand tall & $\text{clip}((\Delta z_{\text{leg}} - 0.6)/0.3, 0, 1)$ & $50$ \\
    & Overbend knee pen. & $\max(q_{\text{knee}} - 1.0, 0)$ & $-50$ \\ \addlinespace

    \textbf{4. Slide} 
    & Pitch balance & $R_{\text{gate}} \cdot \exp(-5.0 \cdot \theta_{\text{pitch}}^2)$ & $2$ \\
    & Stand tall & $R_{\text{gate}} \cdot \exp(-10.0(0.95 - \Delta z_{\text{leg}})^2)$ & $2$ \\
    & Slope forward lean & $\text{clip}(-\mathbf{u}_{\text{chest}} \cdot \mathbf{\hat{z}} / 0.3, 0, 1)$ & $20$ \\
    & Backward lean pen. & $\max(\mathbf{u}_{\text{chest}} \cdot \mathbf{\hat{z}}, 0)$ & $-100$ \\ \addlinespace

    \textbf{5. Open door} 
    & Hand attractor field & $\exp(-1.5 \|\Delta \mathbf{p}_{\text{handle}}\|)$ & $400$ \\
    & Open hinge & $\text{clip}(\theta_{\text{door}} / 1.2, 0, 1) \cdot \text{prox\_gate}$ & $500$ \\
    & Pass threshold & Sparse bonuses for $x > 1.0, 2.0$ & $2000$ \\
    & Dive/Split penalties & Penalize pelvis lead or wide feet & $-300/-400$ \\ \addlinespace

    \textbf{6. Reach} 
    & Dense reach dist & $\exp(-4.0 \|\Delta \mathbf{p}_{\text{reach}}\|)$ & $50$ \\
    & Target velocity proj & $\text{clip}(\mathbf{v}_{\text{hand}} \cdot \mathbf{u}_{\text{reach}}, 0, 2.0)$ & $20$ \\
    & Target touch bonus & $\mathbb{I}(\|\Delta \mathbf{p}_{\text{reach}}\| < 0.15\text{m})$ & $200$ \\ \addlinespace

    \textbf{7. Walk \& sit} 
    & Backwards precision & $\exp(-\|\Delta \mathbf{p}_{\text{chair}}^{xy}\|^2 / 0.005)$ & $100$ \\
    & Fast descent bonus & $\mathbb{I}(\dot{z}_{\text{pelvis}} < -1.5\text{m/s})$ & $50$ \\
    & Relaxed recline & $\mathbb{I}(\mathbf{u}_{\text{torso}} \cdot \mathbf{\hat{z}} < 0.8)$ (once seated) & $200$ \\
    & 90-degree pose & $\exp(-5(\bar{q}_{\text{hip}} - 1.5)^2) + \dots$ & $100$ \\ \addlinespace

    \textbf{8. Chin up} 
    & Spatial gradient & $\exp(-5.0 \max(0, h_{\text{bar}} - z_{\text{head}}))$ & $50$ \\
    & Muscle force filter & $\text{mean}(F_{\text{pull\_muscles}})$ & $5$ \\
    & Explosive accel & $\max(0, \ddot{z}_{\text{pelvis}})$ & $40$ \\
    & Anti-kipping wiggle & $\|\dot{q}_{\text{legs}}\|$ & $-20$ \\ \addlinespace

    \textbf{9. Catch} 
    & Hand-object reach & $\exp(-3.0 \cdot d_{\text{catch}})$ & $100$ \\
    & Interception bonus & $\mathbb{I}(d_{\text{catch}} < 0.15\text{m} \land z_{\text{cube}} > 0.5\text{m})$ & $500$ \\
    & Drop penalty & $\mathbb{I}(z_{\text{cube}} < 0.1\text{m})$ & $-200$ \\ \addlinespace

    \textbf{10. Pole walk} 
    & Target feet phases & $\exp(-10.0(\Delta x_{\text{foot}} \pm 0.25\sin\phi)^2)$ & $400$ \\
    & Dynamic radar gap & E.g., $\text{clip}(2.0 \cdot v_{y, \text{CoM}}, 0, 1)$ if left gap & $800$ \\
    & Fatal collision & $\mathbb{I}_{\text{fatal\_contact}}$ (head/torso to pole) & $-1500$ \\ 
    \bottomrule
    \end{tabular}
\end{table*}
\begin{table*}[!htbp]
    \centering
    \caption{Task Settings, Specific Observations, and Termination Criteria (Part 1: Stabilization Series). \textit{Base Observation ($O_{\text{base}}$): All tasks fundamentally include proprioceptive states $(q, \dot{q}, \mathbf{v}_{\text{CoM}})$ and muscle states (length, velocity, force, activation). Only task-specific exteroceptive or goal variables are listed below.}}
    \label{tab:task_setup_stab}
    \renewcommand{\arraystretch}{1.15}
    \begin{tabular}{
        @{} 
        >{\raggedright\arraybackslash}p{2.5cm} 
        >{\raggedright\arraybackslash}p{4.5cm} 
        >{\raggedright\arraybackslash}p{5.0cm} 
        >{\raggedright\arraybackslash}p{4.5cm} 
        @{}
    }
    \toprule
    \textbf{Task} & \textbf{Specific Obs ($O_{\text{task}}$)} & \textbf{Initialization (Brief)} & \textbf{Termination Criteria} \\ \midrule

    \textbf{1. Stand} & Torso quat, feet height & Default neutral, zero vel. & $z_{\text{CoM}} < 0.8$ or tilt $>0.8$ \\
    \textbf{2. Powerlift} & Dumbbell $\mathbf{p}_{\text{bar}}$, $\Delta \mathbf{p}_{\text{bar}}$, tilt & High CoM, pre-loaded lumbar & $z_{\text{pelvis}} < 0.7$ or $q_{\text{lumbar}} > 0.5$ \\
    \textbf{3. Singleleg} & Torso quat, feet/CoM $z$ & Flamingo pose, noise & $z_{\text{pelvis}} < 0.6$, dropped foot \\
    \textbf{4. Sit} & Chair relative $\Delta \mathbf{p}_{\text{chair}}$ & In front of chair, hips pre-flexed & $z_{\text{pelvis}} < 0.25$, invalid contact \\
    \textbf{5. Balance} & Board $\mathbf{p}, \mathbf{q}$, CoM/Ball $\Delta \mathbf{p}$ & Drop $z=1.28$m, guard stance & $z_{\text{pelvis}} < 0.70$, pitch fall \\
    \textbf{6. Squat} & Phase $\phi$, target $z_{\text{target}}$ & Squat start pose, barbell hold & $z_{\text{pelvis}} < 0.35$, kneeling \\ 
    \bottomrule
    \end{tabular}
\end{table*}

\begin{table*}[!htbp]
    \centering
    \caption{Task Settings, Specific Observations, and Termination Criteria (Part 2: Locomotion Series).}
    \label{tab:task_setup_loco}
    \renewcommand{\arraystretch}{1.15}
    \begin{tabular}{
        @{} 
        >{\raggedright\arraybackslash}p{2.5cm} 
        >{\raggedright\arraybackslash}p{4.5cm} 
        >{\raggedright\arraybackslash}p{5.0cm} 
        >{\raggedright\arraybackslash}p{4.5cm} 
        @{}
    }
    \toprule
    \textbf{Task} & \textbf{Specific Obs ($O_{\text{task}}$)} & \textbf{Initialization (Brief)} & \textbf{Termination Criteria} \\ \midrule

    \textbf{1. Walk} & Phase $\phi$, error $\Delta q = q - q_{\text{ref}}$ & Sampled from synthetic walk & $z_{\text{CoM}} < 0.5$ or tilt $>0.8$ \\
    \textbf{2. Crawl} & Torso quat, feet height & Prone quadruped, ground $z=0.25$ & Fwd move, twist, flip \\
    \textbf{3. Run} & Torso angle, feet height & Standard standing & $z_{\text{pelvis}} < 0.45$, fall backward \\
    \textbf{4. Jump} & Torso angle, $d_{xy, \text{feet}}$ & Micro-squat, pre-flexed & $z_{\text{pelvis}} < 0.45$, wide split \\
    \textbf{5. Walk turn} & Target $\Delta \mathbf{p}$, ID, facing $a$ & Stand, 20-step novice protection & $z_{\text{pelvis}} < 0.55$, $z_{\text{head}} < 0.8$ \\
    \textbf{6. Sidestep} & Phase $\phi$, lat dist $\Delta y_{\text{feet}}$ & Crab guard stance, hips wide & $z_{\text{pelvis}} < 0.45$, split $>0.5$m \\ 
    \bottomrule
    \end{tabular}
\end{table*}

\begin{table*}[!htbp]
    \centering
    \caption{Task Settings, Specific Observations, and Termination Criteria (Part 3: Interaction with Environment Series).}
    \label{tab:task_setup_interact}
    \renewcommand{\arraystretch}{1.15}
    \begin{tabular}{
        @{} 
        >{\raggedright\arraybackslash}p{2.5cm} 
        >{\raggedright\arraybackslash}p{4.5cm} 
        >{\raggedright\arraybackslash}p{5.0cm} 
        >{\raggedright\arraybackslash}p{4.5cm} 
        @{}
    }
    \toprule
    \textbf{Task} & \textbf{Specific Obs ($O_{\text{task}}$)} & \textbf{Initialization (Brief)} & \textbf{Termination Criteria} \\ \midrule

    \textbf{1. Stairs} & Phase $\phi$, radar $\mathbf{z}_{\text{terrain}} \in \mathbb{R}^9$ & Walk posture, initial fwd vel & $z_{\text{pelvis}} < 0.75$, lat drift \\
    \textbf{2. Hurdle} & Phase $\phi$, nearest $\Delta \mathbf{h}$ & Standing, scan dynamic hurdles & $z_{\text{pelvis}} < 0.55$, heavy yaw \\
    \textbf{3. Step stones}& Phase $\phi$, stones $\Delta \mathbf{p}_{\text{stones}}$ & Elevated platform $z=3.05$m & Fall from height, $z_{\text{pelvis}}<2.5$ \\
    \textbf{4. Slide} & Phase $\phi$, normal $\mathbf{n}_{\text{slope}}$ & Elevated over wave terrain & Backwards fall, split $>1.2$m \\
    \textbf{5. Open door} & Handle $\Delta \mathbf{p}$, door $\theta$, target & Stand front of closed door & $z_{\text{pelvis}} < 0.65$, head $< 1.1$ \\
    \textbf{6. Reach} & Phase $\phi$, reach $\Delta \mathbf{p}_{\text{reach}}$ & Quarter-squat, hands raised & Knee drop $< 0.15$m \\
    \textbf{7. Walk \& sit} & Chair $\Delta \mathbf{p}$, phase $\mathbb{I}_{\text{seated}}$ & Face away, inject backward vel & Ground contact (head/knees) \\
    \textbf{8. Chin up} & Phase (0), bar dist $\Delta z_{\text{chin}}$ & Hanging, flexed shoulders & Fall off bar ($z_{\text{pelvis}} < 0.5$) \\
    \textbf{9. Catch} & Cube $\mathbf{p}, \mathbf{v}$, dist $d_{\text{catch}}$ & Anticipatory basket, cube spawn & Drop cube ($z_{\text{cube}} < 0.05$) \\
    \textbf{10. Pole walk} & Phase $\phi$, radar $\mathbf{z}_{\text{radar}} \in \mathbb{R}^7$ & Elevated $1.05$m, high core gain & Fatal upper-body collision \\ 
    \bottomrule
    \end{tabular}
\end{table*}

\subsection{Evaluation Details}
\label{sec:app-evaluation_details}
In this section, we present a detailed quantitative evaluation of the benchmarked tasks. We systematically assess the performance, sample efficiency, and zero-shot robustness of various reinforcement learning baselines under increasingly difficult conditions. The detailed results for the Stabilization (Table~\ref{tab:benchmark_robustness_stabilization}), Locomotion (Table~\ref{tab:benchmark_robustness}), and Interaction (Table~\ref{tab:benchmark_robustness_interaction}) series are summarized in the tables below.

The full-suite robustness tables use the benchmark success criteria shared by the four reward-based baselines. The focused MuscleMimic--DepRL comparison in Fig.~\ref{fig:musclemimic_vs_deprl} instead uses a matched cross-paradigm diagnostic protocol: within each task, both methods are evaluated using identical success conditions, perturbation distributions, and perturbation scales. The Jump diagnostic uses a less stringent completion condition than the full-suite benchmark, with the same condition applied to both methods. Consequently, the DepRL points in Fig.~\ref{fig:musclemimic_vs_deprl}, including the Jump sweep, are not expected to reproduce the corresponding entries in Table~\ref{tab:benchmark_robustness}. The focused comparison is reported only as a within-protocol diagnostic and is excluded from the full-suite aggregate robustness scores.

\subsubsection{Success Rate}
The reported train-environment success rate restores the training environment, including its native noise. It is distinct from the zero-scale point of a separately configured robustness sweep. Within each task, the four full-suite reward-based baselines share observations, rewards, termination conditions, and evaluation. Task-family and all-task success rates are arithmetic means over tasks; nonzero-success coverage counts tasks with observed SR $>0$.

Success is evaluated separately from environment termination. The termination rules in Tables~\ref{tab:task_setup_stab}--\ref{tab:task_setup_interact} determine when a rollout stops, whereas the evaluator assigns a binary label using the task-specific conditions in Table~\ref{tab:success_criteria}. A native positive \texttt{solved} signal, when provided by an environment, is accepted as success; otherwise the tracked completion condition is used. Reaching a survival horizon explicitly requires that the episode did not terminate earlier. For the remaining tasks, an earlier termination stops further progress but is not introduced as a second, universal success predicate. Formally, for $N$ evaluation episodes,
\begin{equation}
\begin{aligned}
\text{Success Rate} &= \frac{1}{N} \sum_{i=1}^{N} \mathbb{I}\!\left(\mathcal{S}_{\mathrm{task}}^{(i)}\right),
\end{aligned}
\end{equation}
where $\mathbb{I}(\cdot)$ is the indicator function and $\mathcal{S}_{\mathrm{task}}^{(i)}$ is the corresponding code-level condition below. Threshold crossings are accumulated over the episode: $\max$ and $\min$ therefore refer to the recorded rollout history rather than only the terminal state.

\begin{table*}[t]
    \centering
    \caption{Task-specific success criteria used by the shared full-suite evaluator for PPO, SAC, DepRL, and DynSyn-SAC. These criteria are distinct from the environment termination rules. A positive native \texttt{solved} signal also counts as success for every task.}
    \label{tab:success_criteria}
    \scriptsize
    \setlength{\tabcolsep}{3.5pt}
    \begin{tabularx}{\textwidth}{>{\raggedright\arraybackslash}p{0.105\textwidth} X >{\raggedright\arraybackslash}p{0.105\textwidth} X}
        \toprule
        \textbf{Task} & \textbf{Success criterion} & \textbf{Task} & \textbf{Success criterion} \\
        \midrule
        Stand & Reach 1,000 evaluation steps without earlier termination. & Sidestep & Maximum absolute lateral pelvis displacement from the start is at least $9.0$ m. \\
        Powerlift & Reach 900 steps and attain a maximum dumbbell height of at least $1.6$ m. & Stairs & Maximum pelvis height is at least $1.6$ m. \\
        Single-leg stand & Reach 1,000 evaluation steps without earlier termination. & Hurdle & The recorded number of cleared hurdles reaches the environment's total (five if unavailable). \\
        Sit & The environment's \texttt{has\_sat\_down} flag becomes true. & Step stones & \texttt{current\_stone\_idx} reaches the final stone index. \\
        Balance & Run for more than 350 steps and attain a board--ball horizontal separation below $0.5$ m. & Slide & Maximum forward pelvis displacement from the start is at least $18.0$ m. \\
        Squat & Pelvis height first falls below $0.75$ m and subsequently rises above $0.90$ m. & Door open & Forward displacement is at least $2.0$ m, the door-open flag is true, or maximum door angle exceeds $1.0$ rad. \\
        Walk & Reach 1,000 evaluation steps without earlier termination. & Reach & Accumulate at least two target touches. \\
        Crawl & Maximum backward pelvis displacement from the start is at least $2.0$ m. & Walk and sit & The sit flag becomes true and the rollout reaches 500 steps. \\
        Run & Maximum forward pelvis displacement from the start is at least $15.0$ m. & Chin-up & Keep the head more than $0.1$ m above the bar for at least 25 consecutive steps. \\
        Jump & Complete at least 10 hops; a hop is counted after pelvis height crosses above $1.00$ m and then below $0.95$ m. & Catch & The catch flag becomes true and the rollout reaches 200 steps. \\
        Walk turn & \texttt{current\_target\_idx} reaches at least 2. & Pole walk & Maximum forward pelvis displacement from the start is at least $1.6$ m. \\
        \bottomrule
    \end{tabularx}
\end{table*}

\subsubsection{Robustness}
Action noise, observation noise, and dynamics randomization are introduced to assess robustness~\cite{he2024dynsyn, radosavovic2024real} of the methods. It evaluates whether a policy can maintain balance, preserve task performance, or quickly recover its intended trajectory when facing stochastic perturbations and randomized dynamics~\cite{he2024dynsyn, radosavovic2024real}. To comprehensively test robustness under Sim2Real-style perturbations~\cite{tobin2017domain, peng2018sim, tan2018sim, akkaya2019solving}, we consider the following:

First, \textbf{action noise} mimics stochastic execution errors in motor-neural excitation signals or low-level actuator commands. Sim-to-real robot learning studies that model stochastic action effects and transition mismatch between simulation and the real world~\cite{desai2020stochastic}, together with continuous-control reinforcement learning studies that examine additive action noise~\cite{hollenstein2022action}, motivate such action-space perturbations. In our implementation, Gaussian noise is applied to the normalized action vector generated by the policy. Since the policy action space is bounded within $[-1, 1]$, the perturbed action is clipped to this valid action range:
\begin{equation}
\begin{aligned}
\mathbf{u}'_t &= \operatorname{clip}_{[-1,1]} \left( \mathbf{u}_t + \boldsymbol{\epsilon}_t^{\mathrm{act}} \right), \\
\boldsymbol{\epsilon}_t^{\mathrm{act}} &\sim \mathcal{N} \left( \mathbf{0}, \sigma_{\mathrm{act}}^2 \mathbf{I} \right),
\end{aligned}
\end{equation}
where $\mathbf{u}_t$ denotes the policy's original normalized action output, $\boldsymbol{\epsilon}_t^{\mathrm{act}}$ denotes Gaussian action noise, $\sigma_{\mathrm{act}}$ controls the action noise magnitude, and $\mathbf{u}'_t$ denotes the noisy action applied to the environment.

Second, \textbf{observation noise} mimics sensing errors from external sensors such as IMUs or proprioceptive feedback such as muscle spindles and Golgi tendon organs. This perturbation is also consistent with prior sim-to-real studies that perturb or randomize sensory inputs to improve robustness under observation mismatch~\cite{tobin2017domain, peng2018sim, akkaya2019solving}. In our implementation, Gaussian noise is added to the normalized policy observation vector before it is passed to the policy network:
\begin{equation}
\begin{aligned}
\tilde{\mathbf{s}}'_t &= \operatorname{clip}_{[-20,20]} \left( \tilde{\mathbf{s}}_t + \boldsymbol{\epsilon}_t^{\mathrm{obs}} \right), \\
\boldsymbol{\epsilon}_t^{\mathrm{obs}} &\sim \mathcal{N} \left( \mathbf{0}, \sigma_{\mathrm{obs}}^2 \mathbf{I} \right),
\end{aligned}
\end{equation}
where $\tilde{\mathbf{s}}_t$ denotes the normalized observation vector before perturbation, 
$\boldsymbol{\epsilon}_t^{\mathrm{obs}}$ denotes Gaussian observation noise, 
$\sigma_{\mathrm{obs}}$ controls the observation noise level, and 
$\tilde{\mathbf{s}}'_t$ denotes the perturbed normalized observation received by the policy. The clipping operation is applied element-wise to keep the noisy normalized observation within the predefined numerical range.

Third, \textbf{dynamics randomization} mimics individual variations in muscle physiology and uncertainty in muscle force capacity. To improve robustness against real-world dynamics mismatch~\cite{peng2018sim, tan2018sim, akkaya2019solving}, this setting follows the broader sim-to-real practice of randomizing physical parameters, such as mass, friction, actuator properties, and other dynamics-related quantities. At the beginning of each evaluation episode, we independently sample a random perturbation for each muscle and keep it fixed throughout the episode. Specifically, for the $m$-th muscle, we sample:
\begin{equation}
\begin{aligned}
\delta_m &\sim \mathcal{U} \left( -\sigma_{\mathrm{dyn}}, \sigma_{\mathrm{dyn}} \right), \\
r_m &= 1 + \delta_m, \\
\mathrm{Gain}'_{m} &= \mathrm{Gain}_{m} \times r_m,
\end{aligned}
\end{equation}
where $\delta_m$ denotes the independently sampled perturbation for the $m$-th muscle, 
$r_m$ denotes the corresponding multiplicative gain factor, 
$\mathrm{Gain}_{m}$ denotes the original maximum force-generation parameter of the $m$-th muscle, and 
$\mathrm{Gain}'_{m}$ denotes the randomized muscle gain used during evaluation. The parameter $0 \leq \sigma_{\mathrm{dyn}} < 1$ controls the perturbation strength and ensures positive muscle gains. For example, $\sigma_{\mathrm{dyn}} = 0.20$ means that each muscle's maximum force capacity is independently scaled by a random factor in $[0.8, 1.2]$.

Across these perturbation settings, we report the average reward, survival steps, and true success rate to quantify how reliably each policy maintains task performance under noise and randomized dynamics. Higher values of these metrics under perturbation indicate stronger robustness. When compared with the clean evaluation setting, smaller performance degradation indicates better robustness.

Action and dynamics perturbation scales are $(0,0.05,0.10,0.15,0.20)$, and observation scales are $(0,0.02,0.05,0.08,0.10)$. For each task and perturbation type, robustness is the trapezoidal area under success versus perturbation scale, divided by the tested scale range. The aggregate score averages these normalized areas equally over tasks and perturbation types, on a 0--100 scale when success is expressed as a percentage. Task-specific cumulative rewards are reported separately and are never pooled across tasks.

\subsubsection{Activation Cost}
Activation cost~\cite{zuo2024self, wei2026scaling, he2024dynsyn, selinger2015humans, michaud2021fair} is the mean squared muscle activation across muscles and time. It is an effort proxy, not a direct measurement of metabolic energy. It is interpreted only for successful, behaviorally comparable policies. Formally:
\begin{equation}
\begin{aligned}
E_{\mathrm{muscle}} &= \frac{1}{T} \sum_{t=1}^{T} \left( \frac{1}{M} \sum_{m=1}^{M} a_{m,t}^{2} \right),
\end{aligned}
\end{equation}
where $a_{m,t} \in [0,1]$ denotes the actuator activation of the $m$-th muscle at time step $t$, $T$ denotes the total number of simulation steps in each episode, and $M$ denotes the total number of muscles. A lower $E_{\mathrm{muscle}}$ indicates lower mean squared activation; it does not by itself establish better task performance or greater physiological fidelity.

\subsubsection{Joint Smoothness}
Joint smoothness is evaluated by analyzing high-frequency fluctuations in joint motion. Prior work has used smoothness-related kinematic measures to characterize coordinated and biologically plausible movement~\cite{flash1985coordination, balasubramanian2015analysis}. Lower high-frequency jerk indicates smoother motor behavior with fewer abrupt joint-level changes, which is consistent with the low-pass characteristics of biological motor control~\cite{flash1985coordination, balasubramanian2015analysis}. 

In our implementation, smoothness is computed from the joint angular velocity signal returned by the physics engine, denoted as $v_{j,t}$ for the $j$-th joint at time step $t$. Since our implementation applies a second-order finite difference to joint velocity, the resulting finite-difference quantity is proportional to the third-order derivative of joint angle, commonly referred to as jerk:
\begin{equation}
\begin{aligned}
\mathrm{D}^{(2)} v_{j,t} &= \frac{ v_{j,t} - 2v_{j,t-1} + v_{j,t-2} }{ \Delta t^2 } \\
&\approx \frac{\mathrm{d}^2 v_j(t)}{\mathrm{d}t^2} = \frac{\mathrm{d}^3 \theta_j(t)}{\mathrm{d}t^3}.
\end{aligned}
\end{equation}
We therefore quantify joint-level high-frequency fluctuation using the mean squared finite-difference jerk over all joints and time steps:
\begin{equation}
\begin{aligned}
J_{\mathrm{sq}} &= \frac{1}{(N-2)N_{\mathrm{joint}}} \sum_{t=3}^{N} \sum_{j=1}^{N_{\mathrm{joint}}} \left( \mathrm{D}^{(2)} v_{j,t} \right)^2,
\end{aligned}
\end{equation}
where $N_{\mathrm{joint}}$ denotes the total number of joints, $N$ denotes the total number of simulation steps in the episode, $v_{j,t}$ denotes the joint angular velocity of the $j$-th joint at time step $t$, $\Delta t$ denotes the simulation time step, and $\mathrm{D}^{(2)} v_{j,t}$ denotes the second-order finite difference of joint angular velocity.

For visualization and task-level comparison, we report the logarithmic mean squared jerk:
\begin{equation}
\begin{aligned}
\mathrm{Smoothness}_{\mathrm{jerk}} &= \log_{10} \left( J_{\mathrm{sq}} + \epsilon \right),
\end{aligned}
\end{equation}
where $\epsilon$ is a small constant used for numerical stability. The mean squared angular jerk has units of rad$^2$/s$^6$ before logarithmic display. Lower values indicate fewer high-frequency joint fluctuations. Like activation cost, smoothness is interpreted only for successful, behaviorally comparable policies and does not establish human similarity.

\subsubsection{Peak Efficiency Steps}
We introduce this metric to identify the training phase at which an algorithm achieves its peak evaluation performance. Peak Efficiency Steps pinpoints the number of environment interactions required to obtain the highest expected evaluation return over the full training horizon. Formally, it is defined as the evaluated training step $t$ that maximizes the expected evaluation return:
\begin{equation}
\begin{aligned}
T_{\mathrm{max}} &= \operatorname*{arg\,max}_{t \in \mathcal{T}} \mathbb{E} \left[ R_t \right],
\end{aligned}
\end{equation}
where $\mathcal{T}$ denotes the set of evaluated training steps throughout the learning process, and $R_t$ denotes the cumulative reward obtained during evaluation at training step $t$. We also denote the corresponding maximum expected return as
\begin{equation}
\begin{aligned}
R_{\mathrm{max}} &= \max_{t \in \mathcal{T}} \mathbb{E} \left[ R_t \right].
\end{aligned}
\end{equation}
A smaller $T_{\mathrm{max}}$, particularly when paired with a high $R_{\mathrm{max}}$, indicates that the algorithm can rapidly discover a high-performing behavioral policy, reflecting better sample efficiency and faster peak skill acquisition.

\subsubsection{EMG Similarity}
EMG-envelope similarity compares simulated activations with matched human reference envelopes, following MuscleMimic~\cite{li2026towards}. Walking and stairs use Gait120~\cite{boo2025comprehensive}; running uses the 3-m/s recordings of Van Hooren and Meijer~\cite{vanhooren2024running}. Walking results in Table~\ref{tab:residual_results} report three selected muscles individually; running reports the mean over 11 independently recorded lower-limb EMG channels, and stairs reports the mean over 12 matched lower-limb muscles. Simulated muscle activation is treated as an activation-envelope proxy, not as a unique ground-truth measure of physiological control.

For each selected lower-limb muscle, simulated activations and human EMG recordings are processed in a gait-cycle-normalized form. Human reference envelopes are min--max normalized per muscle, resampled to 101 gait-cycle points, and cycle-averaged. Simulated profiles are cyclically shifted to maximize correlation, as detailed in App.~\ref{sec:app-emg_data_analysis}. Muscle activations from RL are segmented into gait cycles, interpolated to the same normalized time base, and averaged across cycles to obtain representative activation profiles. We then compare the waveform similarity between the simulated activation envelope and the processed human EMG envelope using Pearson correlation:
\begin{equation}
\begin{aligned}
r_m &= \frac{ \sum_{t=1}^{T} \left( a_{\mathrm{sim},m}(t) - \bar{a}_{\mathrm{sim},m} \right) \left( e_{\mathrm{EMG},m}(t) - \bar{e}_{\mathrm{EMG},m} \right) }{ \sqrt{ \sum_{t=1}^{T} \left( a_{\mathrm{sim},m}(t) - \bar{a}_{\mathrm{sim},m} \right)^2 } } \\
&\quad \times \frac{ 1 }{ \sqrt{ \sum_{t=1}^{T} \left( e_{\mathrm{EMG},m}(t) - \bar{e}_{\mathrm{EMG},m} \right)^2 } },
\end{aligned}
\end{equation}
where $a_{\mathrm{sim},m}(t)$ denotes the gait-cycle-normalized simulated activation envelope of muscle $m$ at normalized time $t$, $e_{\mathrm{EMG},m}(t)$ denotes the corresponding processed human-reference EMG envelope, and $\bar{a}_{\mathrm{sim},m}$ and $\bar{e}_{\mathrm{EMG},m}$ denote their temporal means. We report both per-muscle correlations and the average correlation across the selected muscles:
\begin{equation}
\begin{aligned}
r_{\mathrm{EMG}} &= \frac{1}{M} \sum_{m=1}^{M} r_m,
\end{aligned}
\end{equation}
where $M$ is the number of muscles included in the comparison.

Higher $r_m$ indicates closer phase-optimized waveform shape. Because signals are normalized and cyclically shifted, this metric does not measure absolute amplitude or absolute timing agreement. It complements task success and kinematic tracking; improved success does not necessarily imply improved EMG-envelope agreement.

\begin{table*}[htbp]
    \vspace*{5pt}
    \centering
    \rmfamily 
    
    \caption{Benchmark Performance and Robustness Evaluation: Stabilization. Train-env. SR retains native training noise; robustness columns report separate perturbation sweeps (App.~\ref{sec:app-evaluation_details}).}
    \label{tab:benchmark_robustness_stabilization}
    
    \setlength{\tabcolsep}{1.8pt}
    \scriptsize
    
    \resizebox{\linewidth}{!}{
    \begin{tabular}{lc c *{15}{c} ccc}
        \toprule
        \multirow{3}{*}{\textbf{Task}} 
        & \multirow{3}{*}{\textbf{Algorithm}} 
        & \textbf{Activation} 
        & \multicolumn{15}{c}{\textbf{Robustness (\%)}} 
        & \textbf{Max} 
        & \textbf{Train-env.} 
        & \textbf{$\log_{10}$smooth} \\
        
        & & \textbf{Cost} 
        & \multicolumn{5}{c}{Action} 
        & \multicolumn{5}{c}{Observation} 
        & \multicolumn{5}{c}{Dynamic} 
        & \textbf{Steps} ($\times 10^7$) 
        & \textbf{SR (\%)} 
        & \textbf{(rad$^2$/s$^6$)} \\
        
        \cmidrule(lr){4-8} 
        \cmidrule(lr){9-13} 
        \cmidrule(lr){14-18}
        
        & & 
        & 0 & 0.05 & 0.1 & 0.15 & 0.2 
        & 0 & 0.02 & 0.05 & 0.08 & 0.10 
        & 0 & 0.05 & 0.1 & 0.15 & 0.2 
        & & & \\
        
        \midrule
        
        \multirow{4}{*}{stand}
        & DepRL & 0.4488
        & 94 & 94 & 90 & 96 & 98
        & 100 & 12 & 0 & 0 & 0
        & 98 & 48 & 28 & 12 & 4
        & 3.74 & 86 & 3.152 \\
        
        & SAC & 0.0484
        & 100 & 100 & 100 & 100 & 96
        & 100 & 0 & 0 & 0 & 0
        & 100 & 100 & 100 & 100 & 100
        & 7.13 & 100 & 1.875 \\
        
        & PPO & 0.3691
        & 0 & 0 & 0 & 0 & 0
        & 0 & 0 & 0 & 0 & 0
        & 0 & 0 & 0 & 0 & 0
        & 1.97 & 0 & 5.479 \\
        
        & DynSyn-SAC & 0.2859
        & 52 & 42 & 56 & 46 & 36
        & 46 & 46 & 22 & 30 & 14
        & 40 & 32 & 40 & 30 & 30
        & 5.44 & 52 & 2.104 \\
        
        \cmidrule{1-21}
        
        \multirow{4}{*}{powerlift}
        & DepRL & 0.3356
        & 68 & 68 & 64 & 56 & 52
        & 62 & 62 & 38 & 4 & 0
        & 66 & 70 & 52 & 42 & 38
        & 4.14 & 38 & 2.631 \\
        
        & SAC & 0.2106
        & 98 & 94 & 94 & 88 & 80
        & 96 & 0 & 0 & 0 & 0
        & 98 & 94 & 88 & 64 & 2
        & 6.78 & 90 & 2.825 \\
        
        & PPO & 0.3955
        & 0 & 0 & 0 & 0 & 0
        & 0 & 0 & 0 & 0 & 0
        & 0 & 0 & 0 & 0 & 0
        & 1.76 & 0 & 5.042 \\
        
        & DynSyn-SAC & 0.3631
        & 100 & 100 & 100 & 98 & 92
        & 100 & 100 & 96 & 94 & 90
        & 100 & 98 & 94 & 90 & 82
        & 8.50 & 100 & 5.781 \\
        
        \cmidrule{1-21}
        
        \multirow{4}{*}{squat}
        & DepRL & 0.3781
        & 44 & 46 & 38 & 32 & 30
        & 24 & 56 & 34 & 62 & 26
        & 46 & 38 & 12 & 20 & 10
        & 14.02 & 42 & 2.984 \\
        
        & SAC & 0.1184
        & 0 & 0 & 0 & 4 & 0
        & 0 & 0 & 0 & 0 & 0
        & 4 & 0 & 0 & 0 & 0
        & 4.79 & 0 & 1.708 \\
        
        & PPO & 0.3455
        & 0 & 0 & 0 & 0 & 0
        & 0 & 0 & 0 & 0 & 0
        & 0 & 0 & 0 & 0 & 0
        & 2.01 & 0 & 5.390 \\
        
        & DynSyn-SAC & 0.1955
        & 48 & 38 & 32 & 26 & 22
        & 50 & 44 & 38 & 36 & 30
        & 48 & 42 & 32 & 26 & 18
        & 7.00 & 42 & 6.503 \\
        
        \cmidrule{1-21}
        
        \multirow{4}{*}{sit}
        & DepRL & 0.3327
        & 96 & 100 & 100 & 98 & 92
        & 92 & 100 & 100 & 100 & 94
        & 100 & 92 & 96 & 100 & 90
        & 4.88 & 96 & 2.872 \\
        
        & SAC & 0.1164
        & 80 & 80 & 74 & 70 & 64
        & 84 & 0 & 0 & 0 & 0
        & 78 & 100 & 76 & 70 & 52
        & 7.39 & 64 & 3.204 \\
        
        & PPO & 0.4517
        & 0 & 0 & 0 & 0 & 0
        & 0 & 0 & 0 & 0 & 0
        & 0 & 0 & 0 & 0 & 0
        & 2.37 & 0 & 5.350 \\
        
        & DynSyn-SAC & 0.3302
        & 80 & 72 & 70 & 68 & 60
        & 80 & 80 & 84 & 82 & 78
        & 92 & 84 & 78 & 60 & 26
        & 9.27 & 82 & 3.145 \\
        
        \cmidrule{1-21}
        
        \multirow{4}{*}{singlestand}
        & DepRL & 0.3977
        & 14 & 16 & 28 & 18 & 10
        & 24 & 30 & 8 & 0 & 0
        & 12 & 28 & 6 & 8 & 12
        & 2.10 & 14 & 2.854 \\
        
        & SAC & 0.0849
        & 0 & 0 & 0 & 0 & 0
        & 0 & 0 & 0 & 0 & 0
        & 0 & 0 & 0 & 0 & 0
        & 4.33 & 0 & 1.954 \\
        
        & PPO & 0.3981
        & 0 & 0 & 0 & 0 & 0
        & 0 & 0 & 0 & 0 & 0
        & 0 & 0 & 0 & 0 & 0
        & 1.84 & 0 & 5.436 \\
        
        & DynSyn-SAC & 0.3845
        & 24 & 28 & 24 & 20 & 18
        & 26 & 22 & 16 & 0 & 0
        & 22 & 18 & 8 & 0 & 0
        & 10.00 & 24 & 2.931 \\
        
        \cmidrule{1-21}
        
        \multirow{4}{*}{balance}
        & DepRL & 0.3775
        & 60 & 40 & 26 & 20 & 14
        & 56 & 0 & 0 & 0 & 0
        & 46 & 12 & 36 & 22 & 18
        & 8.86 & 26 & 2.822 \\
        
        & SAC & 0.1459
        & 0 & 0 & 0 & 0 & 0
        & 0 & 0 & 0 & 0 & 0
        & 0 & 0 & 0 & 0 & 0
        & 8.11 & 0 & 2.547 \\
        
        & PPO & 0.3859
        & 0 & 0 & 0 & 0 & 0
        & 0 & 0 & 0 & 0 & 0
        & 0 & 0 & 0 & 0 & 0
        & 1.25 & 0 & 4.861 \\
        
        & DynSyn-SAC & 0.2735
        & 40 & 38 & 20 & 16 & 14
        & 54 & 40 & 8 & 2 & 0
        & 52 & 8 & 0 & 0 & 0
        & 10.50 & 44 & 6.066 \\
        
        \bottomrule
    \end{tabular}
    }
\end{table*}

\begin{table*}[htbp]
    \vspace*{5pt}
    \centering
    \rmfamily 
    
    \caption{Benchmark Performance and Robustness Evaluation: Locomotion. Train-env. SR retains native training noise; robustness columns report separate perturbation sweeps (App.~\ref{sec:app-evaluation_details}).}
    \label{tab:benchmark_robustness}
    
    \setlength{\tabcolsep}{1.8pt}
    \scriptsize
    
    \resizebox{\linewidth}{!}{
    \begin{tabular}{lc c *{15}{c} ccc}
        \toprule
        \multirow{3}{*}{\textbf{Task}}
        & \multirow{3}{*}{\textbf{Algorithm}}
        & \textbf{Activation}
        & \multicolumn{15}{c}{\textbf{Robustness(\%)}}
        & \textbf{Max}
        & \textbf{Train-env.}
        & \textbf{$\log_{10}$smooth} \\

        & & \textbf{Cost}
        & \multicolumn{5}{c}{Action}
        & \multicolumn{5}{c}{Observation}
        & \multicolumn{5}{c}{Dynamic}
        & \textbf{Steps}($\times 10^7$)
        & \textbf{SR (\%)}
        & \textbf{(rad$^2$/s$^6$)} \\

        \cmidrule(lr){4-8}
        \cmidrule(lr){9-13}
        \cmidrule(lr){14-18}

        & & 
        & 0 & 0.05 & 0.1 & 0.15 & 0.2
        & 0 & 0.02 & 0.05 & 0.08 & 0.10
        & 0 & 0.05 & 0.1 & 0.15 & 0.2
        & & & \\

        \midrule

        \multirow{4}{*}{walk forward}
        & DepRL
        & 0.4206
        & 28 & 28 & 32 & 32 & 44
        & 34 & 32 & 46 & 24 & 30
        & 40 & 20 & 12 & 8 & 6
        & 3.32 & 64 & 3.151 \\

        & SAC
        & 0.1427
        & 0 & 0 & 0 & 0 & 0
        & 0 & 0 & 0 & 0 & 0
        & 0 & 0 & 0 & 0 & 0
        & 9.88 & 0 & 3.262 \\

        & PPO
        & 0.3940
        & 0 & 0 & 0 & 0 & 0
        & 0 & 0 & 0 & 0 & 0
        & 0 & 0 & 0 & 0 & 0
        & 2.42 & 0 & 5.558 \\

        & DynSyn-SAC
        & 0.2771
        & 82 & 88 & 68 & 60 & 48
        & 82 & 82 & 80 & 76 & 76
        & 82 & 76 & 76 & 64 & 8
        & 5.00 & 76 & 3.417 \\

        \cmidrule{1-21}

        \multirow{4}{*}{run}
        & DepRL
        & 0.3354
        & 30 & 22 & 32 & 36 & 42
        & 18 & 18 & 30 & 32 & 26
        & 32 & 16 & 12 & 8 & 2
        & 6.72 & 18 & 3.343 \\

        & SAC
        & 0.0940
        & 0 & 0 & 0 & 0 & 0
        & 0 & 0 & 0 & 0 & 0
        & 0 & 0 & 0 & 0 & 0
        & 2.13 & 0 & 2.854 \\

        & PPO
        & 0.3696
        & 0 & 0 & 0 & 0 & 0
        & 0 & 0 & 0 & 0 & 0
        & 0 & 0 & 0 & 0 & 0
        & 2.16 & 0 & 5.524 \\

        & DynSyn-SAC
        & 0.3221
        & 96 & 94 & 84 & 70 & 68
        & 96 & 94 & 90 & 86 & 80
        & 96 & 88 & 2 & 2 & 0
        & 10.00 & 96 & 3.435 \\

        \cmidrule{1-21}

        \multirow{4}{*}{walk turn}
        & DepRL
        & 0.3385
        & 46 & 60 & 62 & 66 & 68
        & 34 & 40 & 58 & 40 & 26
        & 48 & 46 & 40 & 36 & 36
        & 6.38 & 42 & 3.057 \\

        & SAC
        & 0.1571
        & 0 & 0 & 0 & 0 & 0
        & 0 & 0 & 0 & 0 & 0
        & 0 & 0 & 0 & 0 & 0
        & 0.99 & 0 & 2.610 \\

        & PPO
        & 0.3747
        & 0 & 0 & 0 & 0 & 0
        & 0 & 0 & 0 & 0 & 0
        & 0 & 0 & 0 & 0 & 0
        & 1.05 & 0 & 5.331 \\

        & DynSyn-SAC
        & 0.3112
        & 96 & 88 & 84 & 78 & 74
        & 94 & 90 & 90 & 88 & 84
        & 92 & 24 & 0 & 0 & 6
        & 8.00 & 86 & 3.248 \\

        \cmidrule{1-21}

        \multirow{4}{*}{jump}
        & DepRL
        & 0.3616
        & 8 & 24 & 20 & 12 & 10
        & 18 & 18 & 8 & 4 & 0
        & 14 & 10 & 12 & 6 & 0
        & 6.89 & 10 & 3.188 \\

        & SAC
        & 0.1342
        & 0 & 0 & 0 & 0 & 0
        & 0 & 0 & 0 & 0 & 0
        & 0 & 0 & 0 & 0 & 0
        & 3.92 & 0 & 2.853 \\

        & PPO
        & 0.3880
        & 0 & 0 & 0 & 0 & 0
        & 0 & 0 & 0 & 0 & 0
        & 0 & 0 & 0 & 0 & 0
        & 2.07 & 0 & 5.640 \\

        & DynSyn-SAC
        & 0.2700
        & 70 & 40 & 8 & 0 & 0
        & 62 & 54 & 50 & 48 & 42
        & 42 & 30 & 2 & 2 & 0
        & 8.00 & 54 & 3.449 \\

        \cmidrule{1-21}

        \multirow{4}{*}{sidestep}
        & DepRL
        & 0.3888
        & 40 & 38 & 62 & 56 & 32
        & 34 & 12 & 2 & 0 & 0
        & 24 & 22 & 18 & 10 & 6
        & 4.38 & 28 & 3.349 \\

        & SAC
        & 0.1045
        & 0 & 0 & 0 & 0 & 0
        & 0 & 0 & 0 & 0 & 0
        & 0 & 0 & 0 & 0 & 0
        & 0.000032 & 0 & 3.166 \\

        & PPO
        & 0.3729
        & 0 & 0 & 0 & 0 & 0
        & 0 & 0 & 0 & 0 & 0
        & 0 & 0 & 0 & 0 & 0
        & 4.92 & 0 & 5.408 \\

        & DynSyn-SAC
        & 0.3630
        & 74 & 66 & 54 & 42 & 30
        & 74 & 78 & 74 & 78 & 54
        & 64 & 72 & 60 & 40 & 40
        & 5.58 & 76 & 3.223 \\

        \cmidrule{1-21}

        \multirow{4}{*}{crawl}
        & DepRL
        & 0.3540
        & 100 & 100 & 100 & 100 & 98
        & 96 & 100 & 100 & 100 & 100
        & 100 & 100 & 98 & 100 & 100
        & 5.12 & 86 & 3.158 \\

        & SAC
        & 0.0338
        & 0 & 0 & 0 & 0 & 0
        & 0 & 0 & 0 & 0 & 0
        & 0 & 0 & 0 & 0 & 0
        & 2.87 & 0 & 2.516 \\

        & PPO
        & 0.4036
        & 0 & 0 & 0 & 0 & 0
        & 0 & 0 & 0 & 0 & 0
        & 0 & 0 & 0 & 0 & 0
        & 1.34 & 0 & 5.674 \\

        & DynSyn-SAC
        & 0.2766
        & 100 & 100 & 100 & 100 & 100
        & 100 & 100 & 100 & 100 & 100
        & 100 & 100 & 100 & 100 & 100
        & 10.00 & 100 & 3.444 \\

        \bottomrule
    \end{tabular}
    }
\end{table*}

\begin{table*}[htbp]
    \vspace*{5pt}
    \centering
    \rmfamily 
    
    \caption{Benchmark Performance and Robustness Evaluation: Interaction with Environment. Train-env. SR retains native training noise; robustness columns report separate perturbation sweeps (App.~\ref{sec:app-evaluation_details}).}
    \label{tab:benchmark_robustness_interaction}
    
    \setlength{\tabcolsep}{1.8pt}
    \scriptsize
    
    \resizebox{\linewidth}{!}{
    \begin{tabular}{lc c *{15}{c} ccc}
        \toprule
        \multirow{3}{*}{\textbf{Task}}
        & \multirow{3}{*}{\textbf{Algorithm}}
        & \textbf{Activation}
        & \multicolumn{15}{c}{\textbf{Robustness (\%)}}
        & \textbf{Max}
        & \textbf{Train-env.}
        & \textbf{$\log_{10}$smooth} \\
        
        & & \textbf{Cost}
        & \multicolumn{5}{c}{Action}
        & \multicolumn{5}{c}{Observation}
        & \multicolumn{5}{c}{Dynamic}
        & \textbf{Steps} ($\times 10^7$)
        & \textbf{SR (\%)}
        & \textbf{(rad$^2$/s$^6$)} \\
        
        \cmidrule(lr){4-8}
        \cmidrule(lr){9-13}
        \cmidrule(lr){14-18}
        
        & & 
        & 0 & 0.05 & 0.1 & 0.15 & 0.2
        & 0 & 0.02 & 0.05 & 0.08 & 0.10
        & 0 & 0.05 & 0.1 & 0.15 & 0.2
        & & & \\
        
        \midrule
        
        \multirow{4}{*}{stair}
        & DepRL & 0.3499
        & 52 & 54 & 38 & 42 & 40
        & 34 & 38 & 2 & 0 & 0
        & 32 & 40 & 24 & 14 & 4
        & 1.42 & 36 & 2.982 \\
        
        & SAC & 0.1993
        & 0 & 0 & 0 & 0 & 0
        & 0 & 0 & 0 & 0 & 0
        & 0 & 0 & 0 & 0 & 0
        & 6.98 & 0 & 2.459 \\
        
        & PPO & 0.3854
        & 0 & 0 & 0 & 0 & 0
        & 0 & 0 & 0 & 0 & 0
        & 0 & 0 & 0 & 0 & 0
        & 1.57 & 0 & 5.303 \\
        
        & DynSyn-SAC & 0.2759
        & 0 & 0 & 0 & 0 & 0
        & 0 & 0 & 0 & 0 & 0
        & 0 & 0 & 0 & 0 & 0
        & 5.92 & 0 & 2.854 \\
        
        \cmidrule{1-21}
        
        \multirow{4}{*}{catch}
        & DepRL & 0.3247
        & 80 & 80 & 76 & 76 & 84
        & 62 & 48 & 80 & 72 & 46
        & 78 & 40 & 42 & 38 & 56
        & 7.23 & 28 & 2.919 \\
        
        & SAC & 0.1384
        & 0 & 0 & 0 & 0 & 0
        & 0 & 0 & 0 & 0 & 0
        & 0 & 0 & 0 & 0 & 0
        & 1.37 & 0 & 2.899 \\
        
        & PPO & 0.3964
        & 0 & 0 & 0 & 0 & 0
        & 0 & 0 & 0 & 0 & 0
        & 0 & 0 & 0 & 0 & 0
        & 3.12 & 0 & 5.300 \\
        
        & DynSyn-SAC & 0.3109
        & 70 & 64 & 60 & 60 & 58
        & 60 & 40 & 34 & 26 & 22
        & 68 & 68 & 60 & 54 & 50
        & 7.38 & 66 & 3.134 \\
        
        \cmidrule{1-21}
        
        \multirow{4}{*}{hurdle}
        & DepRL & 0.3715
        & 48 & 48 & 56 & 50 & 40
        & 64 & 44 & 48 & 28 & 0
        & 62 & 44 & 40 & 32 & 20
        & 1.26 & 50 & 3.212 \\
        
        & SAC & 0.1852
        & 0 & 0 & 0 & 0 & 0
        & 0 & 0 & 0 & 0 & 0
        & 0 & 0 & 0 & 0 & 0
        & 4.12 & 0 & 2.816 \\
        
        & PPO & 0.3668
        & 0 & 0 & 0 & 0 & 0
        & 0 & 0 & 0 & 0 & 0
        & 0 & 0 & 0 & 0 & 0
        & 2.46 & 0 & 5.398 \\
        
        & DynSyn-SAC & 0.3717
        & 0 & 0 & 0 & 0 & 0
        & 0 & 0 & 0 & 0 & 0
        & 0 & 0 & 0 & 0 & 0
        & 5.50 & 0 & 7.102 \\
        
        \cmidrule{1-21}
        
        \multirow{4}{*}{slide}
        & DepRL & 0.3330
        & 100 & 22 & 22 & 16 & 12
        & 100 & 16 & 4 & 4 & 0
        & 100 & 8 & 14 & 4 & 6
        & 11.32 & 100 & 3.106 \\
        
        & SAC & 0.0925
        & 0 & 0 & 0 & 0 & 0
        & 0 & 0 & 0 & 0 & 0
        & 0 & 0 & 0 & 0 & 0
        & 4.66 & 0 & 2.752 \\
        
        & PPO & 0.3934
        & 0 & 0 & 0 & 0 & 0
        & 0 & 0 & 0 & 0 & 0
        & 0 & 0 & 0 & 0 & 0
        & 4.13 & 0 & 5.345 \\
        
        & DynSyn-SAC & 0.2649
        & 0 & 2 & 2 & 0 & 0
        & 0 & 0 & 0 & 0 & 0
        & 0 & 2 & 2 & 0 & 0
        & 3.83 & 0 & 2.864 \\
        
        \cmidrule{1-21}
        
        \multirow{4}{*}{steppingstones}
        & DepRL & 0.3606
        & 42 & 46 & 44 & 40 & 38
        & 46 & 42 & 10 & 20 & 2
        & 54 & 38 & 24 & 18 & 18
        & 8.32 & 58 & 2.985 \\
        
        & SAC & 0.1647
        & 0 & 0 & 0 & 0 & 0
        & 0 & 0 & 0 & 0 & 0
        & 0 & 0 & 0 & 0 & 0
        & 7.33 & 0 & 2.140 \\
        
        & PPO & 0.4020
        & 0 & 0 & 0 & 0 & 0
        & 0 & 0 & 0 & 0 & 0
        & 0 & 0 & 0 & 0 & 0
        & 1.27 & 0 & 5.211 \\
        
        & DynSyn-SAC & 0.3569
        & 0 & 0 & 0 & 0 & 0
        & 0 & 0 & 0 & 0 & 0
        & 0 & 0 & 0 & 0 & 0
        & 10.00 & 0 & 6.764 \\
        
        \cmidrule{1-21}
        
        \multirow{4}{*}{polewalk}
        & DepRL & 0.4080
        & 22 & 38 & 16 & 40 & 20
        & 42 & 26 & 6 & 0 & 0
        & 30 & 26 & 20 & 16 & 16
        & 5.14 & 12 & 2.876 \\
        
        & SAC & 0.1504
        & 0 & 0 & 0 & 0 & 0
        & 0 & 0 & 0 & 0 & 0
        & 0 & 0 & 0 & 0 & 0
        & 6.10 & 0 & 1.724 \\
        
        & PPO & 0.4274
        & 0 & 0 & 0 & 0 & 0
        & 0 & 0 & 0 & 0 & 0
        & 0 & 0 & 0 & 0 & 0
        & 2.87 & 0 & 5.377 \\
        
        & DynSyn-SAC & 0.2909
        & 98 & 96 & 98 & 92 & 90
        & 98 & 0 & 0 & 0 & 0
        & 96 & 94 & 94 & 76 & 16
        & 9.00 & 88 & 3.242 \\
        
        \cmidrule{1-21}
        
        \multirow{4}{*}{reach}
        & DepRL & 0.3410
        & 84 & 86 & 84 & 94 & 82
        & 94 & 82 & 90 & 62 & 34
        & 92 & 88 & 82 & 82 & 74
        & 9.34 & 84 & 2.975 \\
        
        & SAC & 0.1705
        & 0 & 0 & 0 & 0 & 0
        & 0 & 0 & 0 & 0 & 0
        & 0 & 0 & 0 & 0 & 0
        & 5.85 & 0 & 2.677 \\
        
        & PPO & 0.3965
        & 0 & 0 & 0 & 0 & 0
        & 0 & 0 & 0 & 0 & 0
        & 0 & 0 & 0 & 0 & 0
        & 2.87 & 0 & 5.152 \\
        
        & DynSyn-SAC & 0.2556
        & 4 & 14 & 0 & 4 & 4
        & 2 & 6 & 6 & 2 & 4
        & 10 & 8 & 6 & 4 & 0
        & 2.10 & 0 & 2.840 \\
        
        \cmidrule{1-21}
        
        \multirow{4}{*}{walkandsit}
        & DepRL & 0.3648
        & 30 & 26 & 28 & 14 & 10
        & 28 & 36 & 32 & 18 & 10
        & 22 & 28 & 22 & 14 & 12
        & 2.68 & 16 & 3.365 \\
        
        & SAC & 0.1180
        & 0 & 0 & 0 & 0 & 0
        & 0 & 0 & 0 & 0 & 0
        & 0 & 0 & 0 & 0 & 0
        & 6.54 & 0 & 2.410 \\
        
        & PPO & 0.3951
        & 0 & 0 & 0 & 0 & 0
        & 0 & 0 & 0 & 0 & 0
        & 0 & 0 & 0 & 0 & 0
        & 3.03 & 0 & 5.636 \\
        
        & DynSyn-SAC & 0.3391
        & 48 & 58 & 66 & 60 & 52
        & 48 & 54 & 40 & 36 & 34
        & 42 & 48 & 40 & 36 & 18
        & 0.45 & 58 & 7.145 \\
        
        \cmidrule{1-21}
        
        \multirow{4}{*}{chinup}
        & DepRL & 0.3998
        & 50 & 59 & 50 & 62 & 54
        & 46 & 72 & 64 & 62 & 56
        & 70 & 62 & 64 & 60 & 52
        & 5.46 & 32 & 2.862 \\
        
        & SAC & 0.4618
        & 64 & 86 & 76 & 66 & 58
        & 64 & 50 & 74 & 72 & 66
        & 70 & 70 & 68 & 64 & 60
        & 1.37 & 28 & 2.703 \\
        
        & PPO & 0.3620
        & 62 & 74 & 72 & 66 & 60
        & 64 & 66 & 82 & 80 & 72
        & 74 & 66 & 56 & 60 & 76
        & 1.97 & 74 & 4.710 \\
        
        & DynSyn-SAC & 0.2797
        & 0 & 0 & 0 & 0 & 0
        & 0 & 0 & 0 & 0 & 0
        & 0 & 0 & 0 & 0 & 0
        & 10.00 & 0 & 6.989 \\
        
        \cmidrule{1-21}
        
        \multirow{4}{*}{open door}
        & DepRL & 0.3304
        & 96 & 100 & 100 & 100 & 98
        & 100 & 94 & 74 & 62 & 42
        & 100 & 96 & 96 & 78 & 66
        & 4.71 & 98 & 3.030 \\
        
        & SAC & 0.0962
        & 0 & 0 & 0 & 0 & 0
        & 0 & 0 & 0 & 0 & 0
        & 0 & 0 & 0 & 0 & 0
        & 1.97 & 0 & 2.836 \\
        
        & PPO & 0.4112
        & 0 & 0 & 0 & 0 & 0
        & 0 & 0 & 0 & 0 & 0
        & 0 & 0 & 0 & 0 & 0
        & 1.77 & 0 & 5.327 \\
        
        & DynSyn-SAC & 0.2784
        & 100 & 98 & 98 & 96 & 92
        & 100 & 100 & 100 & 98 & 98
        & 100 & 100 & 14 & 4 & 0
        & 9.50 & 100 & 6.461 \\
        
        \bottomrule
    \end{tabular}
    }
\end{table*}

\subsection{Detailed Descriptions of Reward-Based RL Baselines}
\label{appendix: pure_rl_baselines}

In Section~\ref{sec: benchmark}, we categorized several standard and exploration-enhanced reinforcement learning methods under the reward-based RL methods. The detailed descriptions of these individual baselines are provided below:

\begin{itemize}
    \item \textbf{PPO (Proximal Policy Optimization).} This~\cite{schulman2017proximal} serves as the quintessential on-policy actor--critic baseline for continuous-control tasks. By employing a clipped surrogate objective function, PPO ensures conservative policy updates, preventing destructively large parameter shifts that frequently occur in highly nonlinear biomechanical simulations. While it offers excellent training stability and reliable convergence, its purely reward-driven, unstructured exploration often proves sample-inefficient when navigating the high-dimensional, highly redundant muscle excitation spaces inherent to full-body musculoskeletal humanoids.

    \item \textbf{SAC (Soft Actor-Critic).} This~\cite{haarnoja2018soft} is utilized as the primary off-policy baseline, distinguished by its maximum-entropy reinforcement learning framework. By jointly maximizing expected return and policy entropy, SAC explicitly encourages broad and robust exploration of the action space. This stochastic exploration strategy is particularly advantageous for musculoskeletal control: the entropy regularization prevents premature convergence to sub-optimal local minima—a common pathology in muscle-driven systems where multiple distinct muscle activation patterns can yield identical kinematic joint torques. Furthermore, its off-policy nature and replay buffer significantly improve sample efficiency over on-policy methods.

    \item \textbf{DepRL.} This~\cite{schumacher2022dep} represents a specialized, exploration-enhanced reinforcement learning baseline tailored for musculoskeletal systems. Traditional independent Gaussian action noise often fails to activate muscles in a biologically coordinated manner, typically resulting in conflicting co-contractions and ``stiff'' joint behaviors. DepRL addresses this by augmenting the RL policy with Differential Extrinsic Plasticity (DEP)---an unsupervised, self-organizing exploration mechanism. Instead of injecting blind stochastic noise, the DEP controller dynamically updates a sensorimotor mapping matrix based on the temporal correlations between recent motor commands and the derivatives of proprioceptive feedback. This generates highly structured, mechanically grounded exploratory actions that intrinsically leverage the agent's physical embodiment. By superimposing these synergistic exploration patterns onto the RL policy outputs, DepRL effectively mitigates stiff co-contractions and significantly accelerates the discovery of functional locomotion gaits in highly overactuated models.

    \item \textbf{DynSyn-SAC.} This~\cite{he2024dynsyn} is a state-of-the-art action-representation algorithm designed specifically for efficient learning and control in overactuated embodied systems. Built upon the maximum-entropy SAC framework, DynSyn-SAC introduces a dynamical synergistic structure directly into the network architecture. Specifically, it groups functionally related actuators and processes state observations into a lower-dimensional latent representation. A dedicated \textit{DynSyn Layer} then maps this latent output to full-body muscle activations using group-shared synergistic weights. To ensure stable exploration during the early phases of learning, the algorithm employs a linear scheduling mechanism that dynamically bounds and scales the amplitude of these synergistic weights over the course of training. This hierarchical approach allows the policy to effectively predict group-level synergistic actions alongside state-dependent, muscle-level fine-tuning, thereby significantly accelerating sample efficiency in high-dimensional musculoskeletal models.
\end{itemize}

\subsubsection{Training and Checkpoint Protocol}
The released benchmark training entry points and per-task configuration files specify the settings summarized in Table~\ref{tab:reward_baseline_training}. PPO and SAC use eight parallel training environments and two evaluation environments; DepRL uses $4\times4=16$ training environments; DynSyn-SAC uses 64 training environments. All released configurations use seed 0. ``Update/data'' below reports the configured optimizer-update cadence relative to newly collected transitions, while the PPO entry instead reports its on-policy sample-reuse epochs. Command-line overrides are supported by the released code; the table records the released defaults/configurations rather than unspecified library settings.

\begin{table*}[t]
    \centering
    \caption{Released training and checkpoint protocol for the four reward-based baselines. MLP widths list hidden layers for the actor and value/critic networks. ``Det.'' denotes deterministic evaluation episodes.}
    \label{tab:reward_baseline_training}
    \scriptsize
    \setlength{\tabcolsep}{3.2pt}
    \begin{tabularx}{\textwidth}{>{\raggedright\arraybackslash}p{0.065\textwidth} >{\raggedright\arraybackslash}p{0.135\textwidth} >{\raggedright\arraybackslash}p{0.105\textwidth} >{\raggedright\arraybackslash}p{0.06\textwidth} X >{\raggedright\arraybackslash}p{0.075\textwidth} X}
        \toprule
        \textbf{Method} & \textbf{Actor / critic MLP} & \textbf{Learning rate} & \textbf{Batch} & \textbf{Collection, rollout, and update/data cadence} & \textbf{Budget} & \textbf{Checkpoint used for reporting} \\
        \midrule
        PPO & $2\times1024$ / $2\times1024$ & $3\times10^{-4}$ & 256 & 2,048 steps per environment and eight environments ($16{,}384$ transitions per rollout); 10 epochs, i.e., 10 passes over each rollout. & $2\times10^8$ steps & Highest mean evaluation return; 10 det. episodes every $5\times10^5$ training steps. \\
        SAC & $2\times1024$ / twin $2\times1024$ Q networks & $1\times10^{-4}$ & 256 & Continuous off-policy collection; one vectorized step (eight new transitions) followed by one gradient step. The configured \texttt{gradient\_steps}/\texttt{train\_freq} ratio is 1. & $2\times10^8$ steps & Highest mean evaluation return; 10 det. episodes every $2\times10^5$ training steps. \\
        DepRL & $2\times1024$ / $2\times1024$ & Actor $5\times10^{-5}$; critic $8\times10^{-5}$; dual $2\times10^{-3}$ & 256 & Continuous replay with 3-step returns; after $10^5$ warm-up transitions, 30 minibatch updates per 1,000 new transitions (0.03 update iterations per transition). & $1\times10^8$ steps & Last step-numbered checkpoint (the completed run ends at $10^8$); saved every $2\times10^6$ steps. A best-return checkpoint is also written, but is not the evaluator default. \\
        DynSyn-SAC & $3\times1024$ / twin $3\times1024$ Q networks, plus DynSyn layer & Linear $5\times10^{-4}$ to $1.25\times10^{-4}$ & 256\textsuperscript{*} & Continuous off-policy collection; one vectorized step (64 new transitions) followed by one gradient step. The configured \texttt{gradient\_steps}/\texttt{train\_freq} ratio is 1. & $5\times10^7$ steps & Highest mean evaluation return; 3 det. episodes every $5\times10^6$ training steps. \\
        \bottomrule
    \end{tabularx}
    \vspace{1mm}
    \parbox{0.98\textwidth}{\scriptsize \textsuperscript{*}DynSyn-SAC does not override \texttt{batch\_size}; 256 is the Stable-Baselines3 SAC default in the declared 2.x dependency range. It likewise inherits a one-step \texttt{train\_freq}; the released configuration explicitly sets one gradient step.}
\end{table*}

For off-policy methods, a single gradient step after one vectorized collection step should not be read as one update per scalar transition: SAC collects eight transitions and DynSyn-SAC collects 64 transitions in that step. DepRL's 30/1,000 cadence is defined directly in scalar environment-step counts. The success-rate scripts then evaluate the selected checkpoint for 50 episodes per task by default, using the shared criteria in Table~\ref{tab:success_criteria} rather than selecting checkpoints by success rate.

\subsection{Detailed Muscle Activation Analysis: 416 vs. 700 Muscle Models}
\label{appendix: muscle_heatmaps}

\begin{figure*}[!htbp]
    \centering
    \includegraphics[width=\linewidth]{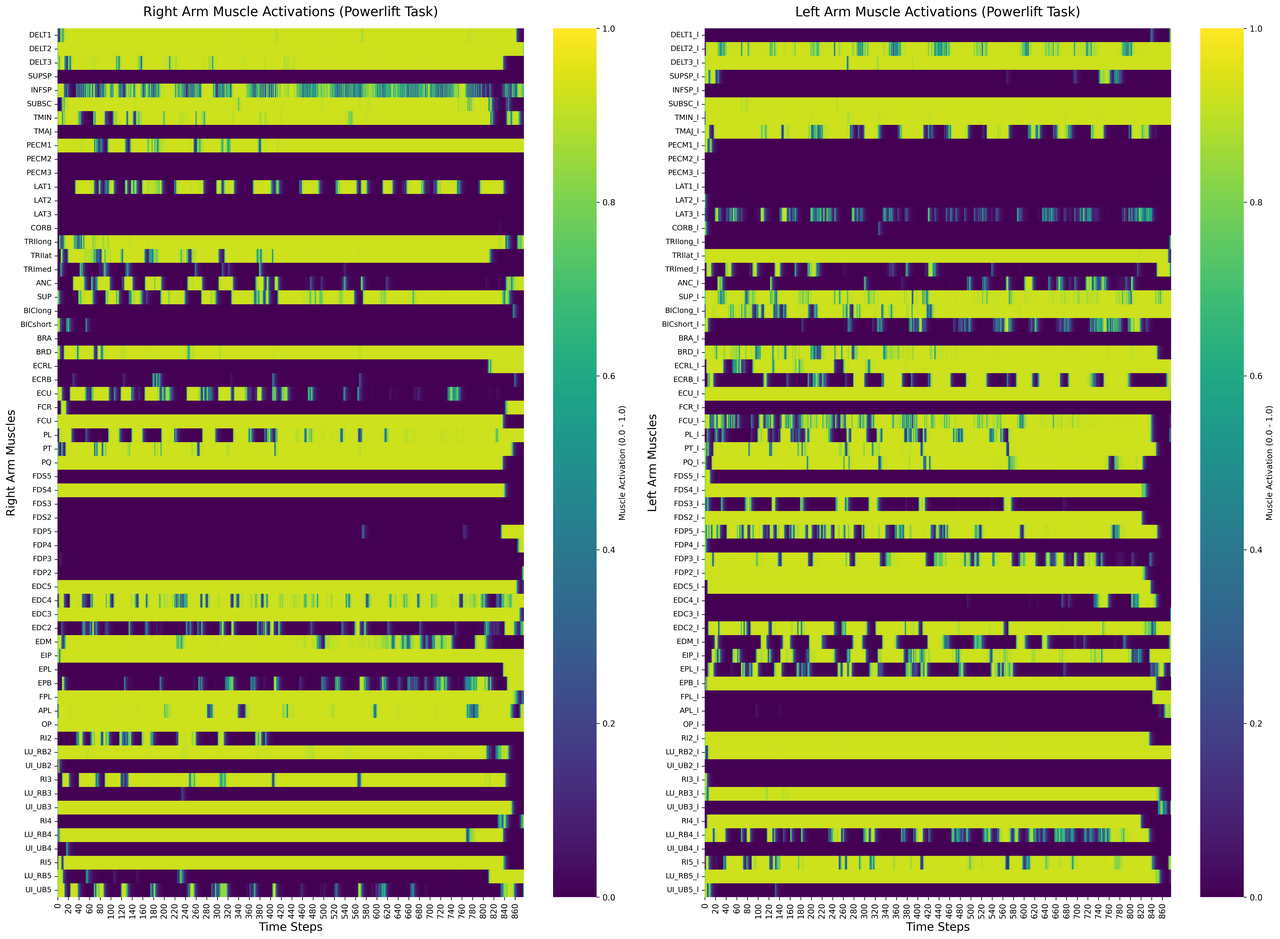}
    \caption{\textbf{Upper limb muscle activation heatmap for the 416-muscle model (Powerlift).} The heatmap displays the intense, sustained activations of the available arm and shoulder musculature required to stabilize the barbell overhead. Note that this model lacks the detailed deep axial stabilizers of the neck and spine.}
    \label{fig:heatmap_416}
\end{figure*}

\begin{figure*}[!htbp]
    \centering
    \includegraphics[width=0.88\linewidth,height=0.83\textheight,keepaspectratio]{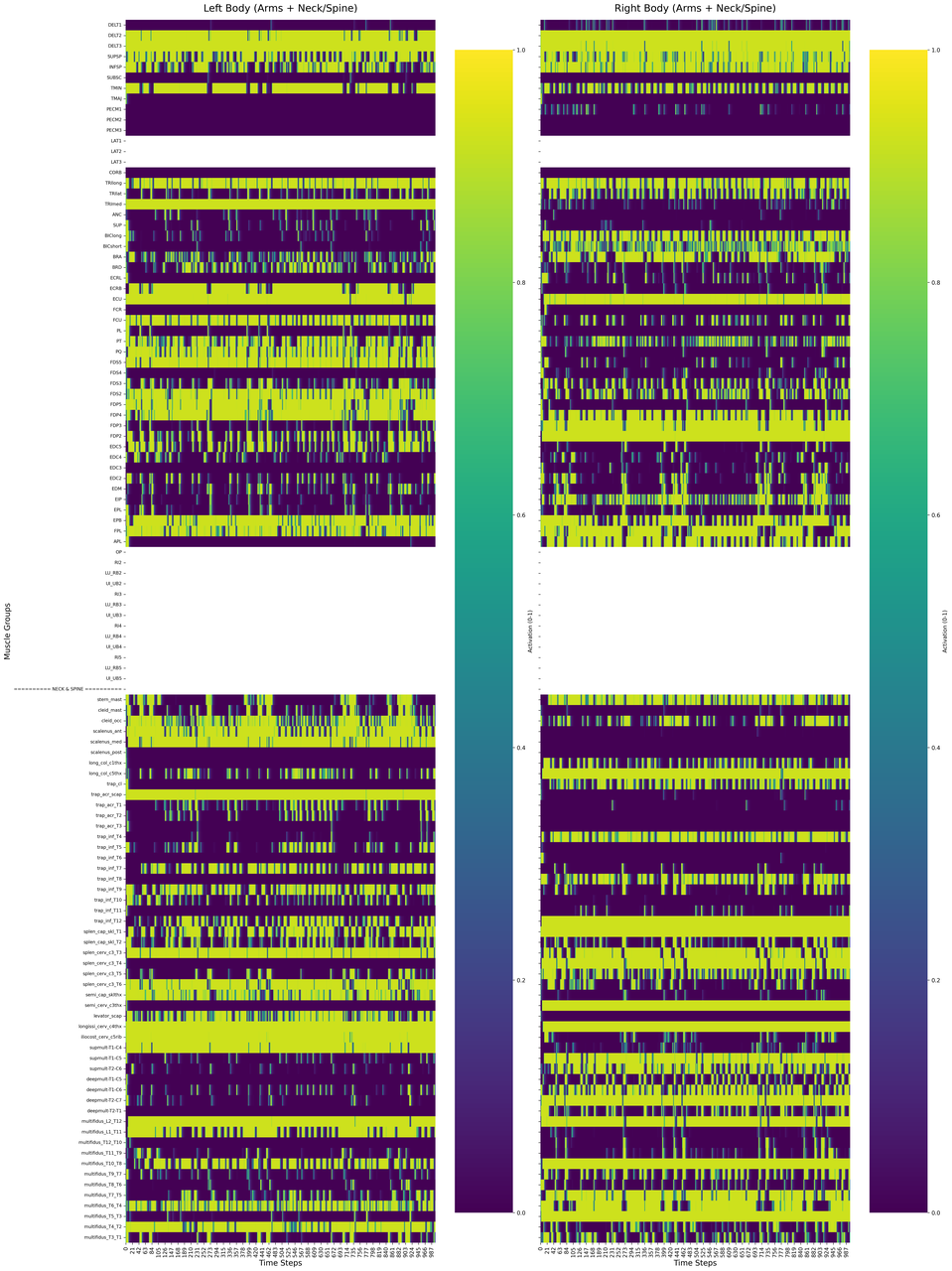}
    \caption{\textbf{Expanded muscle activation heatmap for the 700-muscle model (Powerlift).} The top section includes the upper limb muscles, where the distinct "blank gaps" represent the fine-grained scapular and arm muscles exclusive to the 700-model. The bottom section explicitly details the heavy recruitment of the newly available neck and spine axial stabilizers (e.g., \textit{multifidus}, \textit{splenius}).}
    \label{fig:heatmap_700}
\end{figure*}

In the main text, we discussed how anatomical resolution changes the recruitment groups available for analysis during the \textbf{powerlift} task. To provide deeper insight into these internal control differences, we present the time-step resolved muscle activation heatmaps for both the 416-muscle (Fig.~\ref{fig:heatmap_416}) and the highly detailed 700-muscle (Fig.~\ref{fig:heatmap_700}) models.

\textbf{Biomechanical Analysis.}
The heatmaps describe model-specific recruitment over the recorded powerlift motions. The 700-muscle model includes additional neck, deep-trunk, and deep-scapular groups, expanding the recruitment pathways available for inspection. Differences in upper-limb and axial activation show that anatomical resolution affects muscle-use diagnostics. These observations do not establish improved human similarity, clinically correct force transfer, or a causal explanation for the observed recruitment differences.

\subsection{Agentic Reward Tuning}
\label{sec:app-agentic_reward_tuning}

In our framework, we introduce \textit{Agentic Reward Tuning}, a training mechanism that autonomously adjusts reward weights during reinforcement learning according to the agent's observed performance. Instead of relying on a fully fixed reward design, the reward composition is periodically refined by an external reasoning module based on recent gait quality, balance stability, and task-progress statistics.

Formally, let $\mathbf{r}_t$ denote the vector of sub-rewards at timestep $t$, and let $\mathbf{w}_t$ denote the corresponding reward weights. The dense reward is computed as:
\begin{equation}
R_t = \sum_i w_{i,t} r_{i,t}.
\end{equation}

During training, the reward weights are not treated as static hyperparameters. Every $K$ environment steps, the system aggregates recent performance statistics:
\begin{equation}
\mathbf{s}_{\mathrm{stats}} =
\mathrm{Aggregate}
\left(
\{\mathbf{o}_{\tau}, \mathbf{r}_{\tau}\}_{\tau=t-K}^{t}
\right),
\end{equation}
where $\mathbf{s}_{\mathrm{stats}}$ contains task-relevant indicators such as forward velocity, torso pitch, posture stability, and tracking quality.

The reasoning module then updates the reward weights according to the current reward configuration and the collected statistics:
\begin{equation}
\mathbf{w}_{t+1}
=
\mathrm{Reasoner}
\left(
\mathbf{w}_{t}, \mathbf{s}_{\mathrm{stats}}
\right),
\end{equation}
In the original Agentic-DepRL setting, updates occur every 20,000 environment steps using mean forward speed and torso pitch. Each nonzero coefficient changes by at most 20\% per update; this relative bound does not prevent the unnormalized magnitudes from growing over time.

For example, if the agent exhibits insufficient forward speed, the tuning module increases the velocity-tracking weight. If the agent develops a hunched posture, the weights of uprightness and anti-hunching rewards are increased. In this way, Agentic Reward Tuning acts as an adaptive reward-shaping mechanism that continuously emphasizes the currently deficient aspects of the learned behavior.

In implementation, the tuning process is triggered periodically during environment rollout. Recent statistics, such as average forward speed and torso pitch, are cached over a fixed interval. A reasoning module analyzes these statistics and returns an updated reward-weight dictionary. To support parallel training, the updated weights are written to a shared file and synchronized across environments. A file-lock mechanism is used to prevent multiple workers from simultaneously calling the reasoning module, and fallback exception handling ensures that training continues even when the external call fails.

The matched reasoner comparison uses DeepSeek-V4-Flash~\cite{deepseekai2026deepseekv4} and GPT-6 Astra~\cite{openai2026gpt6astra} with the same DepRL learner, architecture, training budget, and evaluation, and the identical constraint $\|w\|_1=40$. Table~\ref{tab:reasoner_comparison} reports survival, tracking RMSE, and activation. This constrained comparison is distinct from the original unnormalized coefficient-update study. The reported point estimates show metric-dependent trade-offs rather than a uniform benefit from reward adaptation.

\subsection{Direct GPT-6 Control: Experimental Details}
\label{sec:app-gpt6_direct_details}

The aggregate outcomes of the direct-action diagnostic are reported in
Table~\ref{tab:gpt_direct_control}. Here we specify the controller boundary,
state serialization, rollout procedure, and metric computation used to obtain
those values.

\textbf{Controller-visible state.}
At decision index $k$, GPT-6 Astra was allowed to read only two JSON files:
static environment metadata and the current serialized
policy observation. The metadata records the
task, actuator names and ordering, action bounds, observation layout, simulator
timesteps, motion-prior metadata, and decision budget. The observation file is a
lossless restructuring of the 2,418-dimensional policy observation; flattening
the serialized fields exactly reconstructs the original \texttt{float32} vector.
Its composition is given in Table~\ref{tab:gpt6_observation_layout}.

\begin{table*}[t]
    \centering
    \caption{Controller-visible observation supplied at each GPT-6 decision.
    Dimensions are identical for Walk, Run, and Stairs.}
    \label{tab:gpt6_observation_layout}
    \begin{tabular}{lr}
        \toprule
        Observation block & Dimension \\
        \midrule
        Root pose without global $x$--$y$ & 5 \\
        Joint positions & 82 \\
        Root velocity & 6 \\
        Joint velocities & 82 \\
        Muscle state ($354\times$ length, velocity, force, excitation, activation) & 1,770 \\
        Foot/toe touch sensors & 4 \\
        Goal and reference features & 469 \\
        \midrule
        Total & 2,418 \\
        \bottomrule
    \end{tabular}
\end{table*}

The 469-dimensional goal block contains current relative positions, angles, and
velocities for 16 tracked sites; current reference-site positions; four
look-ahead reference blocks containing root-position delta, root-velocity delta,
and site positions; and the motion phase. Before choosing an action, the
controller could not inspect the saved trajectory, privileged diagnostics,
rendered frames, reference-controller outputs, future observations, or future
reference states beyond these policy-visible look-ahead features. Thus, the GPT
interface exposed the structured information already present in the policy
observation rather than simulator-internal state.

\textbf{Action contract.}
Each response was stored as a bare JSON array whose index order follows the
MuJoCo actuator ordering recorded in \texttt{env\_info.json}. The runner required
exactly 354 numeric, finite values, rejected Boolean entries, and checked every
component against the environment action bound $[-1,1]$. Accepted arrays were
cast to \texttt{float32} and applied without clipping, smoothing, interpolation,
optimizer refinement, baseline-action addition, or other post-processing. The
direct environment was instantiated with its base-policy callback disabled, so
the accepted array was the complete muscle-excitation command.

\textbf{Pause, replay, and stepping protocol.}
The interaction was implemented as a file-mediated paused session. A session
first reset the environment with seed~0 and exported decision~0. After one action
file was submitted, the runner validated it, rebuilt the same seeded environment,
deterministically replayed all previously committed actions, and applied the new
action until its simulation-time deadline. If the rollout remained valid, the
runner exported exactly one new observation and exited again. This design kept
MuJoCo paused while GPT-6 selected the next command and made each decision
recoverable from the manifest and committed action sequence.

The physics timestep was 0.001~s and ten physics substeps formed one 0.01-s
native control step. An accepted action was held with zero-order hold until
simulation time $0.2(k+1)$, corresponding to 20 native control steps. Termination,
truncation, and the fall gate were evaluated after every native step rather than
only at GPT decision boundaries. Observation-export and action-receipt timestamps
were retained for auditing, but wall-clock delay was excluded from the simulated
horizon and was not treated as model inference latency.

\textbf{Fall gate and analyzed prefix.}
Privileged quantities used only for stopping and offline analysis were never
written to the controller-visible observation. Let $h$ denote root height and let
$q=(w,x,y,z)$ be the normalized root quaternion. Root tilt was computed from
\begin{equation}
    \theta = \cos^{-1}\!\left(\operatorname{clip}
    \left(1-2(x^2+y^2),-1,1\right)\right).
\end{equation}
A sample was marked fallen when the generalized position or velocity contained a
non-finite value, $h<0.6$~m, or $\theta>70^{\circ}$. Once marked, the current hold
was stopped immediately. Metrics and videos use the inclusive prefix ending at
the first fallen sample. The immutable Walk audit log was produced before this
stop gate was repaired and therefore contains states after its first fallen sample
at 0.75~s; these post-fall states are excluded from every value and visualization
reported in the paper.

\textbf{Metric computation.}
At native step $t$, site-tracking error was calculated over the 16 tracked sites
as
\begin{equation}
    e_t = \sqrt{\frac{1}{3N}
    \sum_{j=1}^{N}\left\|\mathbf{x}_{j,t}-
    \mathbf{x}^{\mathrm{ref}}_{j,t}\right\|_2^2},
    \qquad N=16.
\end{equation}
The reported tracking RMSE is
$\sqrt{T^{-1}\sum_{t=1}^{T}e_t^2}$ over the first-fall-truncated prefix, including
all native-control samples rather than only GPT decision points.

The first-divergence timestamp uses a data-derived reference instead of a manually
selected task threshold. For each task, a reference trajectory was generated with
the official \texttt{mm-fullbody-base} MuscleMimic checkpoint and zero residual
correction. Let $e_t^{\mathrm{GPT}}$ be the GPT tracking error,
$\widetilde{e}_t^{\mathrm{ref}}$ the reference error interpolated at the same
simulation time, and $Q_{0.95}^{\mathrm{ref}}$ the 95th percentile of reference
errors. Divergence is the first sample of the first five-sample run satisfying
\begin{equation}
    e_t^{\mathrm{GPT}} > \widetilde{e}_t^{\mathrm{ref}}
    \quad\text{and}\quad
    e_t^{\mathrm{GPT}} > Q_{0.95}^{\mathrm{ref}}.
\end{equation}
Five native samples correspond to 0.05~s. If this sustained condition is not met
before the analyzed trajectory ends, the timestamp is reported as N/A. The
reference rollout is used here only to define the divergence criterion; its
performance is not included as an additional result in the direct-control table.

\subsection{Latent-Action RL}
\label{sec:app-latent_action_rl}

The \textit{Latent-Action RL} framework integrates expert-informed action filtering with reinforcement learning to guide musculoskeletal humanoid control. In the hard mode used in this work, the policy is not allowed to directly execute arbitrary high-dimensional muscle actions. Instead, each policy action is first projected through a pretrained anatomical encoder-decoder module, and the final executable action is constrained to remain within a bounded neighborhood of the expert-informed action. This design reduces unstable exploration in the high-dimensional muscle action space while preserving the adaptability of reinforcement learning.

\textbf{1. Expert Teacher Pretraining:}  
An expert teacher model is first trained using behavior cloning on high-quality demonstration data. The model adopts an encoder-decoder architecture:
\begin{itemize}
    \item \textbf{Encoder:} maps the expert muscle action $a_t$ to a low-dimensional latent intent $z_t$, which captures the underlying control intention.
    \item \textbf{Decoder (Full Anatomical Transformer):} reconstructs the muscle action conditioned on $z_t$ and the current musculoskeletal state, including muscle priors, muscle states, and joint moment information.
\end{itemize}

Formally, for a batch of expert actions $A$ and associated anatomical features $(P, S, M)$, the reconstructed action is given by:
\begin{equation}
\hat{A} = \mathrm{Decoder}(\mathrm{Encoder}(A), P, S, M),
\end{equation}
where $P$ denotes muscle-level physical priors, $S$ denotes muscle state features such as muscle length, velocity, force, and activation, and $M$ denotes the muscle-joint moment matrix. The behavior cloning objective is:
\begin{equation}
\mathcal{L}_{\mathrm{BC}} =
\frac{1}{N}
\sum_i
\left\|
\hat{A}_i - \frac{A_i + 1}{2}
\right\|^2 .
\end{equation}
Here, the transformation $(A_i+1)/2$ maps the original normalized action range $[-1,1]$ to the muscle activation range $[0,1]$.

\textbf{2. Hard-mode Action Filtering in RL:}  
During reinforcement learning, the policy outputs a raw action $a_t^{\mathrm{RL}} \in [-1,1]^M$. This action is first normalized to the muscle activation range:
\begin{equation}
a_{t,01}^{\mathrm{RL}} = \frac{a_t^{\mathrm{RL}} + 1}{2}.
\end{equation}

The pretrained encoder and decoder then generate an expert-informed filtered action:
\begin{equation}
\hat{a}_t =
\mathrm{Decoder}
\left(
\mathrm{Encoder}(a_t^{\mathrm{RL}}),
P_t, S_t, M_t
\right),
\end{equation}
where $\hat{a}_t \in [0,1]^M$ represents the anatomical action prior predicted by the transformer decoder.

In hard mode, the final executable action is clipped within a fixed-radius action tube around the expert-informed action:
\begin{equation}
a_{t,01}^{\mathrm{env}}
=
\mathrm{clip}
\left(
a_{t,01}^{\mathrm{RL}},
\hat{a}_t - \delta,
\hat{a}_t + \delta
\right),
\end{equation}
where $\delta$ is the tube radius controlling the maximum allowed deviation from the anatomical prior. The action is then mapped back to the environment action range:
\begin{equation}
a_t^{\mathrm{env}}
=
2a_{t,01}^{\mathrm{env}} - 1 .
\end{equation}

This hard constraint prevents the policy from producing physiologically implausible muscle activations while still allowing local exploration around the expert action manifold.

\textbf{3. Deviation Penalty:}  
To further discourage unnecessary deviation from the expert-informed action, the hard mode introduces a penalty based on the violation between the original RL action and the clipped safe action:
\begin{equation}
\mathcal{L}_{\mathrm{dev}}
=
\left\|
a_{t,01}^{\mathrm{RL}} - a_{t,01}^{\mathrm{env}}
\right\|^2 .
\end{equation}

The environment reward is modified as:
\begin{equation}
R_t'
=
R_t
-
\lambda_t
\mathcal{L}_{\mathrm{dev}},
\end{equation}
where $\lambda_t$ is a decaying coefficient. At the early stage of training, the deviation penalty strongly guides the policy toward the expert action manifold. As training progresses, the penalty gradually decreases, allowing the policy to adapt more freely to task-specific objectives.

\textbf{4. Transformer-based Muscle Attention:}  
The Full Anatomical Transformer models inter-muscle dependencies through stacked Muscle Attention Blocks. Each block consists of multi-head self-attention, layer normalization, feed-forward transformations, and residual connections. By attending across muscles, the decoder captures global muscle synergy patterns and joint-level coordination, which are difficult to model using independent muscle-wise mappings.

\textbf{5. Overall Principle:}  
In hard mode, Latent-Action RL can be viewed as reinforcement learning inside an expert-guided anatomical action tube. The pretrained encoder-decoder provides a biologically plausible action manifold, while reinforcement learning performs bounded exploration around this manifold. This mechanism reduces the difficulty of high-dimensional muscle control, suppresses unstable or energy-wasting actions, and improves the stability and physiological plausibility of learned humanoid behaviors.

\subsection{EMG Data Processing and Comparative Analysis}
\label{sec:app-emg_data_analysis}

This section describes the processing of simulated activations and human EMG envelopes. Human references are Gait120 for walking and stairs~\cite{boo2025comprehensive} and the 3-m/s running recordings of Van Hooren and Meijer~\cite{vanhooren2024running}.

\textbf{1. Data Normalization and Gait Cycle Alignment}  
EMG signals are first normalized to the gait cycle. Each signal is resampled to 101 points per cycle using cubic interpolation:
\begin{equation}
\tilde{x} = \text{interp1d}(x_\text{old}, x_\text{raw}, \text{kind='cubic'})(x_\text{new}),
\end{equation}
where $x_\text{old}$ represents the original sample points and $x_\text{new}$ is the normalized gait percentage from 0\% to 100\%.  
Signals are then min-max normalized to $[0,1]$:
\begin{equation}
x_\text{norm} = \frac{x - x_\text{min}}{x_\text{max} - x_\text{min} + \epsilon},
\end{equation}
with $\epsilon$ a small constant to prevent division by zero.

\textbf{2. Multi-Cycle Averaging}  
For EMG recordings with multiple gait cycles, individual cycles are extracted by detecting peaks corresponding to gait events (e.g., heel strikes). Each cycle is normalized and resampled, then averaged across cycles to produce a mean activation profile. The standard deviation across cycles is also computed and normalized by the peak-to-peak amplitude to quantify variability:
\begin{equation}
\sigma_\text{norm} = \frac{\text{std}(x_\text{cycles})}{x_\text{max} - x_\text{min} + \epsilon}.
\end{equation}

\textbf{3. Simulation-Human Alignment}  
Simulated EMG signals are aligned with the human reference by cyclically shifting the normalized simulation profile to maximize the Pearson correlation coefficient $r$:
\begin{equation}
r = \frac{\text{cov}(\tilde{x}_\text{sim}, \tilde{x}_\text{human})}{\sigma_\text{sim} \sigma_\text{human}}.
\end{equation}
The shift that maximizes $r$ is applied to both the mean and standard deviation of the simulated signal, ensuring phase alignment with the human gait cycle.

\textbf{4. Comparative Analysis}  
For each target muscle, $r$ quantifies phase-optimized envelope-shape agreement. Normalization and cyclic shifting remove absolute amplitude and phase information, so this correlation does not establish absolute timing or amplitude fidelity. Across-cycle standard deviation describes variability around the mean profile. Task success and EMG agreement are assessed separately.
\subsection{Muscle Recruitment Categorization}
\label{sec:app-muscle_categorization}

To analyze whole-body muscle recruitment patterns, we categorize the 416 muscle actuators into anatomically and functionally meaningful muscle families. This grouping converts the high-dimensional muscle activation vector into an interpretable recruitment profile, allowing us to quantify how different body regions contribute to each learned behavior.

For each trained policy, we perform a rollout in the corresponding musculoskeletal environment and record the activation of each muscle actuator at every simulation timestep. Let $a_{m,t}$ denote the activation of the $m$-th muscle at timestep $t$, where $m \in \{1,\dots,M\}$ and $M=416$. The mean activation of each muscle over one rollout is computed as:
\begin{equation}
\bar{a}_m =
\frac{1}{T}
\sum_{t=1}^{T}
a_{m,t},
\end{equation}
where $T$ denotes the number of simulated timesteps before termination or the predefined maximum rollout length.

Each muscle actuator is assigned to an anatomical family according to its actuator name. Specifically, each category contains a set of predefined name keywords. A muscle is assigned to category $C_k$ if its actuator name contains at least one keyword belonging to that category:
\begin{equation}
m \in C_k
\quad \Longleftrightarrow \quad
\exists q \in \mathcal{Q}_k
\ \mathrm{s.t.}\
q \subseteq \mathrm{name}(m),
\end{equation}
where $\mathcal{Q}_k$ denotes the keyword set of the $k$-th muscle category.

The absolute activation mass of each muscle family is then computed by summing the mean activations of all muscles assigned to that family:
\begin{equation}
A_k =
\sum_{m \in C_k}
\bar{a}_m .
\end{equation}
Thus, muscle-family activation is computed by summing the time-averaged activations of its constituent actuators. The quantity is not a temporal integral or a cumulative sum over timesteps.

The total activation mass across all muscle families is:
\begin{equation}
A_{\mathrm{total}}
=
\sum_{k=1}^{K}
A_k .
\end{equation}

Finally, the relative contribution of each muscle family is reported as:
\begin{equation}
P_k =
\frac{A_k}{A_{\mathrm{total}}}
\times 100\% .
\end{equation}

The full anatomical grouping rule used in our analysis is summarized in Table~\ref{tab:muscle_categories}. This categorization covers the major trunk, upper-limb, lower-limb, and hand muscle groups in the 416-muscle model.

\begin{table*}[!htbp]
    \centering
    \caption{Anatomical muscle-family categories used for whole-body recruitment analysis. Each actuator is assigned to a category according to whether its name contains one of the corresponding keywords.}
    \label{tab:muscle_categories}
    \renewcommand{\arraystretch}{1.15}
    \begin{tabular}{
        @{}
        >{\raggedright\arraybackslash}p{3.8cm}
        >{\raggedright\arraybackslash}p{4.2cm}
        >{\raggedright\arraybackslash}p{7.8cm}
        @{}
    }
    \toprule
    \textbf{Body Region} & \textbf{Muscle Category} & \textbf{Actuator Name Keywords} \\
    \midrule

    \multirow{6}{*}{Core and Back}
    & Paraspinals (Deep) 
    & \texttt{Ps\_} \\

    & Iliocostalis 
    & \texttt{IL\_} \\

    & Longissimus (Erector Spinae) 
    & \texttt{LTpT}, \texttt{LTpL} \\

    & Abdominals 
    & \texttt{rect\_abd}, \texttt{EO}, \texttt{IO} \\

    & Other Trunk/Back 
    & \texttt{LAT}, \texttt{LD}, \texttt{QL}, \texttt{MF\_m}, \texttt{TRAP}, \texttt{RHOM} \\

    & Pectorals 
    & \texttt{PEC} \\
    \midrule

    \multirow{8}{*}{Upper Extremity}
    & Deltoids 
    & \texttt{DELT} \\

    & Rotator Cuff 
    & \texttt{SUPSP}, \texttt{INFSP}, \texttt{SUBSC}, \texttt{TMIN}, \texttt{TMAJ}, \texttt{CORB} \\

    & Elbow Flexors 
    & \texttt{BIC}, \texttt{BRA}, \texttt{BRD} \\

    & Elbow Extensors 
    & \texttt{TRI}, \texttt{ANC} \\

    & Forearm Pronators/Supinators 
    & \texttt{SUP}, \texttt{PT}, \texttt{PQ} \\

    & Wrist/Finger Flexors 
    & \texttt{FCR}, \texttt{FCU}, \texttt{PL}, \texttt{FDS}, \texttt{FDP}, \texttt{FPL} \\

    & Wrist/Finger Extensors 
    & \texttt{ECR}, \texttt{ECU}, \texttt{EDC}, \texttt{EDM}, \texttt{EIP}, \texttt{EPL}, \texttt{EPB}, \texttt{APL} \\

    & Hand Intrinsics 
    & \texttt{OP}, \texttt{RI2}, \texttt{RI3}, \texttt{RI4}, \texttt{RI5}, \texttt{LU\_}, \texttt{UI\_} \\
    \midrule

    \multirow{6}{*}{Lower Extremity}
    & Glutes (Hip Extensors) 
    & \texttt{GMAX}, \texttt{GMED}, \texttt{GMIN}, \texttt{PIR}, \texttt{GEM}, \texttt{OINT}, \texttt{QFM}, \texttt{glmax}, \texttt{glmed}, \texttt{glmin}, \texttt{piri} \\

    & Quadriceps (Knee Extensors) 
    & \texttt{VAS}, \texttt{RF}, \texttt{VI}, \texttt{VL}, \texttt{VM}, \texttt{vasint}, \texttt{vaslat}, \texttt{vasmed}, \texttt{recfem} \\

    & Hip Flexors/Adductors 
    & \texttt{ILIACUS}, \texttt{PSOAS}, \texttt{ADD}, \texttt{PECT}, \texttt{GRAC}, \texttt{TFL}, \texttt{iliacus}, \texttt{psoas}, \texttt{add}, \texttt{grac}, \texttt{tfl}, \texttt{sart} \\

    & Hamstrings (Knee Flexors) 
    & \texttt{BFLH}, \texttt{BFSH}, \texttt{SEMIM}, \texttt{SEMIT}, \texttt{bflh}, \texttt{bfsh}, \texttt{semiten}, \texttt{semimem} \\

    & Calves (Plantar Flexors) 
    & \texttt{GAS}, \texttt{SOL}, \texttt{soleus}, \texttt{gaslat}, \texttt{gasmed} \\

    & Other Leg/Foot 
    & \texttt{TA}, \texttt{EDL}, \texttt{EHL}, \texttt{FDL}, \texttt{FHL}, \texttt{TP}, \texttt{PER}, \texttt{edl}, \texttt{ehl}, \texttt{fdl}, \texttt{fhl}, \texttt{perbrev}, \texttt{tibant}, \texttt{tibpost}, \texttt{perlong} \\

    \bottomrule
    \end{tabular}
\end{table*}

Muscles that do not match any predefined keyword are assigned to an \textit{Unclassified} group. During analysis, we monitor the activation mass of this group to verify that the predefined anatomical categories cover the majority of meaningful muscle activity. If the activation mass of the unclassified group is negligible, the resulting recruitment distribution can be considered a reliable summary of whole-body muscle usage.

This categorization allows us to compare policies not only by task reward or success rate, but also by biomechanical recruitment strategy. For example, an efficient powerlifting policy should exhibit stronger recruitment of the posterior chain and trunk stabilizers, while a locomotion policy should show coordinated contributions from hip extensors, quadriceps, hamstrings, and plantar flexors.
\subsection{Detailed Descriptions of Residual RL Policy}
\label{appendix: residual_rl}

To adapt the base tracking policy for robust locomotion and overcome the limitations of strict kinematic imitation across various terrains, we introduce several critical modifications to the standard Residual Reinforcement Learning (RRL) framework. The reported results characterize the combined pipeline of residual corrections, task-objective changes, and reference pacing. Tracking divergence on stairs alone does not identify the cause of failure, and these experiments do not isolate the contribution of each component.

\subsubsection{Action Space and Residual Scaling}
Instead of allowing the residual policy to fully overwrite the base policy, we constrain the residual action space to ensure stability. The final blended action $a_t$ sent to the musculoskeletal simulation is defined as:
\begin{equation}
\begin{aligned}
a_t &= \text{clip}\left( a_{\text{base}} + \alpha \cdot \text{clip}(a_{\text{res}}, -1.0, 1.0), -1.0, 1.0 \right)
\end{aligned}
\end{equation}
where $a_t$ is the final blended action at time step $t$, $a_{\text{base}}$ is the action inferred by the frozen base policy, $a_{\text{res}}$ is the output of the residual network, and $\alpha$ is the residual scaling factor. This control ratio grants the residual policy sufficient authority to adjust muscle activations while preserving the fundamental synergistic patterns learned by the base policy. Furthermore, we deliberately disable standard kinematic penalties (e.g., action rate and activation-cost penalties) to encourage aggressive exploration, retaining only a minimal L2 regularization on the residual action ($c_{\text{res}} = -0.02$).

\subsubsection{Adaptations for Stair Navigation}
A fundamental challenge in mimicking flat-ground or mismatched reference trajectories on physical stairs is the accumulation of vertical errors. If penalized for absolute height discrepancies, the policy prematurely terminates. 

\textbf{Z-Axis Agnostic Imitation Objective:} We address this by masking the Z-axis displacement in the root position error calculation:
\begin{equation}
\begin{aligned}
d_{\text{root}} &= \| \mathbf{p}_{xy} - \mathbf{\hat{p}}_{xy} \|_2^2
\end{aligned}
\end{equation}
where $d_{\text{root}}$ is the calculated root position error, and $\mathbf{p}_{xy}$ and $\mathbf{\hat{p}}_{xy}$ are the 2D planar coordinates of the simulated and reference root positions, respectively. This height-invariant formulation allows the humanoid to step onto elevated platforms without suffering catastrophic drops in the tracking reward. 

\textbf{Elastic Trajectory Pacing:} To prevent the policy from being dragged forward when encountering physical obstacles (e.g., tripping on a stair edge), we decouple the reference step from the physical simulation time. The environment implements an \textit{elastic tracking} mechanism: the reference advances normally when the planar root deviation is below 0.4~m and slows by 50\% otherwise.

\textbf{Task-Driven Progress Reward:} To break the deadlock of purely conservative imitation, we augment the imitation reward with a task-driven progress reward:
\begin{equation}
\begin{aligned}
r_{\text{prog}} &= w_{\text{prog}} \Big( \text{clip}(v_{\text{proj}}, 0, 2.0) \\
&\quad + 0.5 \cdot \text{clip}(v_z, 0, 1.5) \Big)
\end{aligned}
\end{equation}
where $r_{\text{prog}}$ is the calculated progress reward, $w_{\text{prog}} = 0.3$ is the weight coefficient for this reward, $v_{\text{proj}}$ is the current root velocity projected onto the directional vector pointing to the future reference pelvis position (1.0s ahead), and $v_z$ is the absolute vertical climbing velocity. 

\subsubsection{Walking Adaptation}
While navigating stairs requires aggressive spatial exploration, flat-ground walking prioritizes energy efficiency and stable transitions between stance and swing phases. Strict kinematic imitation often forces the musculoskeletal model into stiff, metabolically inefficient gaits. 

\textbf{Phase-Dependent Residual Authority:} We dynamically modulate the residual network's authority based on the gait phase. The swing leg requires more flexibility to adjust foot clearance, whereas the stance leg demands strict synergistic stability. The control ratio is formulated as:
\begin{equation}
\begin{aligned}
\alpha_i(t) &= \alpha_{\text{stance}} + (\alpha_{\text{swing}} - \alpha_{\text{stance}}) \cdot \mathcal{I}_{\text{swing}}(i, t)
\end{aligned}
\end{equation}
where $\alpha_i(t)$ represents the dynamic residual scaling factor for limb $i$ at time step $t$, $i$ denotes the respective limb, $\alpha_{\text{stance}} = 0.2$ and $\alpha_{\text{swing}} = 0.6$ are the base control ratios for the stance and swing phases respectively (the latter is used for the stair task as well), and $\mathcal{I}_{\text{swing}}(i, t)$ is a boolean indicator function derived from foot-ground contact sensors.

\textbf{Activation-Based Muscle Regularization:} To eliminate high-frequency, non-functional muscle twitches, we reinstate an activation-magnitude and temporal-change penalty specifically for the walk task:
\begin{equation}
\begin{aligned}
p_{\text{met}} &= -w_{\text{met}} \sum_{j=1}^{N} \left( a_{j, t}^2 + \lambda (a_{j, t} - a_{j, t-1})^2 \right)
\end{aligned}
\end{equation}
where $p_{\text{met}}$ is the activation regularization penalty, $N=416$ is the total number of muscles, $a_{j, t}$ is the activation of muscle $j$ at the current time step $t$, $a_{j, t-1}$ is the activation of the same muscle at the previous time step, $w_{\text{met}} = 0.05$ is the scaling weight for the penalty, and $\lambda = 0.1$ is the smoothing coefficient. This regularizes activation magnitude and temporal variation; task success and human-reference agreement are evaluated separately.

\subsubsection{Adaptations for High-Impact Locomotion (Run Task)}
Running fundamentally differs from walking due to the presence of an aerial flight phase. Discrepancies in ground contact timing between the reference kinematic data and the physics simulation often cause failures during impact.

\textbf{Flight-Phase Propulsive Reward:} During the aerial phase, we temporarily relax the tracking penalty by 50\% and introduce a propulsive velocity reward:
\begin{equation}
\begin{aligned}
r_{\text{propel}} &= w_{\text{vel}} \cdot \max(0, v_x - \hat{v}_x) \cdot \mathcal{I}_{\text{air}}
\end{aligned}
\end{equation}
where $r_{\text{propel}}$ is the propulsive velocity reward, $w_{\text{vel}}$ is the weight coefficient for this reward, $v_x$ is the current forward velocity, $\hat{v}_x$ is the target running speed, and $\mathcal{I}_{\text{air}}$ indicates the flight phase. This encourages explosive muscle forces during toe-off.

\textbf{Ground Reaction Force (GRF) Smoothing:} High-speed impacts cause severe numerical instability in the simulation. We introduce a soft penalty for excessive vertical Ground Reaction Forces (GRF):
\begin{equation}
\begin{aligned}
p_{\text{impact}} &= -w_{\text{grf}} \cdot \max(0, F_z - F_{\text{limit}})
\end{aligned}
\end{equation}
where $p_{\text{impact}}$ is the soft impact penalty, $w_{\text{grf}}$ is the weight coefficient for the GRF penalty, $F_z$ is the total measured vertical contact force, and $F_{\text{limit}}$ is set to $2.5 \times$ the humanoid's body weight. This forces the residual policy to learn active shock absorption strategies prior to touchdown.

\subsection{Extended Discussion on Limitations and Future Directions}
\label{sec:app-limitations}

While the Musculoskeletal Humanoid Benchmark provides a comprehensive and diverse foundation for muscle-driven control, it is essential to acknowledge several limitations in the current scope of this work. These limitations also highlight promising avenues for future research in biomechanical reinforcement learning and humanoid robotics.

\subsubsection{Absence of Dexterous Manipulation} 
The current iteration of our benchmark primarily concentrates on gross motor skills, such as full-body locomotion, postural stabilization, and basic environmental interactions (e.g., pushing a door or reaching). However, one of the most defining characteristics of the human musculoskeletal system is its extraordinary capacity for fine motor control, particularly in the hands and fingers. Modeling the biomechanics of the human hand is highly complex due to the dense routing of tendons, high degrees of freedom, and intricate muscle synergies. Consequently, our benchmark currently lacks fine-grained dexterous manipulation tasks, such as in-hand object manipulation, tool use, or precise grasping. Future work should integrate high-fidelity musculoskeletal hand models into the full-body framework, enabling the study of how gross arm movements dynamically couple with fine finger manipulations.

\subsubsection{Transitioning to Compositional and Long-Horizon Tasks} 
Currently, our benchmark evaluates policies on isolated, single-objective tasks (e.g., exclusively walking, or solely executing a powerlift). Real-world humanoid operation, however, inherently demands the execution of compositional and long-horizon behaviors. For instance, a realistic scenario might require the robot to walk toward a chair, pick up a barbell along the way, and subsequently perform a seated lift. Addressing such long-horizon tasks will require policies to not only master atomic skills but also smoothly transition between them without losing structural balance or experiencing catastrophic energy depletion. Future iterations of this benchmark will aim to include compositional task suites, serving as a testbed for hierarchical reinforcement learning (HRL), skill discovery, and long-term fatigue modeling in muscle-driven systems.

\subsubsection{Constraints in Comprehensive Physiological Validation} 
A core motivation of musculoskeletal modeling is to generate biologically plausible behaviors. However, our rigorous physiological validation—specifically the comparison between policy-generated muscle activations and ground-truth human Electromyography (EMG) signals—currently covers walking, running, and stairs with selected matched lower-limb muscles. The normalized, phase-optimized correlation assesses activation-envelope shape rather than complete biological similarity. This limitation stems from a scarcity of comprehensive, open-source EMG datasets for highly dynamic or unconventional tasks (e.g., hurdle clearance, whole-body crawling, or wobble board balancing). To achieve a more universal physiological benchmark, future research must involve establishing advanced data collection pipelines—synchronizing high-density EMG with full-body motion capture—to build a diverse dataset covering a broader spectrum of human motor skills.

\subsubsection{Sim-to-Real Gap and Hardware Validation} 
Finally, all evaluations within this benchmark are conducted in a simulated physics environment. Although the simulation enforces rigorous biomechanical and physical constraints, it inevitably abstracts away certain real-world complexities, such as sensor noise, communication latency, and unmodeled actuator dynamics. Validating these muscle-driven policies on physical hardware remains a formidable challenge. Real-world musculoskeletal humanoids (e.g., tendon-driven robots) are notoriously difficult to build and maintain. Future directions include bridging the sim-to-real gap by either deploying these policies on state-of-the-art tendon-driven hardware or exploring methodologies to map virtual muscle activations (acting as variable stiffness/compliance signals) onto traditional electric-motor-driven humanoids equipped with whole-body impedance control.

\end{document}